\documentclass[10pt,twocolumn,letterpaper]{article}

\usepackage{wacv}              %

\usepackage{times}
\usepackage{epsfig}
\usepackage{graphicx}
\usepackage{amsmath}
\usepackage{amssymb}
\usepackage{comment}
\usepackage{graphicx}
\usepackage[export]{adjustbox}
\usepackage{caption}
\usepackage{subcaption}
\usepackage{soul}
\usepackage{breqn}
\usepackage{enumitem}
\setlist{nosep,leftmargin=*} %
\usepackage{float}
\usepackage{ragged2e}
\usepackage{setspace}

\usepackage{xcolor}
\newcommand{\colornote}[3]{{\color{#1}\bf{#2 #3}\normalfont}}
\newcommand{\kf}[1]{\colornote{blue}{}{}}

\usepackage[pagebackref,breaklinks,colorlinks]{hyperref}

\usepackage[capitalize]{cleveref}
\crefname{section}{Sec.}{Secs.}
\Crefname{section}{Section}{Sections}
\Crefname{table}{Table}{Tables}
\crefname{table}{Tab.}{Tabs.}

\def\wacvPaperID{1552} %
\def\confName{WACV}
\def\confYear{2024}

\begin{document}

\title{Collage Diffusion}

\author{Vishnu Sarukkai, Linden Li, Arden Ma, Christopher Ré, Kayvon Fatahalian \\ \\ Stanford University}

\setcounter{figure}{1}
\twocolumn[{%
\renewcommand\twocolumn[1][]{#1}%
\maketitle
\vspace{-2.5em}
\begin{center}
	\hypertarget{fig:teaser}{}
   \includegraphics[width=.7\textwidth]{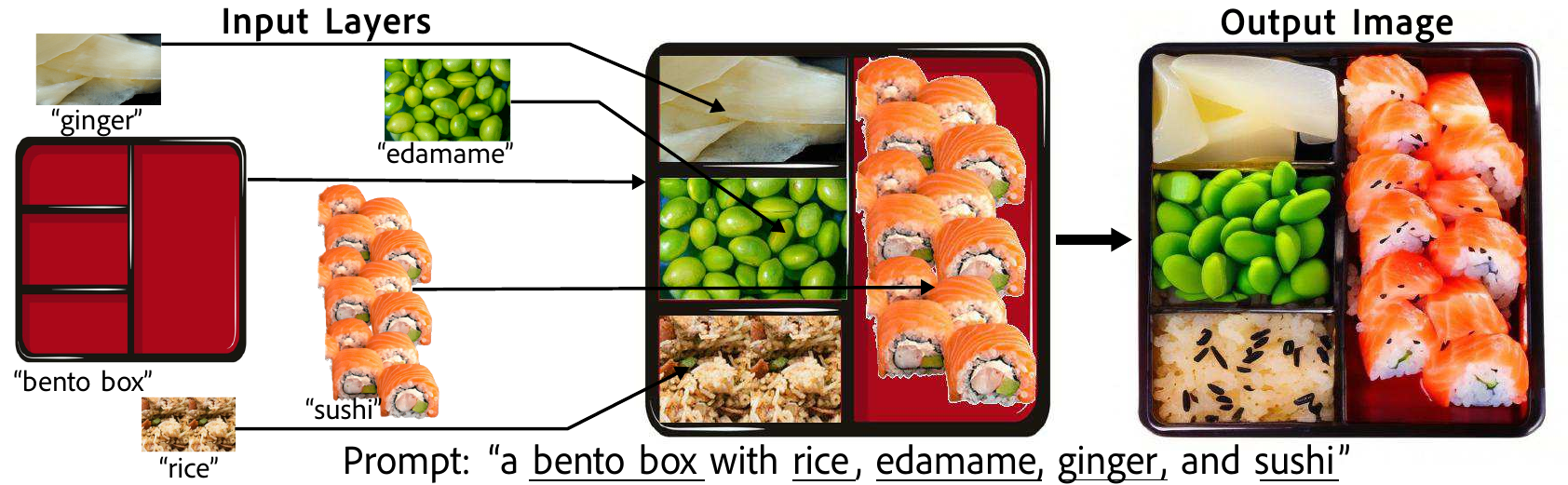}
   \vspace{-1em}

	\begin{justify} Figure 1: A layer is defined as an image-text pair. Given a sequence of layers, \textit{Collage Diffusion} generates an image that is globally harmonized, yet preserves the locations and key visual characteristics of objects in each input layer.
   \end{justify}
\end{center}%
}]

\begin{abstract}
        We seek to give users precise control over diffusion-based image generation by modeling complex scenes as sequences of layers, which define the desired spatial arrangement and visual attributes of objects in the scene.
        \textit{Collage Diffusion} harmonizes the input layers to make objects fit together---the key challenge involves minimizing changes in the positions and key visual attributes of the input layers while allowing other attributes to change in the harmonization process. 
        We ensure that objects are generated in the correct locations by modifying text-image cross-attention with the layers' alpha masks. 
        We preserve key visual attributes of input layers by learning specialized text representations per layer and by extending ControlNet to operate on layers. 
        Layer input allows users to control the extent of image harmonization on a per-object basis, and users can even iteratively edit individual objects in generated images while keeping other objects fixed.
        By leveraging the rich information present in layer input, \textit{Collage Diffusion} generates globally harmonized images that maintain desired object characteristics better than prior approaches. 
\end{abstract}

\section{Introduction}
\label{sec:intro}
\vspace{-0.5em}

Diffusion-based image generation \cite{sohl2015deep,song2019generative,ho2020denoising, dhariwal2021diffusion, dalle2, latentDiffusion} has captured widespread interest with its seemingly magical ability to generate plausible images from a text prompt. Unfortunately, text is a highly ambiguous specification of an image, forcing users to spend significant time tweaking prompt strings to obtain a desired output. 
A body of recent work has therefore focused on providing more precise controls for scene composition via additional inputs: controlling composition via sketching\,\cite{ediffi}, filling in user-provided segmentation masks\,\cite{avrahami2022blended,objectstitch}, providing an image seed for generation\,\cite{sdedit}, etc. Similarly, the desire to precisely dictate object appearance, \emph{``the sushi in THIS reference photo''} rather than \emph{``the sushi''}, has led to approaches that condition generation based on example images \cite{TextualInversion,Dreambooth,CustomDiffusion}.

We seek to give users precise control over image output when creating scenes featuring a collection of objects with a specific spatial arrangement. For example, in Figure \hyperlink{fig:teaser}{1}, ``A bento box with rice, edamame, ginger, and sushi'' neither describes what items go in which Bento bin, nor suggests how each of the items should look. Rather than relying on ambiguous text prompts or forcing the user to sketch scene forms, we return to a traditional and easy-to-create means of expressing artistic intent: \emph{making a sequence of layers} to express a desired scene layout and the appearance of the scene objects. To specify a scene, a user need only acquire reference images of desired scene objects (e.g., via image search or via output from an existing generative model), arrange them on a canvas using a traditional layer-based image editing UI, and pair each object with a text prompt. 

Given these layers, we introduce \emph{Collage Diffusion}, a diffusion-based image harmonization algorithm that generates images that 1) have \emph{fidelity} to the input layers' spatial composition and object appearance, but 2) exhibit global \emph{harmonization} and visual coherence that is representative of ``plausible'' real-world images.  
There is an inherent tradeoff between harmonization and fidelity: harmonization involves changing properties of the input layers so that objects ``fit together'' in a consistent image, while fidelity involves preserving properties of the layers. 
The key challenge is harmonizing a sequence of layers while limiting variation in certain layer properties (color, texture, edge maps, etc.), but allowing variation in other properties. 
We tackle this challenge by leveraging the rich information present in layer input---building upon prior diffusion-based techniques for image harmonization, spatial control, and appearance control, we extend each approach for performance with layers, in particular focusing on mechanisms for per-layer control. 

Specificially we make the following contributions:

\begin{enumerate}%
	\item We introduce layer-conditioned diffusion, where generation is conditioned on alpha-composited RGBA layers as well as text prompts describing the content of each layer. Sequences of layers can be authored by users in minutes, and \textit{Collage Diffusion} generates high-quality images that respect both the desired scene composition and object appearance, even for complex scenes with many layers. 

	\item We extend prior diffusion-based control mechanisms \cite{ediffi,TextualInversion,zhang2023adding} to operate on sequences of layers, ensuring that output images adhere to the composition depicted by the layers (cross-attention) and retain salient visual features of objects in each layer (textual inversion, ControlNet). 

	\item 
	We illustrate how layer input allows users to control the harmonization-fidelity tradeoff on a per-layer basis and also enables users to iteratively refine generated images.
\end{enumerate}
\vspace{-0.5em}
\section{Problem Definition and Goals}
\label{sec:problem}
\vspace{-0.5em}

Our goal is to generate globally harmonized images that respect a user's desired scene composition, both in terms of \textit{spatial fidelity}, i.e., preserving the positions and sizes of the desired objects, as well as \textit{appearance fidelity}, i.e., preserving the visual characteristics of the objects. 
We propose that the user describe their intent by means of a sequence of layers alongside a global text prompt. For brevity, we call this combination a \emph{collage}. 
We first define a collage, then introduce our goals for collage-conditional generation. 

As illustrated in \cref{fig:method}, we define collage $C$ as:
\begin{enumerate}%
    \item A full-image text string $c$, describing the entire image to be generated (``A bento box with rice, edamame, ginger, and sushi'')
    \item A sequence of $n$ layers $l_1, l_2, ... l_n$, ordered from back to front, with each $l_i$ having:
    \begin{enumerate}[labelsep=0.5em,itemsep=-0.25em]
        \item An RGBA image $x_i$ (the alpha-masked input image of sushi), with alpha layer $x^\alpha_i$ 
        \item A text string $c_i$ describing the layer, which is a substring of $c$ (``sushi'')
    \end{enumerate} 
\end{enumerate}

Given input collage $C$, 
we seek to generate output image $x^*_c$ with the following properties:
\begin{enumerate}%
    \item \textit{Global harmonization}: $x^*_c$ is harmonized, having the consistency of a real image. In Figure~\hyperlink{fig:teaser}{1}, the output features consistent perspective, lighting, and occlusions among scene objects. 
    \item \textit{Spatial fidelity}: for all layers $l_i$, the objects described by layer text $c_i$ are generated in the correct regions of $x^*_c$. In Figure~\hyperlink{fig:teaser}{1}, ``edamame,'' ``ginger,'' etc. are all in the same regions of the output image as in the input collage.
    \item \textit{Appearance fidelity}: for all layers $l_i$, in addition to matching layer text $c_i$, regions of $x^*_c$ that depict the contents of the layer share key visual characteristics with $x_i$. In Figure~\hyperlink{fig:teaser}{1}, the ``ginger'' in the output image remains sliced sushi ginger (not whole ginger), etc.
\end{enumerate}

In order to achieve the consistency of a real image, we aim to constrain both the spatial layout of generated images and certain aspects of the appearance of individual objects, allowing other aspects to vary in the harmonization process. 

\vspace{-1.5em}
\section{Related Work}
\vspace{-0.5em}

\begin{figure}
    \centering
    \includegraphics[width=.8\linewidth,valign=m]{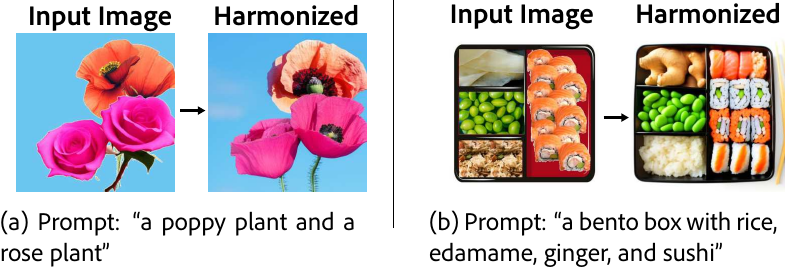}
    \caption{Without layer information, image harmonization can lead to a loss of spatial and appearance fidelity. Added noise can disrupt object-location mappings---on the left, ``poppies'' take the place of the ``roses.'' Added noise also can obscure specifics of an object's appearance---on the right, the generated ``ginger'' is whole instead of sliced.}
    \label{figure:ca_ft_motivation}
\end{figure}

One natural starting point is to ``flatten'' the input layers into image $x_c$ by alpha-compositing the sequence of layer images $x_1, x_2, ...$ into a single image \cite{porter1984compositing}, then use diffusion-based image harmonization to improve the visual quality of the image \cite{sdedit,avrahami2022blended,saharia2022palette}. 
Diffusion-based approaches can harmonize geometry \cite{sdedit,objectstitch}, rather than restricting focus to color and lighting \cite{cong2020dovenet,cong2021bargainnet,hong2022shadow,xue2022dccf}.
The problem with this flatten-then-harmonize algorithm is that generated results may diverge from the content of the initial image, undermining user intent. For example, in \cref{figure:ca_ft_motivation}, noise-based harmonization \cite{sdedit} turns the pink roses into poppies despite the prompt and turns sliced sushi ginger into whole ginger. 
We seek to better maintain the spatial and appearance fidelity of the initial layers. 

\vspace{-1.5em}
\paragraph{Improving Spatial Fidelity}

Prior work has suggested approaches to (1) define spatial layouts of scene objects, and then (2) generate objects according to the desired layout. 
Existing techniques define spatial layouts using segmentation maps, whether defining a region for inpainting \cite{avrahami2022blended,sdedit,objectstitch,chen2023anydoor,yang2023paint} or providing a full-image segmentation map \cite{spatext,ediffi}. 
(Reference-based) inpainting approaches struggle to maintain global coherence with many layers (see Appendix). 
Instead of hand-drawing a segmentation map, we see layers as an intuitive, alternative way to specify spatial composition. 
\vspace{-1.5em}

\paragraph{Improving Appearance Fidelity}
In addition to generating objects in the desired locations, we aim to preserve visual characteristics of input layers. 
Several recent works specialize diffusion models to particular visual concepts (objects, styles, etc.) \cite{TextualInversion,Dreambooth,CustomDiffusion}, requiring several input images and either fine-tuning the model \cite{Dreambooth,CustomDiffusion}, learning a specific textual representation for the object \cite{TextualInversion}, pretraining on reference images \cite{xiao2023fastcomposer}, or reverse-engineering a prompt for a given image \cite{PEZ}. 
These methods struggle to generate high-quality images of scenes with compositions of many objects \cite{TextualInversion,Dreambooth,CustomDiffusion,xiao2023fastcomposer}.
In addition, appraches that fine-tune model weights require either joint multi-concept training or post-hoc combination of model weights, both of which struggle in regimes with several objects \cite{Dreambooth,CustomDiffusion}.
Alternatively, ControlNet \cite{zhang2023adding} enables us to preserve derived features of input layers (edge maps, pose, etc.) without learning a visual concept personalized to the specific object. 

We address the goal of appearance fidelity by extending both textual inversion \cite{TextualInversion} and ControlNet \cite{zhang2023adding} for performance with individual layers. 
We find that the learned representations are effective for maintaining key visual characteristics of input layers when paired with techniques for spatial control. 
When preserving an image structure from an input layer such as an edge map, our extension of ControlNet is effective. 
\vspace{-1.5em}

\begin{figure*}[!t]
    \centering
    \includegraphics[width=.8\linewidth,valign=m]{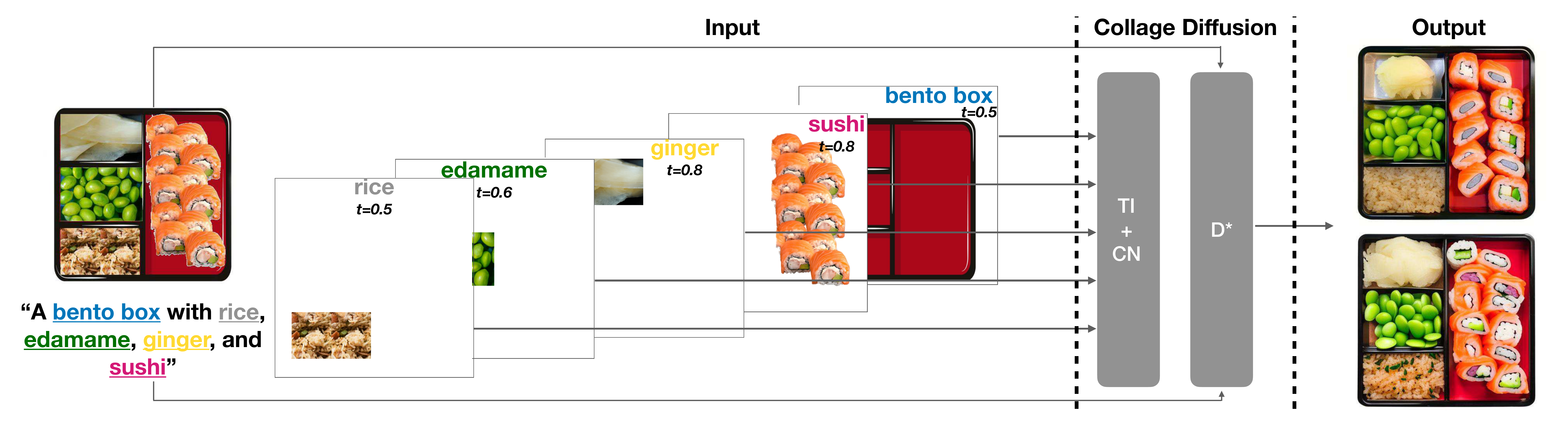}
    \vspace{-1.5em}
    \caption{\textit{Collage Diffusion} takes as input a sequence of layers of RGBA images paired with text (the image of sushi and the text ``sushi''), along with a full-image text string (``A bento box with rice, edamame, ginger, and sushi''). Layer information enables 1) manipulating cross-attention to map individual layers to the corresponding image regions, creating improved diffusion model $D^*$, 2) learning layer-specific representations using textual inversion (TI), 3) having the option to preserve per-layer image structures with ControlNet (CN), and 4) harmonizing layers according to per-layer noise levels $t_i$. \textit{Collage Diffusion} outputs globally-harmonized images that contain objects in the specified locations, and share visual characteristics with the input layer images. 
	\textbf{\emph{In the rest of the paper, for brevity we only display the layer composite image and prompt, and we use underlined substrings to indicate contents of individual layers.}} }
    \label{fig:method}
\end{figure*}

\paragraph{Image-to-Image Approaches} \label{related:img2img}
Constrained image harmonization can also be framed as image stylization: from low-quality layer composite to high-quality harmonized output. 
Stylization can be approached using existing methods for controlled image-to-image diffusion \cite{prompt2prompt,brooks2022instructpix2pix,tumanyan2022plug,mokady2022null,zhang2023adding}. 
Derived features (canny edges, pose, etc.) can provide control \cite{zhang2023adding}, but fails to constrain scene composition---the locations of objects are not preserved. 
Other methods directly \cite{prompt2prompt, tumanyan2022plug,mokady2022null} or indirectly \cite{brooks2022instructpix2pix} manipulate U-Net attention layers (cross-attention \cite{prompt2prompt,brooks2022instructpix2pix,mokady2022null} and self-attention \cite{tumanyan2022plug}) to maintain image structure while making either local edits (adding/removing/modifying objects) or global edits (style, lighting). 
Unfortunately, this approach is insufficient for layer-conditional diffusion.
\emph{Input layers often need to be changed significantly} to fit together in a harmonized image, as objects may need to be rotated, partially occluded, etc. (see the orientation of the sushi in \cref{fig:method}).
This is difficult when preserving the ``structure'' of the input image. 
We evaluate against one constrained image-to-image approach \cite{tumanyan2022plug}, and discuss additional baselines in the Appendix. 
Less constrained harmonization techniques \cite{sdedit} serve as a more useful starting point for \textit{Collage Diffusion} since they allow the desired flexibility in image structure. 
\vspace{-1.5em}

\paragraph{Layered Image and Video Editing}
Layer-based image and video editing is well-established in computer graphics \cite{porter1984compositing,wang1994representing} and is being increasingly adopted in machine learning-driven methods \cite{bar2022text2live, latentPrior, kasten2021layered, lu2020layered}. 
Layered representations allow modification of individual components in images \cite{bar2022text2live, latentPrior} and in video \cite{bar2022text2live, kasten2021layered, lu2020layered}.
This process often requires generating a layered representation from a single input video or image.
In contrast, we assume that layered information is provided as input, using machine learning to synthesize image output from the layers. 

\vspace{-0.5em}
\section{Collage Diffusion}
\vspace{-0.5em}

To frame discussion of layer-based image harmonization, we first recap how text-conditioned diffusion models can perform image harmonization by leveraging added noise. 
Then, we describe how \textit{Collage Diffusion} leverages additional information from individual layers to increase both spatial and appearance fidelity for harmonized output.
\vspace{-0.5em}

\subsection{Global image harmonization} \label{method:SDEdit}
\vspace{-0.5em}

Leveraging only layer composite image $x_c$ and full-image string $c$, the SDEdit algorithm \cite{sdedit} improves image quality by adding Gaussian noise with standard deviation $\sigma(t)$ to $x_c$, then denoising the noised image $x_t = x_c + \mathcal{N}(0,\sigma(t)^2)$ to generate output image $x^*_c$, using a text-conditional diffusion U-Net $D_\theta(x,\sigma(t),c)$ as an image prior \cite{latentDiffusion} ($x$ is a noisy input image, $\sigma(t)$ is the noise level at time $t$, and $c$ is the text conditioning).
Unfortunately, added noise can make it difficult to map objects to the correct image regions and can obscure key visual details, reducing spatial and appearance fidelity to the original layers (\cref{figure:ca_ft_motivation}). 
Layer input, with text $c_i$ and image $x_i$ corresponding to each region of the image, provides additional information facilitating more precise control over individual components of the generated image. 

\subsection{Spatial fidelity: cross-attention manipulation} \label{method:CAC}
\vspace{-0.5em}

To generate an image with the desired objects in the desired locations, 
\textit{Collage Diffusion} modifies the text-image cross-attention 
in text-conditional U-Net model $D_\theta$. 
Not all tokens in full-image input text $c$ correspond to layer strings $c_i$---the start token, end token, several words in the input string, and padding tokens lack specific regional influence. 
We refer to these tokens as ``global'' tokens, while layer-specific tokens are ``layer'' tokens. 
For instance, in \cref{fig:method}, ``with'' is a global token and ``rice'' is a layer token. 
\textit{Collage Diffusion} constrains image generation by restricting the influence of layer tokens to the regions of the image where the corresponding layer is visible.%
The visible layer at pixel coordinate $(a,b)$ is defined as $j = \max\limits_{k \in 1 \dots n}(\{k | (x^\alpha_{k})_{ab} > 0\})$, where $j$ is the layer index of the highest of the $n$ layers with non-zero alpha at pixel coordinate $(a,b)$. 

Cross-attention in $D_\theta$ is computed as $\mathrm{softmax}\big(\frac{QK^T}{\sqrt{d}}\big)V$, where $Q$ is a matrix of query embeddings from image tokens, $K$ is a matrix of key embeddings from text tokens, $V$ is a matrix of value embeddings from text tokens, and $d$ is the embedding dimensionality.
To increase or decrease the influence of a particular token on a part of the image, \textit{Collage Diffusion} alters $QK^T$, an approach similar to the mechanism proposed by eDiffI \cite{ediffi}. 
Like eDiffI, \textit{Collage Diffusion} uses positive attention map $A^{pos}$ to increase the influence of layer tokens on a region relative to global tokens, but unlike eDiffI, \textit{Collage Diffusion} also constructs negative map $A^{neg}$ to prevent layer tokens from influencing regions outside the desired location.

To alter $QK^T$, \textit{Collage Diffusion} constructs attention maps $A^{pos}, A^{neg} \in \mathbb{R}^{N_v \times N_t}$, where $N_v$ is the number of image tokens and $N_t$ is the number of text tokens, and each column $A^{pos}_j$, $A^{neg}_j$ is a flattened alpha mask dependent on visibility of text token $j$. $A_{ij} = 0$ for all global tokens $j$. $A^{pos}_{ij} = 1$ if image token $i$ corresponds to a region of the image that layer token $j$ should influence, and $A^{neg}_{ij} = 1$ if image token $i$ corresponds to a region of the image that layer token $j$ should not influence. 
Along with scalar weights $w^{pos}$ and $w^{neg}$, attention maps $A^{pos}$ and $A^{neg}$ are incorporated into the softmax operation: $\mathrm{softmax}\big(\frac{QK^T + w^{pos}A^{pos} - w^{neg}A^{neg}}{\sqrt{d}}\big)V$. 
With larger weights $w^{pos}$ and $w^{neg}$, the influence of attention maps $A^{pos}$ and $A^{neg}$ on image layout is greater.
Weights $w^{pos}$ and $w^{neg}$ vary dependent on noise level $\sigma(t)$ throughout the diffusion process: $w^{pos} = v^{pos} \cdot \log(1 + \log(1 + \sigma(t))) \cdot \max(QK^T)$ and $w^{neg} = v^{neg} \cdot \log(1 + \log(1 + \sigma(t))) \cdot \max(QK^T)$, where $v^{pos}$ and $v^{neg}$ are scalars specified by the user. 
Denote this modified diffusion model as $D^*_\theta$.

\subsection{Appearance fidelity: inversion and ControlNet} \label{method:TI}
\vspace{-0.5em}

Layer text $c_i$ for a given layer often fails to adequately capture the intended appearance of layer image $x_i$.
For instance, in \cref{fig:method}, layer text ``ginger'' does not capture that the ginger is pickled and sliced. 
Starting image $x_c$ provides some guidance on the desired look of each layer, but the influence of $x_c$ is reduced when noise is added to the image. 
Therefore, to preserve visual fidelity, we refine the layer text to more accurately capture the layer's appearance. To do this, \textit{Collage Diffusion} builds upon Textual Inversion\,\cite{TextualInversion}:
layer text $c_i$ is specialized to image $x_i$ by learning a \textit{modifier} token $a_i$ per layer, prepended to the layer text: $(a_i,c_i)$. 
$a_i$ serves as an adjective describing the object in layer $l_i$, subject to the constraints of the existing layer description $c_i$. 
For instance, string ``ginger'' is modified into new string ``$\langle a_i \rangle$ ginger''. 
The embedding for $a_i$ is learned by optimizing the following loss:

\vspace{-0.5em}
\begin{dmath}
a^*_i = \mathrm{arg}\min_{a_i} E_{\epsilon \sim N(0,\sigma)}[x^\alpha_i\cdot(x_{target_i} - D_\theta(x_{target_i} + \epsilon, \sigma, (a_i,c_i)))]
\end{dmath}
\vspace{-0.5em}

\noindent target image $x_{target_i}$ is constructed by alpha-compositing the first $i$ layers $l_1 \dots l_i$, and layer alpha mask $x^\alpha_i$ restricts the loss to the relevant region of $x_{target_i}$. 

Textual Inversion\,\cite{TextualInversion} learns token $a_i$ as a standalone prompt, and performs optimization using several images of the same object that communicate invariances in pose, lighting, etc. 
\textit{Collage Diffusion} operates in a single-image setting, where $x_{target_i}$ is the only reference for learning $a_i$. Therefore, it leverages the layer textual description $c_i$ to help regularize optimization. 

We also extend ControlNet \cite{zhang2023adding} to preserve image structure on a per-layer basis. 
The ControlNet auxiliary network outputs 2-d feature maps $m_k \in R^{h, w, c}$ from its zero convolutions, where $h$ is height, $w$ is width, and $c$ is number of channels. 
In standard ControlNet, we multiply feature maps $m_k$ by scalar ControlNet weight $w_{all} \in [0, 1]$ that controls the ``strength'' with which ControlNet influences the generated image. 
We replace $w_{all}$ with weight map $w_{layer}$:
the user sets ControlNet weights $w_i$ for each layer $l_i$, and the $w_i$ are converted into single-channel weight map $w_{layer}$: $w_{layer_ab} = t_j$, where $j = \max\limits_{k \in 1 \dots n}(\{k | (x^\alpha_{k})_{ab}  > 0\})$ is the layer index of the highest of the $n$ layers with nonzero alpha for pixel coordinate $(a,b)$, and $w_{layer_ab}$ is the value of $w_{layer}$ at pixel $(a,b)$.
We resize $w_{layer}$ to $[0, 1]^{h,w}$ using bilinear interpolation, then elementwise-multiply $w_{layer} * m_k$ to produce re-weighted ControlNet outputs. 
Now, the user can control the influence of ControlNet on regions corresponding to each layer with per-layer weights $w_i$. 

\vspace{-0.5em}
\subsection{Tuning the Harmonization-Fidelity Tradeoff}
\label{method:layerNoise} 
\vspace{-0.5em}
The content in the input layers must be modified to globally harmonize the image, and users may be willing to accept more variation for some objects than others. 
Layer input allows users to control the harmonization-fidelity tradeoff on a per-object basis by having users specify the desired level of harmonization per layer. 
The user sets noise levels $t_i$ for each layer $l_i$, and the $t_i$ are converted into single-channel noise image $h$: $h_{ab} = t_j$, where $j = \max\limits_{k \in 1 \dots n}(\{k | (x^\alpha_{k})_{ab}  > 0\})$ is the layer index of the highest of the $n$ layers with nonzero alpha for pixel coordinate $(a,b)$, and $h_{ab}$ is the value of $h$ at pixel $(a,b)$.
A Gaussian blur is applied to $h$ to smooth boundaries where the noise level changes sharply. 
Building upon Blended Diffusion \cite{avrahami2022blended}, \textit{Collage Diffusion} modifies the diffusion process so that different levels of noise are added to different regions of the image according to $h$, controlling the harmonization-fidelity tradeoff per layer:

\vspace{-1em}
\begin{dmath}
x'(t-1) = x(t-1) \cdot m(t) + (x_c + \mathcal{N}(0,\sigma(t-1)^2)) \cdot (1 - m(t))
\end{dmath}
\vspace{-0.5em}

\begin{dmath}
m_{ab}(t) =
\begin{cases}
    1 & \text{if }h_{ab}<t\\    
    0 & \text{if }h_{ab}\ge t
\end{cases}
\end{dmath}
\vspace{-0.5em}

\noindent where $x(t)$ is the original solver output at time $t$, $x'(t)$ is the modified solver output at time $t$, and $m(t)$ is a binary mask computed at time $t$ based on the noise image $h$. 
For instance, in \cref{fig:method}, $t_i=0.5$ for both the ``bento box'' and ``rice'' layers, $t_i=0.6$ for the ``edamame'' layer, and $t_i=0.8$ the ``sushi'' and ``ginger'' layers, indicating that the user would like a greater level of harmonization for the ginger and sushi than for the bento box and the rice. 

\vspace{-0.5em}
\subsection{Editing Individual Layers in Generated Images}
\label{method:sequentialGeneration}
\vspace{-0.5em}

Per-layer noise controls also enable layer-by-layer image editing. 
Especially for scenes with many objects, it can be difficult to look through large output galleries to find an example where all objects in the scene look \emph{exactly} as desired.
Rather, the user can simply select a generated image where nearly all the objects look as desired, then refine the image by generating alternate possibilities for the remaining objects. 

Per-layer noise controls enable users to keep a part of an input collage ``fixed'' by setting the noise level to $t=0$ for the layers that should remain constant. 
Having generated an image using \textit{Collage Diffusion}, an individual object may be edited by creating a new two-layer collage, where the generated image is the background layer, and the object to be re-generated is the foreground layer. Setting per-layer noise $t=0$ to the background layer, a variety of possibilities are generated for the foreground layer, harmonized and combined with the fixed background layer. 
Especially for complex scenes, a small part of a generated image might not quite look right.
Here, iterative, layer-driven editing can be the difference between obtaining a final image that is \emph{nearly} satisfactory and one that precisely satisfies the user's image generation goals.   

\vspace{-0.5em}
\subsection{Auto-adjust parameters}
\label{method:parameters} 
\vspace{-0.5em}

The additional parameters provided for tuning spatial and appearance fidelity substantially improve user control over the image harmonization process, but can pose difficultly for novice users to tune. 
Therefore, we introduce a heuristic-based algorithm that automatically generates parameters that qualitatively produce aesthetically pleasing images.
We discuss our parameter-setting algorithm in detail in the Appendix. 
\vspace{-0.5em}
\section{Evaluation}
\vspace{-0.5em}

\begin{figure}
    \includegraphics[width=1.0\linewidth,valign=m]{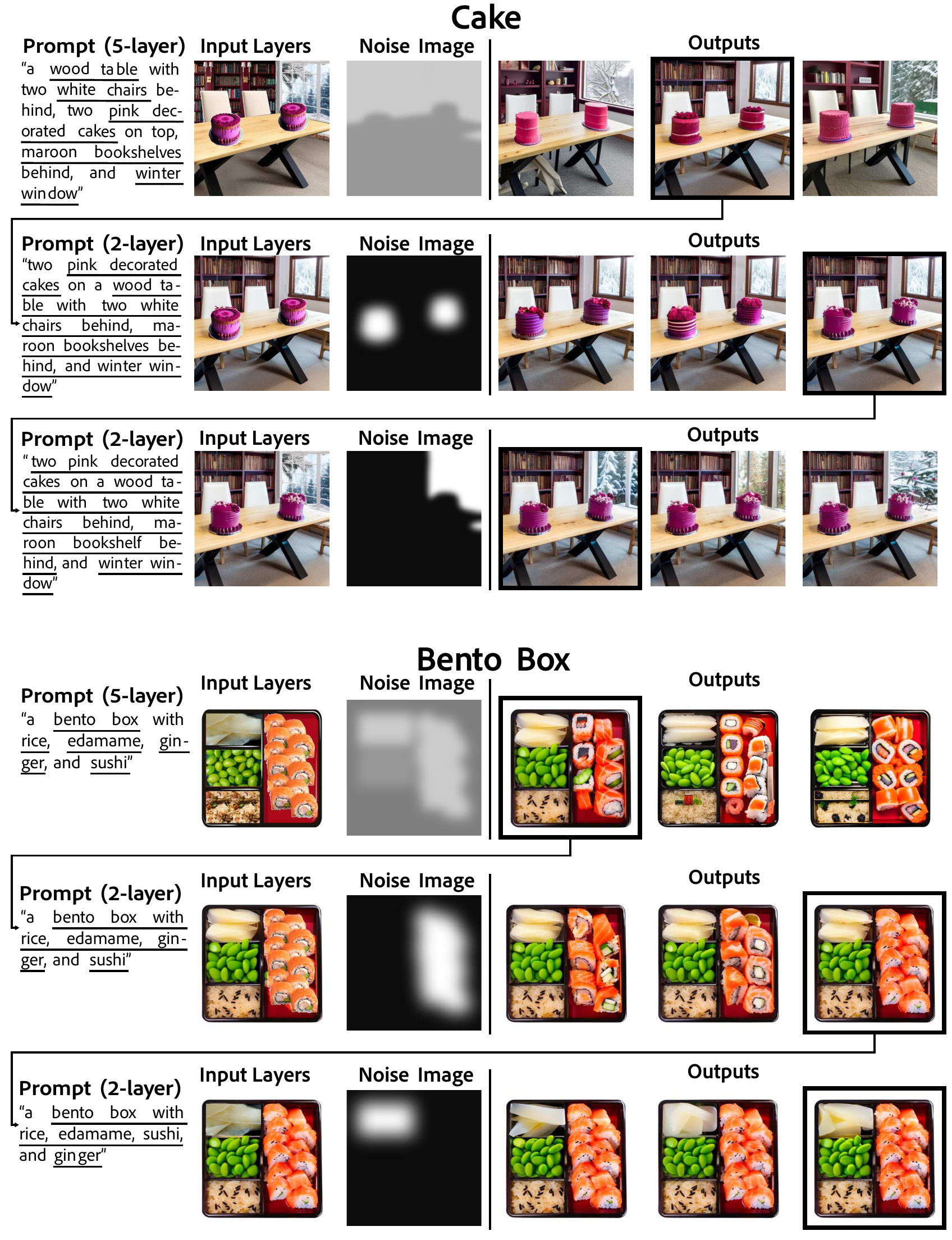}
    \caption{An iterative editing workflow where the user modifies individual layers of generated images for the \textbf{Cake} and \textbf{Bento Box} scenes. In each example, the user generates an initial image using \textit{Collage Diffusion}, then improves the images using two refinement iterations, re-generating one of the original input layers in each refinement iteration. }
    \label{fig:stepwise}
\end{figure}

\begin{figure}[!htbp]
    \centering
    \textbf{Toys}

    ``a \ul{teddy bear}, a \ul{wood train}, and an \ul{american football}, in front of a \ul{tan background}''
    \begin{adjustbox}{max size={\linewidth}{\textheight}}
        \begin{tabular}[t]{p{.32\linewidth}|p{.32\linewidth}p{.32\linewidth}p{.32\linewidth}p{.32\linewidth}p{.32\linewidth}}
        \hfil\textbf{Input Layers (4)} & \hfil\textbf{SA} & \hfil\textbf{GH} & \hfil\textbf{GH+CA} & \hfil\textbf{GH+CA+TI} & \hfil\textbf{GH+CA+TI+LN} \\
        \includegraphics[width=\linewidth,valign=m]{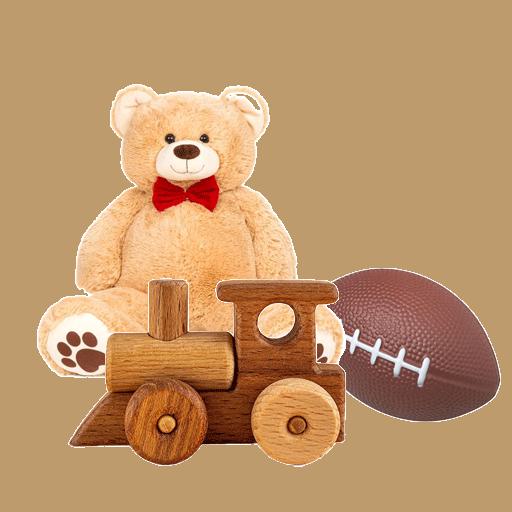}  & \includegraphics[width=\linewidth,valign=m]{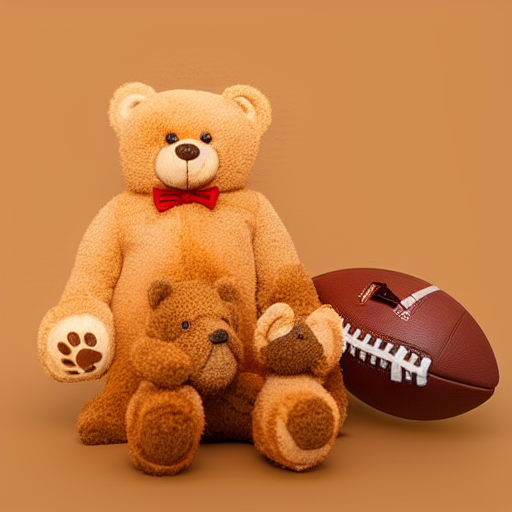} & \includegraphics[width=\linewidth,valign=m]{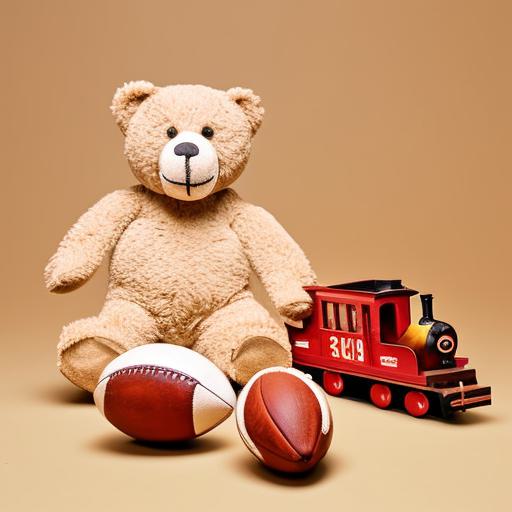} & \includegraphics[width=\linewidth,valign=m]{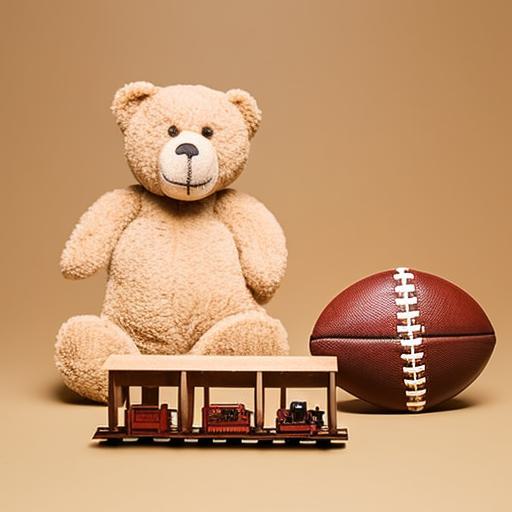} & \includegraphics[width=\linewidth,valign=m]{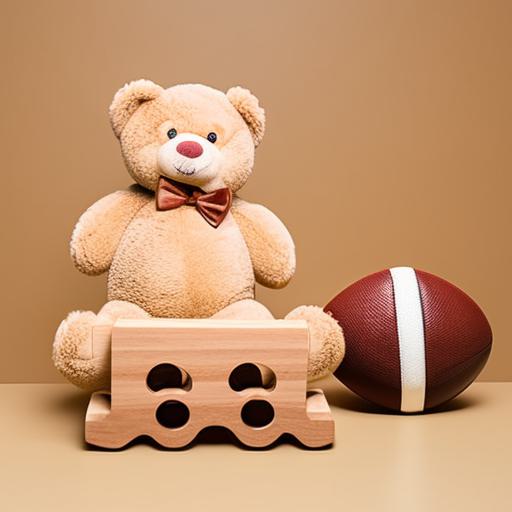} & \includegraphics[width=\linewidth,valign=m]{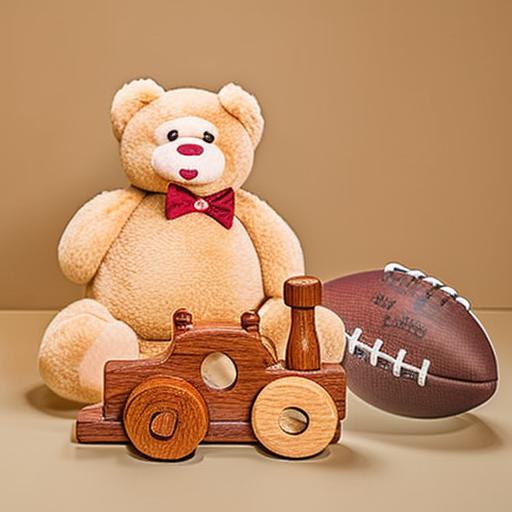} \\
        & Issues with harmonization on the football and merged teddy bears, no wood train in the bottom left & Harmonized image, no wood train in the bottom left & Wood train in the bottom left & Wood train with styling of wood closer to the starting image, white face and tie of teddy bear preserved & Wood train very similar to the original train, red color of tie preserved
    \end{tabular}
    \end{adjustbox}

    \textbf{Bento Box}

    ``a \ul{bento box} with \ul{rice}, \ul{edamame}, \ul{ginger}, and \ul{sushi}''
    \begin{adjustbox}{max size={\linewidth}{\textheight}}
        \begin{tabular}[t]{p{.32\linewidth}|p{.32\linewidth}p{.32\linewidth}p{.32\linewidth}p{.32\linewidth}p{.32\linewidth}}
            \hfil\textbf{Input Layers (5)} & \hfil\textbf{SA} & \hfil\textbf{GH} & \hfil\textbf{GH+CA} & \hfil\textbf{GH+CA+TI} & \hfil\textbf{GH+CA+TI+LN} \\
            \includegraphics[width=\linewidth,valign=m]{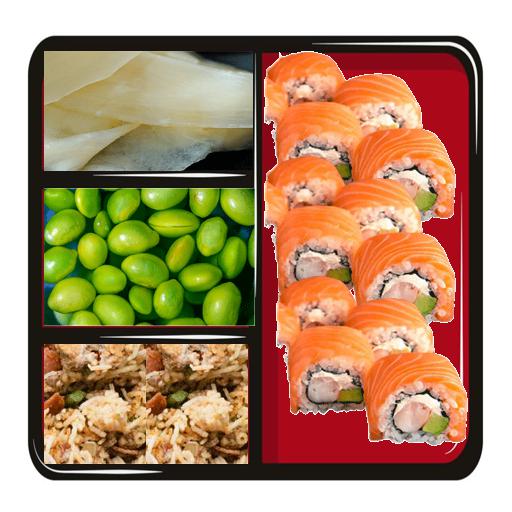} & \includegraphics[width=\linewidth,valign=m]{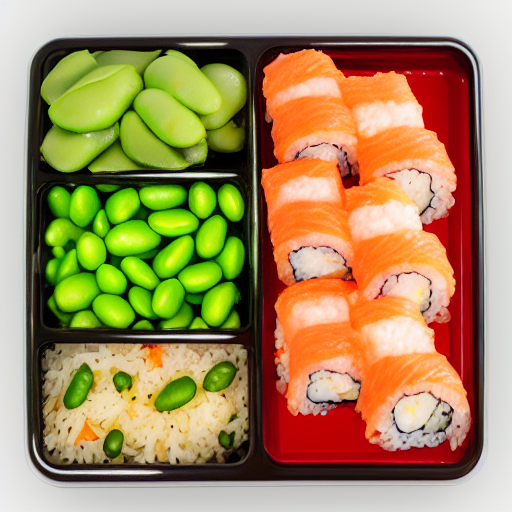} & \includegraphics[width=\linewidth,valign=m]{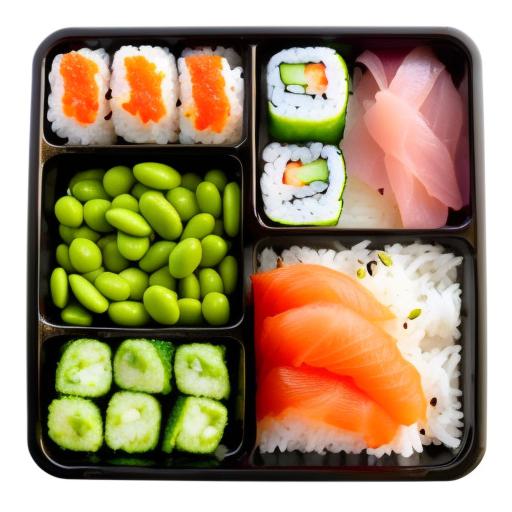} & \includegraphics[width=\linewidth,valign=m]{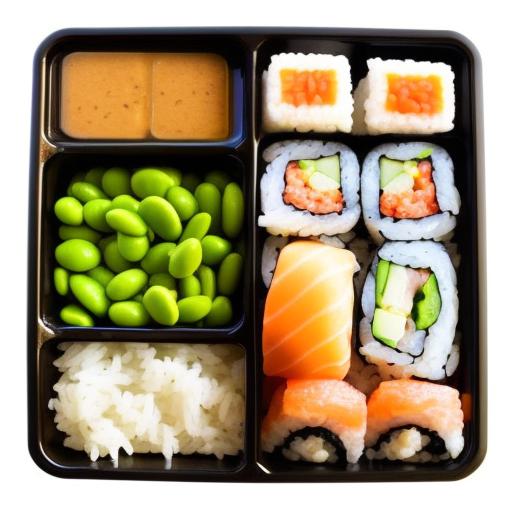} & \includegraphics[width=\linewidth,valign=m]{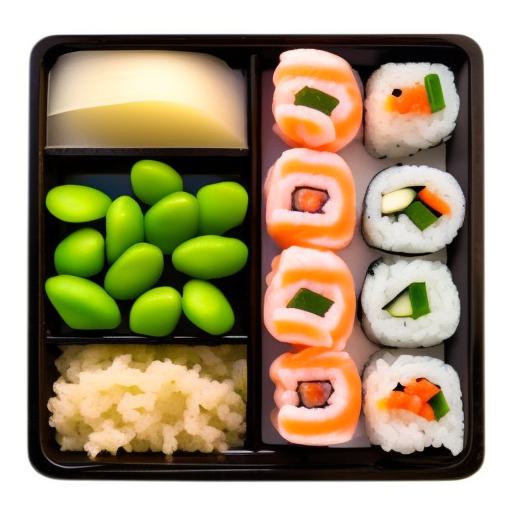} & \includegraphics[width=\linewidth,valign=m]{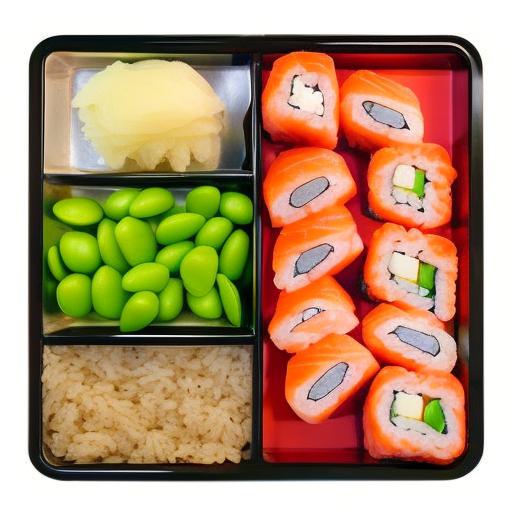} \\
            & Sushi orientation and shading not harmonized, edamame in place of ginger on the top left & Harmonized image, sushi in place of ginger in the top left, wasabi in place of rice in bottom left, no sushi in bottom right & Sushi and rice in currect locations, ginger paste instead of sliced sushi ginger in the top left & Sliced sushi ginger in the top left, darker rice in the bottom left, sushi on right more similar to layer & Sliced sushi ginger in the top left, dark rice in bottom left, sushi on right very similar to layer
        \end{tabular}
    \end{adjustbox}

    \textbf{Cake}

    ``a \ul{wood table} with two \ul{white chairs} behind, two \ul{pink decorated cakes} on top, \ul{maroon bookshelves} behind, and \ul{winter window}'' 
    \begin{adjustbox}{max size={\linewidth}{\textheight}}
        \begin{tabular}[t]{p{.32\linewidth}|p{.32\linewidth}p{.32\linewidth}p{.32\linewidth}p{.32\linewidth}p{.32\linewidth}}
            \hfil\textbf{Input Layers (5)} & \hfil\textbf{SA} & \hfil\textbf{GH} & \hfil\textbf{GH+CA} & \hfil\textbf{GH+CA+TI} & \hfil\textbf{GH+CA+TI+LN} \\
            \includegraphics[width=\linewidth,valign=m]{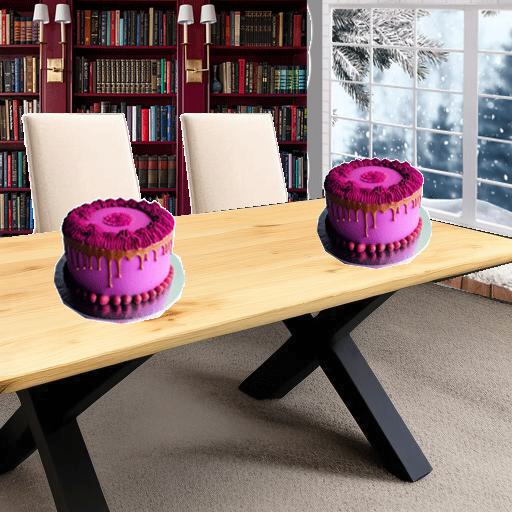} & \includegraphics[width=\linewidth,valign=m]{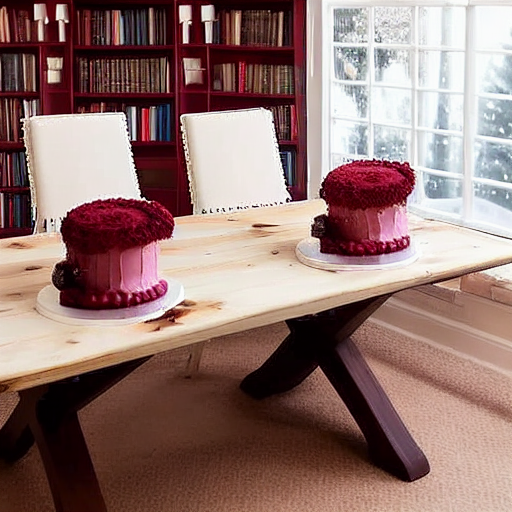} & \includegraphics[width=\linewidth,valign=m]{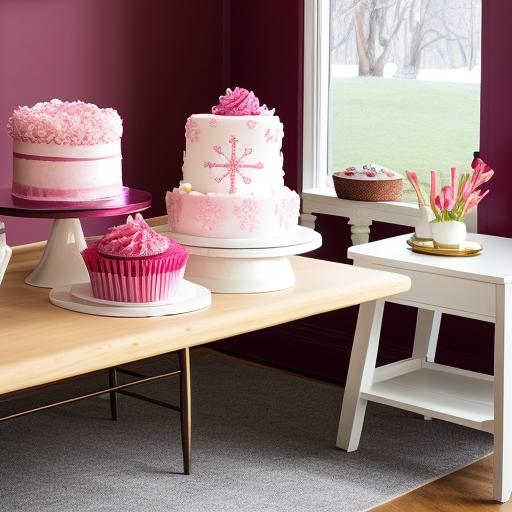} & \includegraphics[width=\linewidth,valign=m]{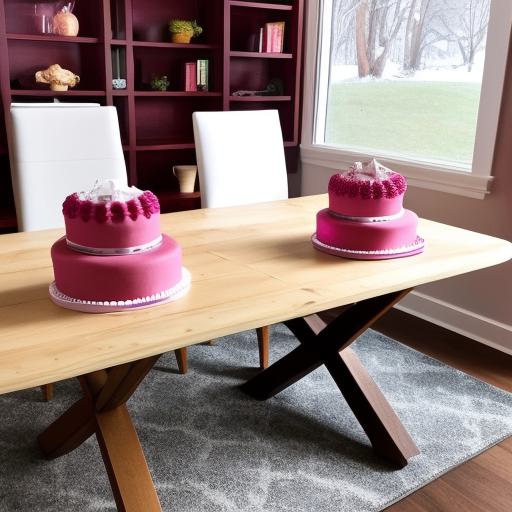} & \includegraphics[width=\linewidth,valign=m]{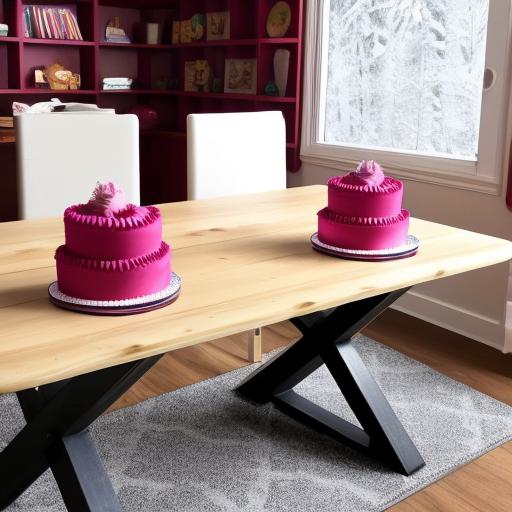} & \includegraphics[width=\linewidth,valign=m]{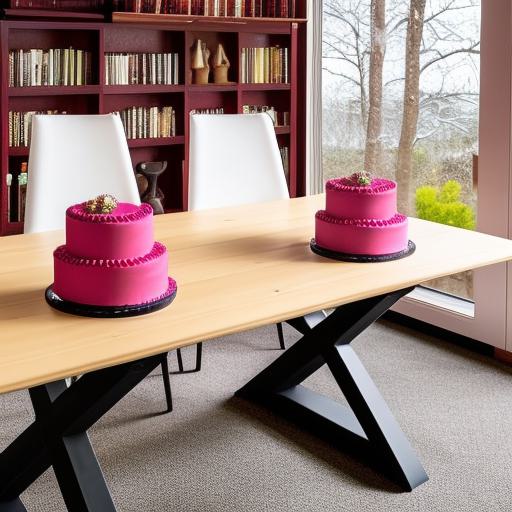} \\
            & Cake orientation not harmonized, bookshelf angle not harmonized, artifacts in the cakes, artifacts on the edges of the chairs & Harmonized image, white cakes in place of the chairs, no bookshelf & Brown table legs instead of black in the bottom right, chairs in the correct locations in top left, not many books on bookshelf in the top left, wooden floor around table & Black table leg in the bottom right, bookshelf with a few more books in the top left, wooden floor around table & Bookshelf with many books in the top left, carpet floor around the table
        \end{tabular}
    \end{adjustbox}
\caption{(Part 1) By leveraging layer information, \textit{Collage Diffusion} generates images with greater spatial and appearance fidelity than the baseline \textbf{GH} approach. For each scene above, there are several aspects in which \textbf{CA}, \textbf{TI}, and \textbf{LN} improve fidelity; we comment on some of these aspects in each row. Compared to \textbf{GH}, \textbf{SA} fails to effectively harmonize input layers; we comment on issues with harmonization in each row. }
\label{fig:mainResults1}
\end{figure}

\begin{figure}
    \centering

    \textbf{Veggie Face}

    ``a face made of vegetables, including a \ul{yellow bell pepper} and a \ul{green bell pepper}, a \ul{white cauliflower}, \ul{red potatoes}, \ul{baby corn}, \ul{small cucumber}, \ul{bean sprouts}, and \ul{floret broccoli}, on a \ul{grey background}''
    \begin{adjustbox}{max size={\linewidth}{\textheight}}
        \begin{tabular}[t]{p{.32\linewidth}|p{.32\linewidth}p{.32\linewidth}p{.32\linewidth}p{.32\linewidth}p{.32\linewidth}}
            \hfil\textbf{Input Layers (9)} & \hfil\textbf{SA} & \hfil\textbf{GH} & \hfil\textbf{GH+CA} & \hfil\textbf{GH+CA+TI} & \hfil\textbf{GH+CA+TI+LN} \\
            \includegraphics[width=\linewidth,valign=m]{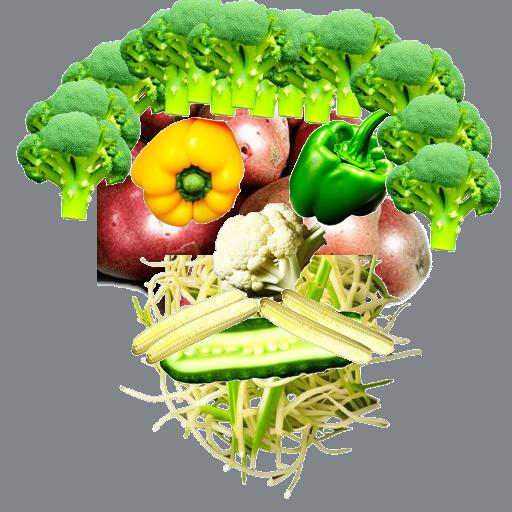} & \includegraphics[width=\linewidth,valign=m]{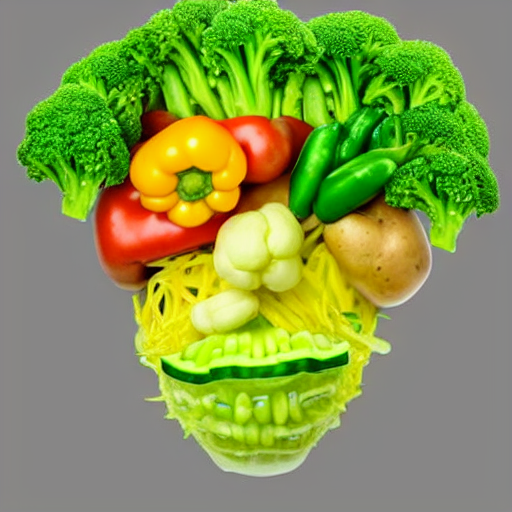} & \includegraphics[width=\linewidth,valign=m]{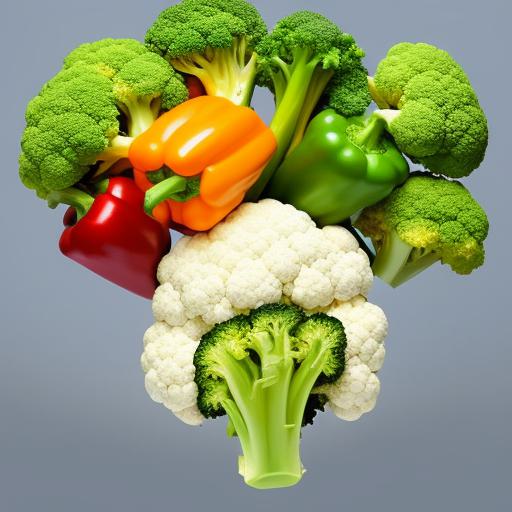} & \includegraphics[width=\linewidth,valign=m]{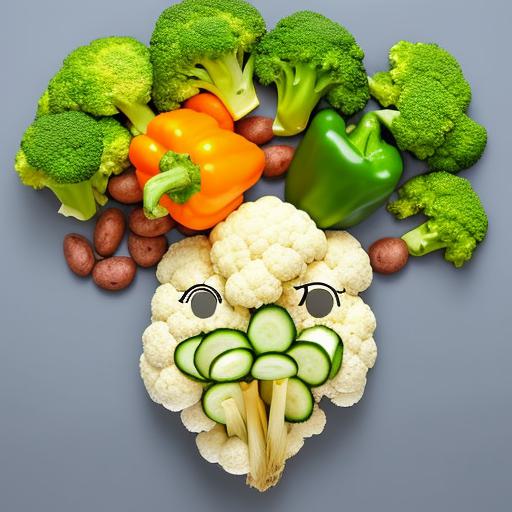} & \includegraphics[width=\linewidth,valign=m]{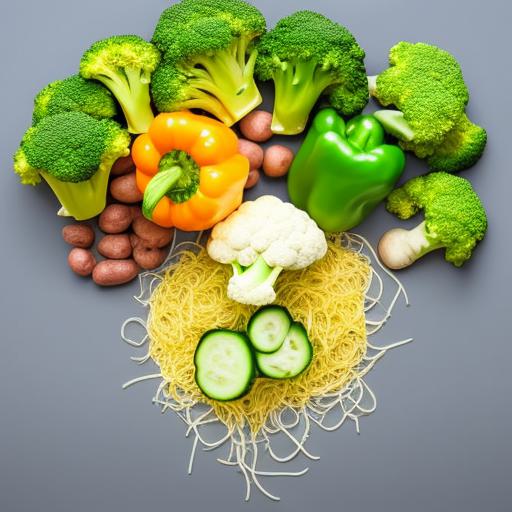} & \includegraphics[width=\linewidth,valign=m]{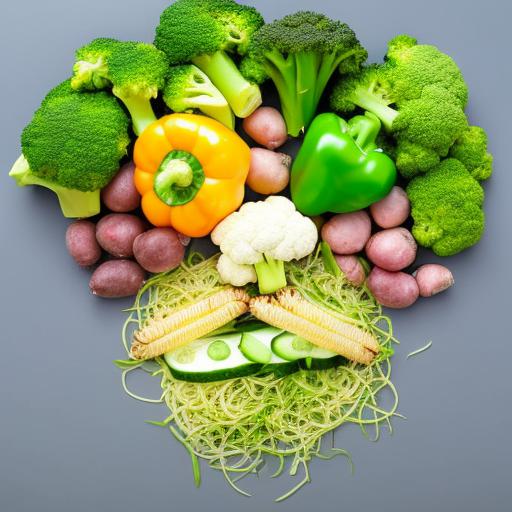} \\
            & Image missing shadows, some potatoes replaced with bell peppers, beans in chin, corn mustache missing & Harmonized image, missing cucumber mouth, corn mustache, sprout beard, red potatoes & Cucumber in correct location, but corn and most bean sprouts missing, red potatoes in correct location & Small cucumber slices for mouth, and some surrounding sprouts that look very different from the ones in the starting image & Bean sprouts more similar to starting image, sliced cucumber mouth more similar to layer, corn mustache in correct location and in natural orientation for a mustache
        \end{tabular}
    \end{adjustbox}

    \textbf{Striped Sweater}

    ``a man wearing \ul{green pants}, a \ul{blue and green striped sweater}, a \ul{plaid scarf}, and a \ul{maroon beanie}''
    \begin{adjustbox}{max size={\linewidth}{\textheight}}
        \begin{tabular}[t]{p{.32\linewidth}|p{.32\linewidth}p{.32\linewidth}p{.32\linewidth}p{.32\linewidth}p{.32\linewidth}}
            \hfil\textbf{Input Layers (4)} & \hfil\textbf{SA} & \hfil\textbf{GH} & \hfil\textbf{GH+CA} & \hfil\textbf{GH+CA+TI} & \hfil\textbf{GH+CA+TI+LN} \\
            \includegraphics[width=\linewidth,valign=m]{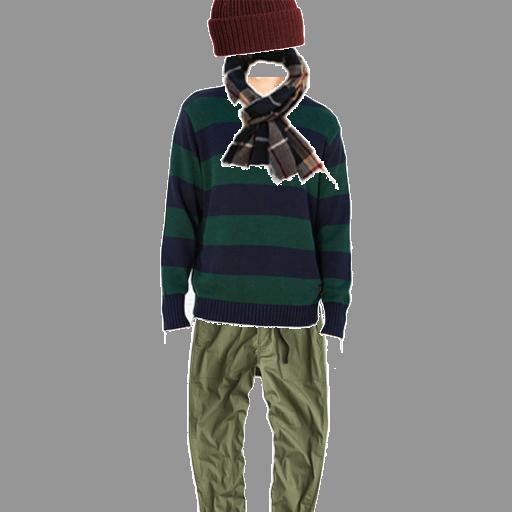} & \includegraphics[width=\linewidth,valign=m]{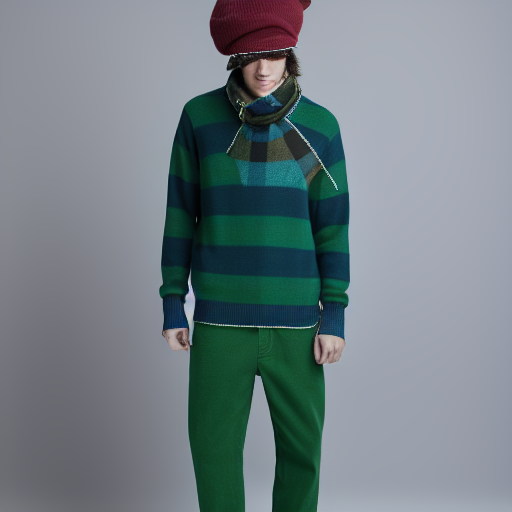} & \includegraphics[width=\linewidth,valign=m]{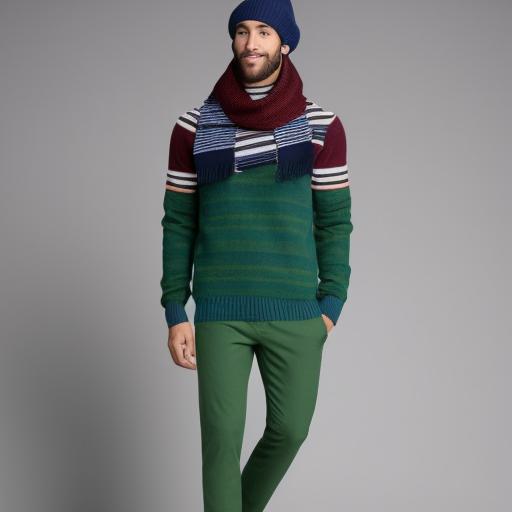} & \includegraphics[width=\linewidth,valign=m]{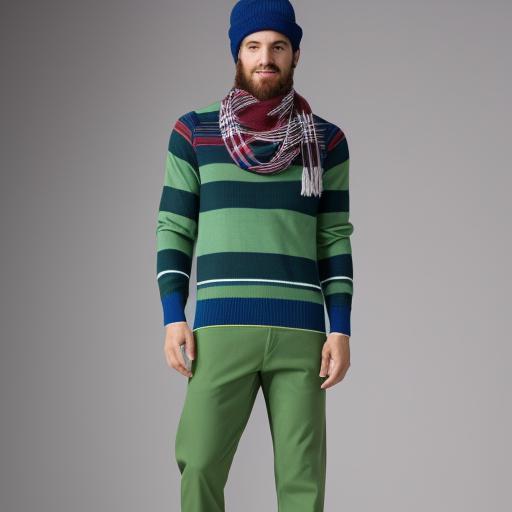} & \includegraphics[width=\linewidth,valign=m]{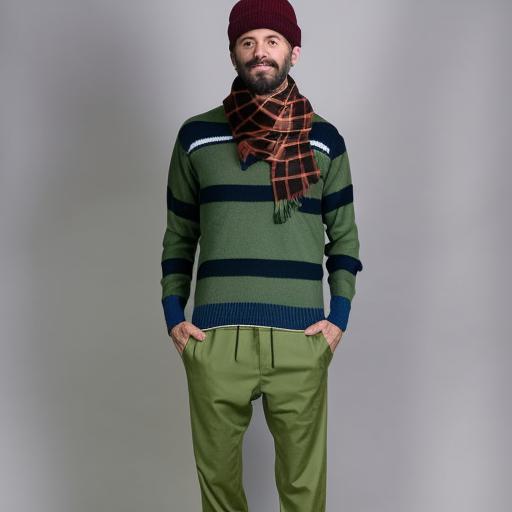} & \includegraphics[width=\linewidth,valign=m]{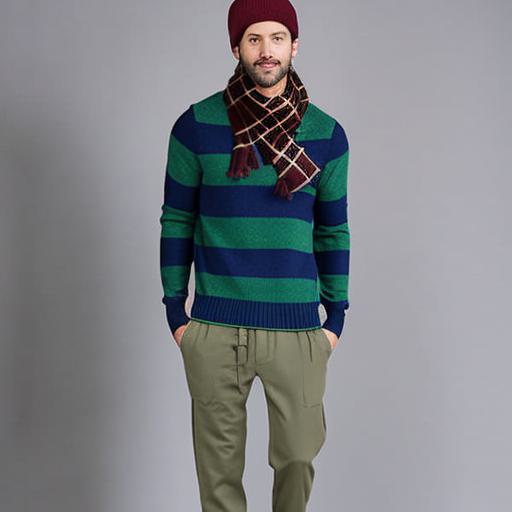} \\
            & Lack of harmonization in both the beanie on the head and the scarf mixing with the sweater, some artifacts from input layers preserved & Harmonized image, a green sweater missing dark stripes, scarf not plaid, blue beanie instead of maroon & A sweater striped with green and blue, plaid scarf, blue beanie instead of maroon & A sweater striped with green and blue that are closer to the original colors, plaid scarf with correct size of squares, pants closer to the style of the input & A sweater with very similar color and pattern to original
        \end{tabular}
    \end{adjustbox}

    \textbf{Ceramic Bowl}

    ``a \ul{blue ceramic bowl} with \ul{red potatoes}, \ul{red apples}, and \\ \ul{red bananas}''
    \begin{adjustbox}{max size={\linewidth}{\textheight}}
        \begin{tabular}[t]{p{.32\linewidth}|p{.32\linewidth}p{.32\linewidth}p{.32\linewidth}p{.32\linewidth}p{.32\linewidth}}
            \hfil\textbf{Input Layers (4)} & \hfil\textbf{SA} & \hfil\textbf{GH} & \hfil\textbf{GH+CA} & \hfil\textbf{GH+CA+TI} & \hfil\textbf{GH+CA+TI+LN} \\
            \includegraphics[width=\linewidth,valign=m]{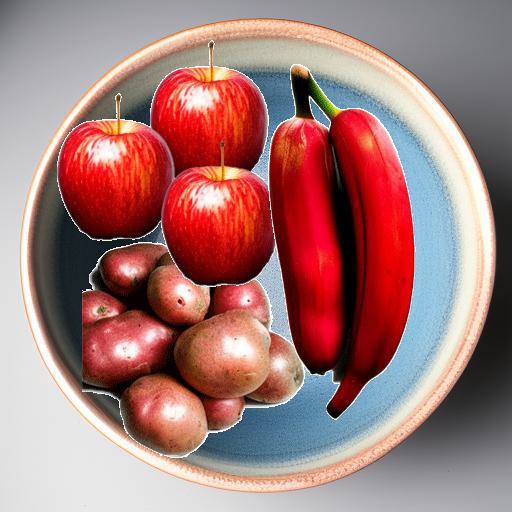} & \includegraphics[width=\linewidth,valign=m]{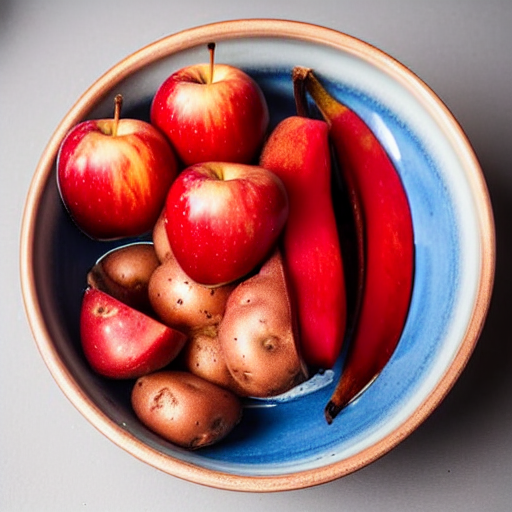} & \includegraphics[width=\linewidth,valign=m]{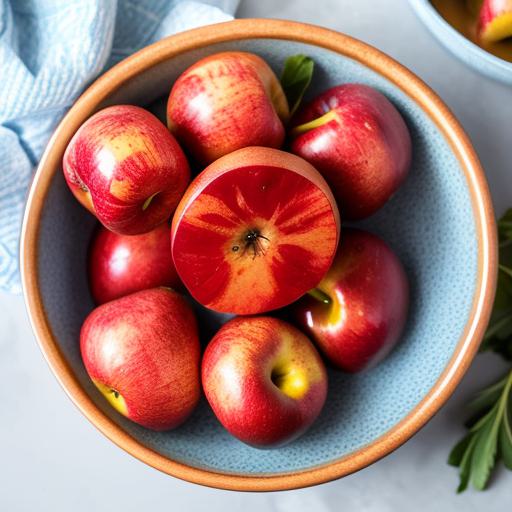} & \includegraphics[width=\linewidth,valign=m]{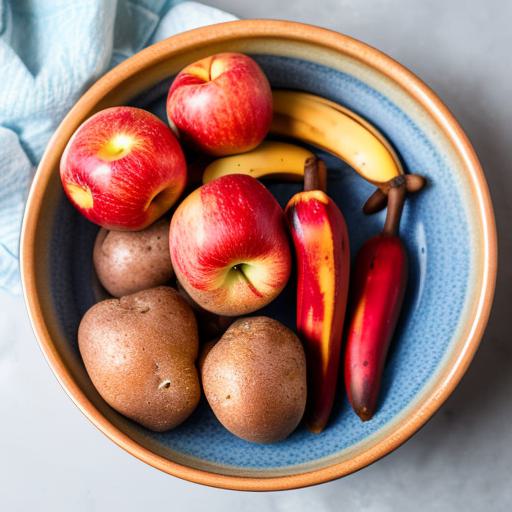} & \includegraphics[width=\linewidth,valign=m]{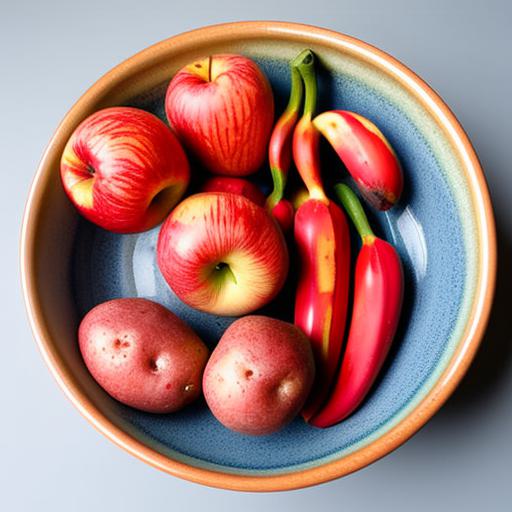} & \includegraphics[width=\linewidth,valign=m]{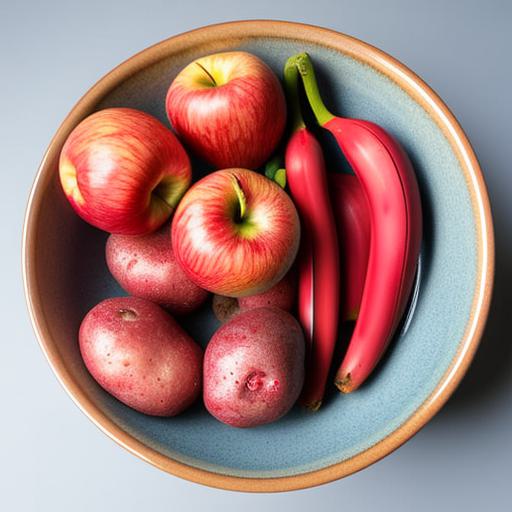} \\
            & Apple orientations not harmonized & Harmonized image, hybrid mixtures of apples, bananas, and potatoes throughout bowl & Brown potatoes instead of red in the bottom left, some yellow in the red bananas & All objects in the desired locations & Banana structure matching layer
        \end{tabular}
    \end{adjustbox}

    \textbf{Red Skirt}

    ``a person wearing a \ul{patterned red skirt}, \ul{buttoned blue blouse}, and \ul{pink summer coat}, in front of a \\ \ul{gray background}''
    \begin{adjustbox}{max size={\linewidth}{\textheight}}
        \begin{tabular}[t]{p{.32\linewidth}|p{.32\linewidth}p{.32\linewidth}p{.32\linewidth}p{.32\linewidth}p{.32\linewidth}}
            \hfil\textbf{Input Layers (4)} & \hfil\textbf{SA} & \hfil\textbf{GH} & \hfil\textbf{GH+CA} & \hfil\textbf{GH+CA+TI} & \hfil\textbf{GH+CA+TI+LN} \\
            \includegraphics[width=\linewidth,valign=m]{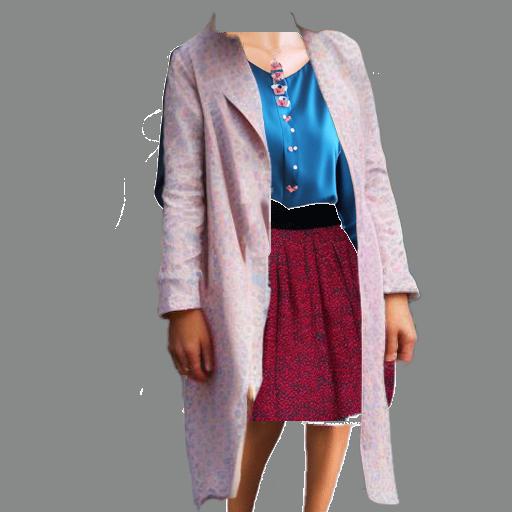} & \includegraphics[width=\linewidth,valign=m]{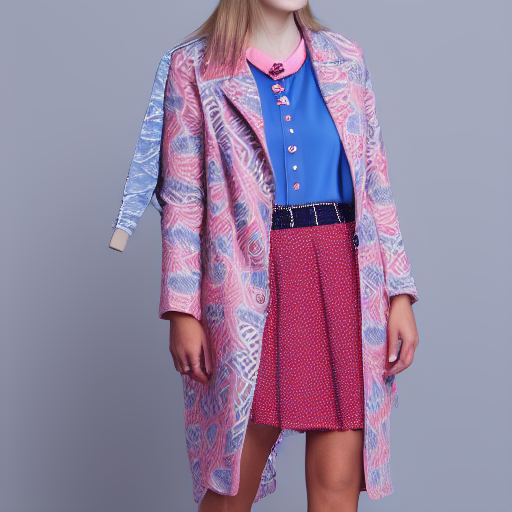} & \includegraphics[width=\linewidth,valign=m]{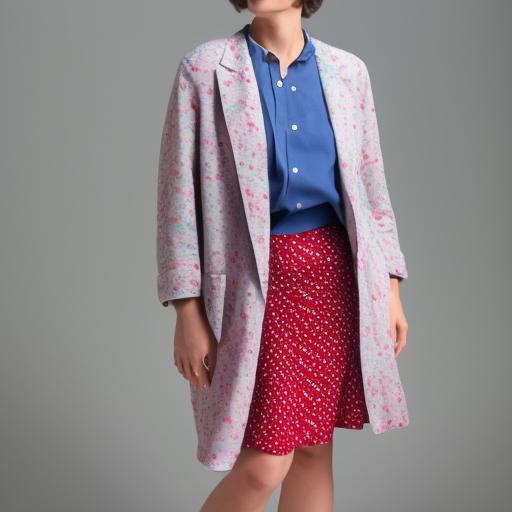} & \includegraphics[width=\linewidth,valign=m]{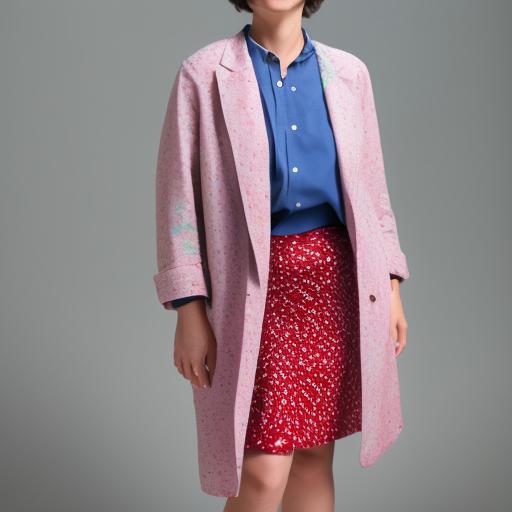} & \includegraphics[width=\linewidth,valign=m]{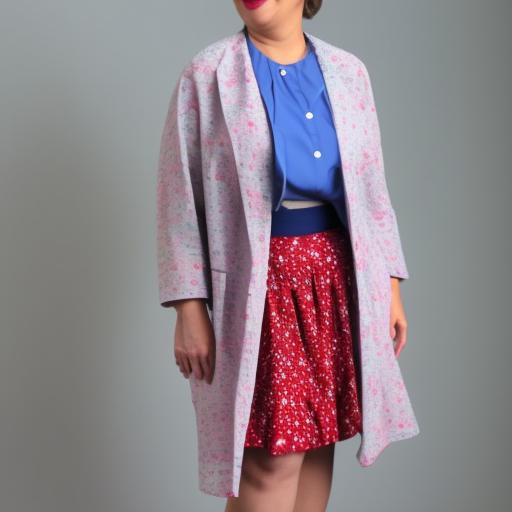}  & \includegraphics[width=\linewidth,valign=m]{claim_1/clothing3/img2img-with_cac-with_ft-no_mask/4.jpg} \\
            & Image artifact on the sleeve, all objects correctly mapped to the desired locations, layer image structure preserved & Harmonized image, all objects correctly mapped to the desired locations & No additional benefit from CA & TI introduces folds in the skirt & No further changes with \textbf{LN}
        \end{tabular}
    \end{adjustbox}

\caption{(Part 2) By leveraging layer information, \textit{Collage Diffusion} generates images with greater spatial and appearance fidelity than the baseline \textbf{GH} approach. See \cref{fig:mainResults1} caption for more detail. }
\label{fig:mainResults2}
\end{figure}

We primarily measure the value of layer information by analyzing the ability of \textit{Collage Diffusion} to generate globally harmonized images that respect the spatial layout and visual characteristics specified by input layers (the goals from \cref{sec:problem}). 
First, we evaluate \textit{Collage Diffusion} with a user in the loop, illustrating sample editing pipelines for multiple scenes.
We also analyze the value of layers for image harmonization in non-interactive settings, comparing \textit{Collage Diffusion} against multiple image harmonization approaches that take text prompt $c$ and composite image $x_c$ as input, flattening the layers rather than directly leveraging them. 
We additionally ablate the impact of the individual components of \textit{Collage Diffusion}. 
Finally, we explore the flexibility of layer-based controls by preserving the image structures of individual layers using ControlNet. 
We choose to focus on qualitative evaluation because our goals are primarily visual and because generative metrics for distributional comparison (FID, etc.) are not applicable in the layer-conditional setting where no ground-truth test dataset for ``the perfect output'' exists. 
Nevertheless, we also present a short quantitative study that mirrors our qualitative observations.

\vspace{-0.5em}
\subsection{Experimental Setup}
\vspace{-0.5em}

We evaluate the capacity of \textit{Collage Diffusion} to generate images without a user in the loop against two prior work baselines that do not use layer information. We also ablate \textit{Collage Diffusion} to create (1) a baseline that omits textual inversion but does modify cross-attention using layer information, and (2) a baseline that modifies cross-attention, leverages textual inversion, but does not enable per-layer control over harmonization.
We evaluate the performance of the following methods for a range of scenes: 
\begin{enumerate}%
    \item \textbf{SA}: Image generation with \textbf{S}elf-\textbf{A}ttention control via Plug-and-Play Diffusion \cite{tumanyan2022plug} applied to composite image $x_c$, with negative prompt ``A collage''. This is a baseline that does not leverage layer information, but maintains the image structure of $x_c$ via self-attention control.
    \item \textbf{GH}: \textbf{G}lobal \textbf{H}armonization by applying SDEdit \cite{sdedit} (\cref{method:SDEdit}) to composite image $x_c$. This is another baseline that does not leverage layer information. 
    \item \textbf{GH+CA}: \textbf{GH} with modified \textbf{C}ross-\textbf{A}ttention (\cref{method:CAC}). This builds upon \textbf{GH} by using layer information to improve spatial fidelity, but lacks specific mechanisms to improve appearance fidelity. 
    \item \textbf{GH+CA+TI}: \textbf{GH} applied to composite image $x_c$ with both \textbf{CA} learned per-layer representations via \textbf{T}extual \textbf{I}nversion \cite{TextualInversion} (\cref{method:TI}). This leverages layer information to improve both spatial and appearance fidelity.
    \item \textbf{GH+CA+TI+LN} (\textit{Collage Diffusion}): \textbf{GH} applied to composite image $x_c$ with both \textbf{CA} and \textbf{TI}, with per-\textbf{L}ayer \textbf{N}oise control (\cref{method:layerNoise}). This leverages layer information to improve both spatial and appearance fidelity, and allows user control over the harmonization-fidelity tradeoff on a per-layer basis.
\end{enumerate}
Controlled image-to-image techniques \cite{prompt2prompt,brooks2022instructpix2pix,tumanyan2022plug,mokady2022null} adhere too closely to starting image structure, as discussed in \cref{related:img2img}, resulting in performance worse than the \textbf{GH} baseline.
Here, we evaluate against one of these methods in \textbf{SA} \cite{tumanyan2022plug}; see the Appendix for additional discussion. 
\vspace{-0.5em}
\paragraph{Scene construction.}
We evaluate \textit{Collage Diffusion} on seven diverse scenes created using an interactive layer editor UI that provides controls similar to those in popular layer-based image editing software.
Creating a scene using the UI is simple and straightforward--see the Appendix for a video example.

\vspace{-0.5em}
\paragraph{Model and optimization.}
We use the Stable Diffusion \cite{latentDiffusion} 2.1 base model as $D_\theta$ for \textbf{GH}, \textbf{GH+CA}, \textbf{GH+CA+TI}, and \textbf{GH+CA+TI+LN}, and generate images using the Euler ancestral solver with 50 steps. 
For each scene, we tune the noise added to the image to qualitatively optimize the harmonization-fidelity tradeoff; values are between $t=0.7$ and $t=0.8$ for all scenes tested. 
We use the official PyTorch implementation of \textbf{SA} \cite{tumanyan2022plug}. 
\vspace{-0.5em}
\paragraph{Metrics} 
\vspace{-0.5em}
We use the following metrics for quantitative evaluation. Our spatial fidelity goals aim for layer text $c_i$ to match the visual content in $x^*_c$ in regions where layer $i$ is visible---we measure this by computing CLIP \cite{radford2021learning} text-image similarity between $c_i$ and the corresponding region of $x^*_c$. 
Appearance fidelity aims for layer image $x_i$ to match the visual content in $x^*_c$ where layer $i$ is visible---we measure this by computing CLIP image-image similarity between $x_i$ and the corresponding region of $x^*_c$. 
We include additional details on metrics in the Appendix.
\vspace{-0.5em}
\subsection{Interactive Editing} \label{exp:interactive}
\vspace{-0.5em}
We illustrate interactive editing with \textit{Collage Diffusion} by repeatedly (1) generating 10 images using different random seeds, (2) allowing the user to select the image they like the most, and (3) selecting an object in this image that they are unsatisfied with and would like to re-generate. This process continues until the user is satisfied with all aspects of the generated image. 

\cref{fig:stepwise} illustrates the value of \emph{Collage Diffusion} for interactively authoring complex scenes.
For the ``Cake'' scene, the user generates a final image in three steps: (1) generating an initial collection of images from the input layers, (2) exploring different options for the cake, and (3) exploring different options for the winter window. 
Similarly, for ``Bento Box,'' the user generates a final image in three steps: (1) generating an initial collection of images from the input layers, (2) exploring different options for the sushi, and (3) exploring different options for the ginger. 
We successfully preserve all previously-generated objects while providing a diverse set of options for each modified object that match the layer specifications. 
This interactive refinement procedure is valuable for ensuring that the user is satisfied with all parts of the generated image.
\vspace{-0.5em}
\subsection{Non-Interactive Generation} \label{exp:mainResults}
\vspace{-0.5em}
\begin{table}[]
    \begin{tabular}{l|llll}
              & \textbf{GH} & \textbf{GH+CA} & \textbf{GH+CA} & \textbf{GH+CA} \\ 
              & & & \textbf{+TI} & \textbf{+TI+LN} \\
    \hline
    $\uparrow$Txt-Img. Sim.& 0.215 & 0.236 & 0.233 & 0.238 \\
    $\uparrow$Img-Img. Sim.& 0.846 & 0.867 & 0.877 & 0.893 \\
    \end{tabular}
    \caption{\textbf{CA}, \textbf{TI}, and \textbf{LN} help \textit{Collage Diffusion} improve both spatial fidelity, as measured by per-layer text-image similarity with the input layers, and appearance fidelity, as measured by per-layer image-image similarity with the input layers. Metrics are averaged across 10 image seeds and all layers for seven scenes.}
    \label{table:clip}
\end{table}
\cref{fig:mainResults1} and \ref{fig:mainResults2} compare the visual output of \textit{Collage Diffusion} with the baseline methods. 
We did not cherry-pick the individual image seeds for each scene---additional examples from each test scene are included in the Appendix, and reflect the same overall trends. 
\vspace{-1em}
\paragraph{GH generates globally-harmonized images, while SA struggles with harmonization.}
Comparison of the \textbf{SA} and \textbf{GH} columns in \cref{fig:mainResults1} and \ref{fig:mainResults2} illustrates the capacity of \textbf{GH} to generate a harmonized image from input $x_c$ while highlighting the downsides of manipulating self-attention to preserve image structure in \textbf{SA}. When image harmonization requires altering the orientations of objects in the scene---the sushi in ``Bento Box,'' the cakes in ``Cake,'' the apples in ``Ceramic Bowl,'' etc.---\textbf{SA} fails to harmonize the image due to the constraints placed on the self-attention maps. In contrast, \textbf{GH} reliably generates globally-harmonized images: the images have consistent perspective and lighting, with fewer artifacts. 
\vspace{-1em}
\paragraph{CA consistently improves spatial fidelity across scenes.} 
Comparison of the \textbf{GH} and \textbf{GH+CA} columns in \cref{fig:mainResults1} and \ref{fig:mainResults2} illustrates the benefits of layer-based cross-attention control.
In ``Bento Box,'' using \textbf{CA} results in ginger and rice in the appropriate locations in the generated output. \textbf{CA} also helps preserve the table legs in ``Cake,'' maps the correct fruits to the correct parts of ``Ceramic Bowl,'' etc.
This trend is also reflected quantitatively: in \cref{table:clip}, \textbf{GH+CA} has a higher average per-layer text-image similarity than \textbf{GH}, indicating better spatial fidelity. 
\vspace{-1em}
\paragraph{TI consistently improves appearance fidelity across scenes.} Having mapped the desired concepts to the desired locations, comparison of the \textbf{GH+CA} and \textbf{GH+CA+TI} columns in \cref{fig:mainResults1} and \ref{fig:mainResults2} illustrates the benefits of layer-based textual representations. 
\textbf{TI} helps generate a wood train with similar style to the starting image in ``Toys,'' the right type of sushi ginger in ``Bento Box,'' the proper legs for the table in ``Cake,'' the correct color and shape for the potatoes in ``Ceramic Bowl,'' the proper saturation of colors and presence of wrinkles in ``Clothing,'' etc. 
This trend is also reflected quantitatively: in \cref{table:clip}, \textbf{GH+CA+TI} has a higher average per-layer image-image similarity than \textbf{GH+CA}, indicating better appearance fidelity. 
\vspace{-1em}
\paragraph{LN consistently helps optimize the harmonization-fidelity tradeoff across scenes}
Having mapped the desired concepts to the desired locations, with textual inversion to increase appearance fidelity, comparison of the \textbf{GH+C+TI} and \textbf{GH+CA+TI+LN} columns in \cref{fig:mainResults1} and \ref{fig:mainResults2} illustrates the benefits of control over per-layer noise.
\textbf{LN} increases the preservation of the structure of the wood train in ``Toys'', the salmon on the sushi in ``Bento Box'', the books on the bookshelves in ``Cake'', the shape of the bananas in ``Ceramic Bowl'', the stripes of the sweater in ``Striped Sweater'', the corn and cucumber in ``Veggie Face,'' etc. For all these scenes, the quality of image harmonization is maintained across \textbf{GH+C+TI} and \textbf{GH+CA+TI+LN}. 
This trend is also reflected quantitatively: in \cref{table:clip}, \textbf{GH+CA+TI+LN} has higher average per-layer text-image and image-image similarity than \textbf{GH+CA+TI}, indicating better spatial and appearance fidelity. 
\vspace{-1.0em}
\paragraph{Where is layer-driven harmonization most helpful?} 
To understand the situations where layer information is most valuable,
we highlight the ``Red Skirt'' (\cref{fig:mainResults2}) and ``Cake'' (\cref{fig:mainResults1}) scenes as examples at either end of the range of difficulty where layers are valuable. 
When harmonization requires limited changes to image structure, \textbf{SA} can be suitable---while \textbf{SA} still produces artifacts on ``Red Skirt'', the approach is more effective than on other scenes because fewer changes in image structure are required to harmonize the image. 
When objects are easy to discriminate even after noise is added (large objects with distinct colors), \textbf{GH} performs well, and \textbf{GH+CA} provides negligible added value.
If the visual attributes that the user cares to preserve in the layer are well-described by the layer prompt, \textbf{TI} may be unnecessary---in \cref{fig:mainResults2}, the only added benefit in ``Red Skirt'' comes from the preservation of the folds on the skirt and the dark band around the waist.

On the other end of the spectrum, when the user is particular on the \emph{exact} appearance of many complex layers, even \textit{Collage Diffusion} may struggle to satisfy user intent across all objects in the scene. For instance, in ``Cake,'' the user may want a specific color and icing pattern on the cake, a snowy pine outside the window, a full bookshelf, etc. For these situations, our iterative editing workflow is valuable, as highlighted in \cref{exp:interactive} and \cref{fig:stepwise}.
\vspace{-0.5em}
\subsection{Flexible per-layer controls with ControlNet} \label{exp:ControlNet}
\vspace{-0.5em}

\begin{figure}
\begin{adjustbox}{max size={\linewidth}{\textheight}}
    \begin{tabular}[t]{p{.32\linewidth}p{.32\linewidth}|p{.32\linewidth}|p{.32\linewidth}p{.32\linewidth}p{.32\linewidth}p{.32\linewidth}}
        \hfil\textbf{Prompt} & \hfil\textbf{Input Layers} & \hfil\textbf{Preserved} & & \hfil\textbf{Outputs} & \\ 
            & & \hfil\textbf{features} & & & \\ 
    \hline
    {\begin{tiny}A \ul{pirate ship} moving across a \ul{stormy ocean with waves} colliding into a \ul{rocky shore} containing a \ul{lighthouse} on top, \ul{dark storm clouds} with lightning in the background\end{tiny}}& \includegraphics[width=\linewidth,valign=t]{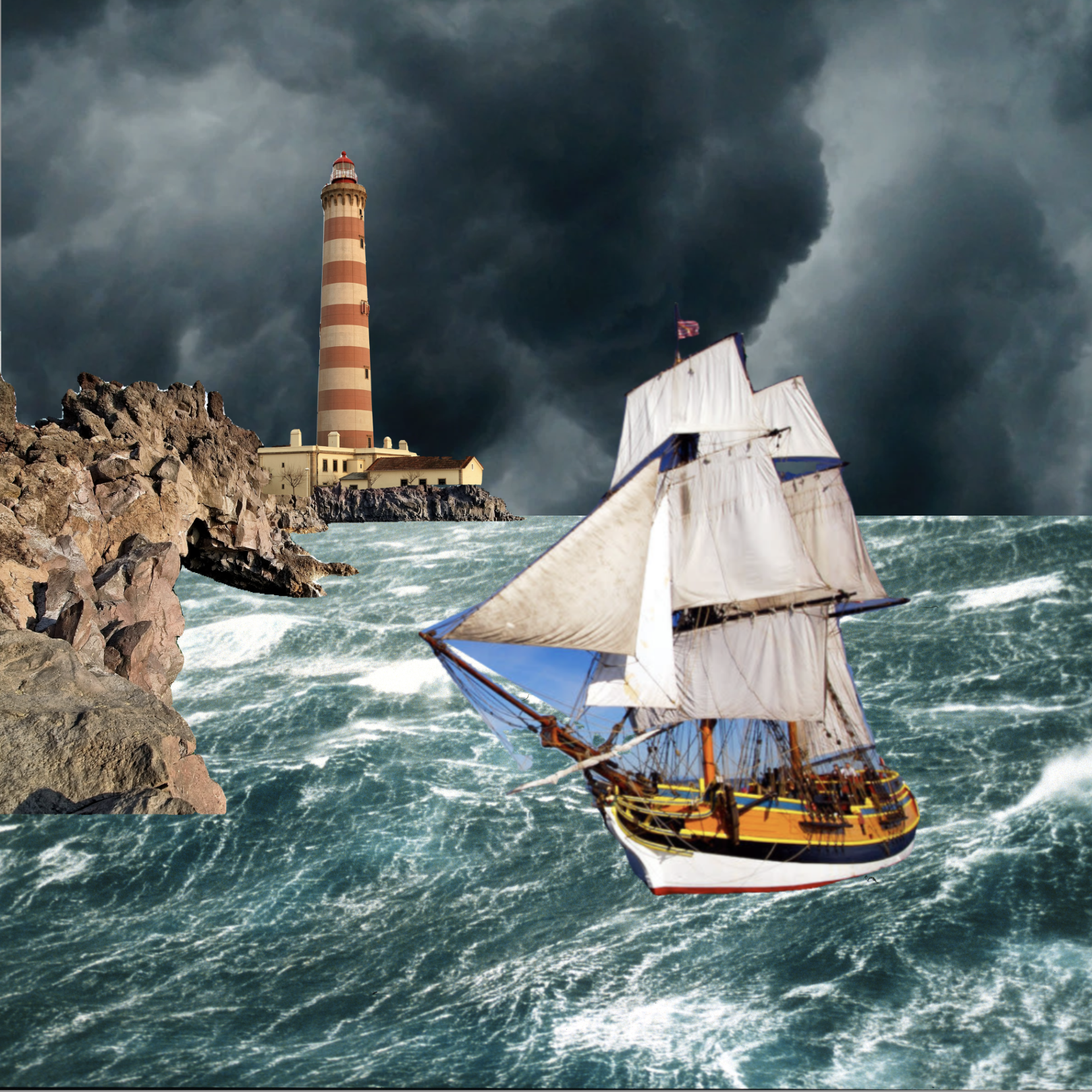}
    & Preserve edges: ship, rocks, lighthouse & \includegraphics[width=\linewidth,valign=t]{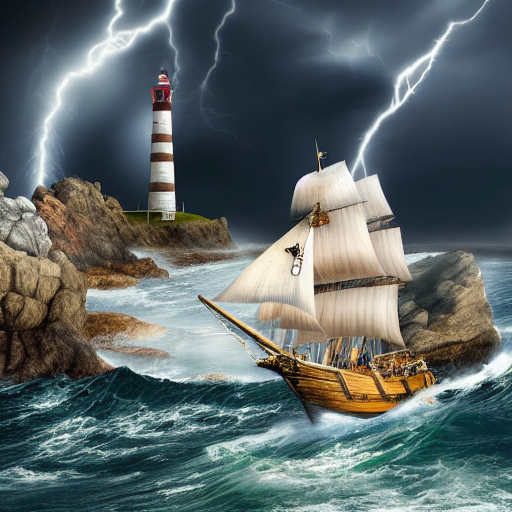} & \includegraphics[width=\linewidth,valign=t]{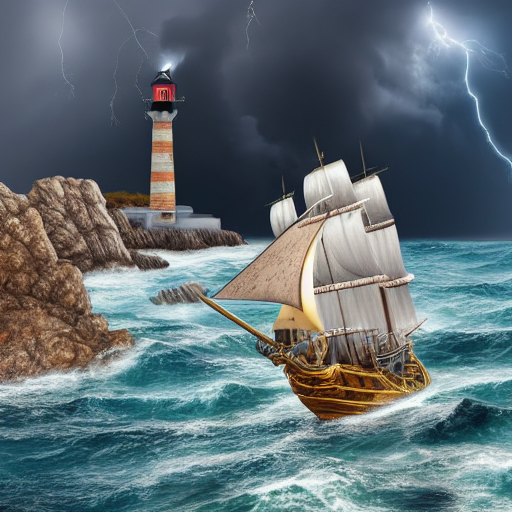} & \includegraphics[width=\linewidth,valign=t]{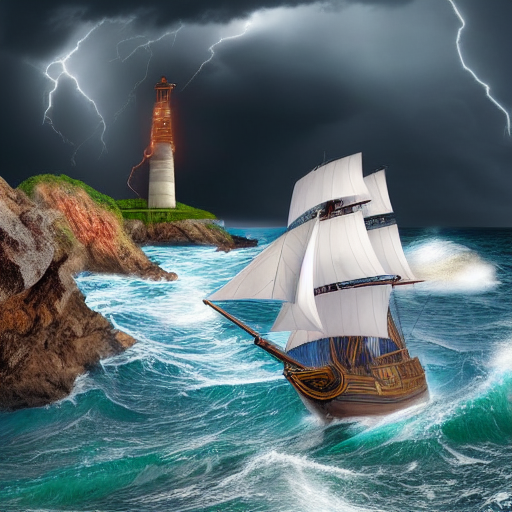} \\
    \begin{tiny}A \ul{house} with a \ul{pink cherry blossom} next to a \ul{swimming pool} with a stone pool deck in the \ul{backyard}, \ul{sky with birds flying} in the background\end{tiny} & \includegraphics[width=\linewidth,valign=t]{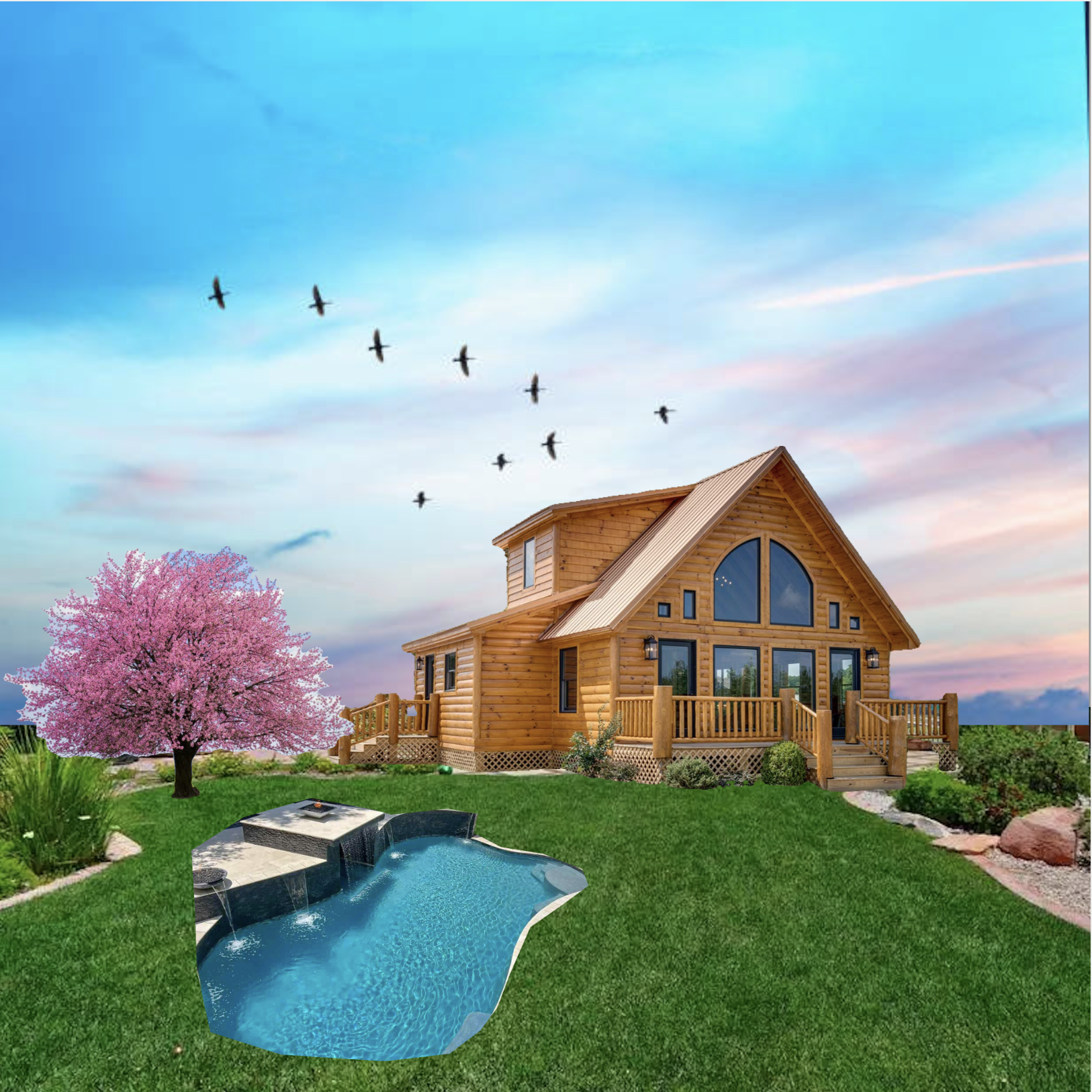}
    & Preserve edges: house, backyard & \includegraphics[width=\linewidth,valign=t]{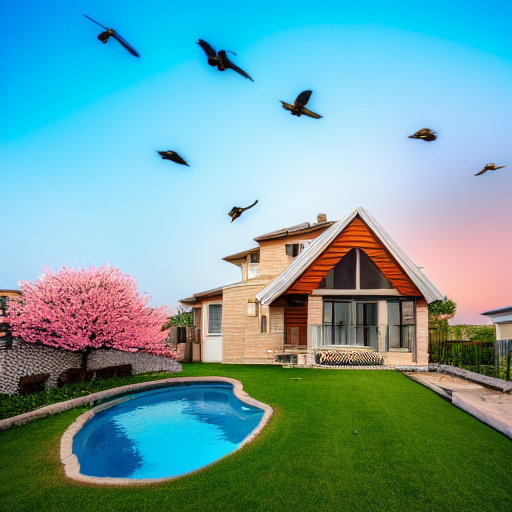} & \includegraphics[width=\linewidth,valign=t]{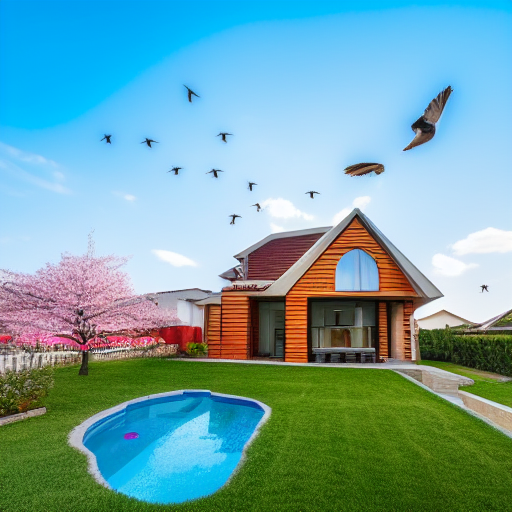} & \includegraphics[width=\linewidth,valign=t]{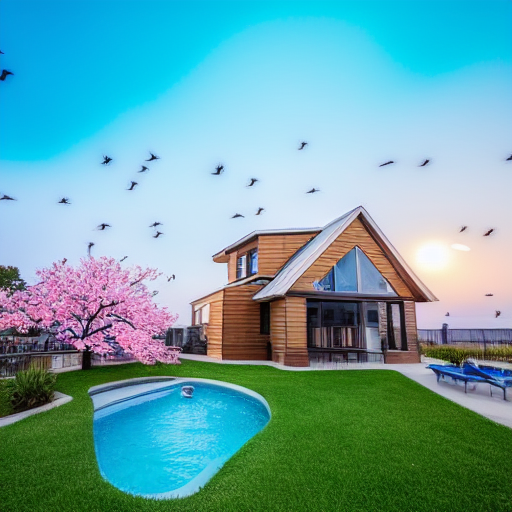} \\
    \end{tabular}
\end{adjustbox}
\caption{ControlNet lets users preserve edge maps on a per-layer basis. First row: high ControlNet weights preserve edge maps for the ships, rocks, and lighthouse. Second row: high ControlNet weights preserve edge maps for the house and the backyard.}
\label{fig:controlnet}
\end{figure}

In \cref{fig:controlnet}, our ControlNet extension enables users to preserve image structures on a per-layer basis. In the first row, high ControlNet weights preserve the edge maps of the ships, rocks, and lighthouse (note that the colors/textures of the rocks and lighthouse vary). The generated images have more variation in the ocean and sky. In the second row, high ControlNet weights strictly preserve the structure of the house, while loosely preserving the layout of the backyard, and allowing variation in the pool shape and pattern of birds in the sky.

\vspace{-1em}
\section{Conclusion}
\vspace{-0.5em}

Given a sequence of layers, \textit{Collage Diffusion} harmonizes these layers while preserving a user's desired spatial layout as well as the desired visual characteristics of the individual objects. 
Layers enable fine-grained control over how various properties of individual objects are preserved during the harmonization process. 
As a result, users can construct a scene matching their precise specifications and edit individual objects in the generated images. 

\paragraph{Acknowledgements} 
We gratefully acknowledge the support of NIH under No. U54EB020405 (Mobilize), NSF under Nos. CCF1763315 (Beyond Sparsity), CCF1563078 (Volume to Velocity), and 1937301 (RTML); US DEVCOM ARL under No. W911NF-21-2-0251 (Interactive Human-AI Teaming); ONR under No. N000141712266 (Unifying Weak Supervision); ONR N00014-20-1-2480: Understanding and Applying Non-Euclidean Geometry in Machine Learning; N000142012275 (NEPTUNE); NXP, Xilinx, LETI-CEA, Intel, IBM, Microsoft, NEC, Toshiba, TSMC, ARM, Hitachi, BASF, Accenture, Ericsson, Qualcomm, Analog Devices, Google Cloud, Salesforce, Total, the HAI-GCP Cloud Credits for Research program, the Stanford Data Science Initiative (SDSI), the Department of Defense National Defense Science and Engineering Graduate Fellowship, and members of the Stanford DAWN project: Facebook, Google, and VMWare. The U.S. Government is authorized to reproduce and distribute reprints for Governmental purposes notwithstanding any copyright notation thereon. Any opinions, findings, and conclusions or recommendations expressed in this material are those of the authors and do not necessarily reflect the views, policies, or endorsements, either expressed or implied, of NIH, ONR, or the U.S. Government.

{\small
\bibliographystyle{ieee_fullname}
\bibliography{collage-diffusion}
}

\pagebreak
\onecolumn
\setcounter{section}{0}
\section{Quantitative Metrics}

Here, we describe the per-layer text-image and image-image similarity metrics in greater detail.

Per-layer text-image similarity aims to measure spatial fidelity, which is defined as having the objects described by layer text $c_i$ matching the correct regions of $x^*_c$. CLIP \cite{radford2021learning} contrastively learns text and image embeddings such that the similarity of two strings or images can be measured by the cosine similarity between embedding vectors for the two concepts. We multiply $x^*_c$ with layer visibility mask $m_i$, where $m_i$ is $1$ where the layer $l_i$ should be visible in the generated image and $0$ otherwise (the same visibility computation as in Section \ref{method:CAC}), to generate a new image $x^*_c \odot m_i = x^*_i$. We compute the normalized CLIP text embedding of $c_i$, the normalized CLIP image embedding of $x^*_i$, and compute the cosine similarity of the two vectors as a proxy for spatial fidelity. 

Per-layer image-image similarity aims to measure appearance fidelity, which is defined as having the objects shown by layer image $x_i$ sharing visual characteristics of the corresponding region in $x^*_c$. We compute the normalized CLIP image embedding of $x_i$, the normalized CLIP image embedding of $x^*_i$, and compute the cosine similarity of the two vectors as a proxy for appearance fidelity. 

\section{Qualitative Metrics}
\label{appendix:qualitative}

We also evaluate collage quality by generating evaluation rubrics to qualitatively measure adherence to the collage diffusion goals specified in section \ref{sec:problem}, measuring performance along the following three axes: 
\begin{enumerate}
    \item \textbf{Image quality}: have we produced a ``high-quality'', globally-coherent image? (0 for No/1 for Yes)
    \item \textbf{Spatial fidelity}: for each desired object, have we correctly generated it in the desired position in the image? (0 for No/1 for Yes)
    \item \textbf{Appearance fidelity}: for each desired object, how closely do its visual features match the visual features of the original image? Note that appearance fidelity requires spatial fidelity as a prerequisite (Construct a list of visual attributes, score between 0.0 and 1.0 per attribute)
\end{enumerate}
Scores are computed for each method on a given scene by using this rubric to ``grade'' images generated from a large range of random seeds. 
We obtained 5 human evaluations for each of 10 image seeds for six scenes (``Toys'', ``Bento Box'', ``Cake'', ``Veggie Face'', ``Striped Sweater'', ``Ceramic Bowl'') across each of four methods (\textbf{GH}, \textbf{GH+CA}, \textbf{GH+CA+TI}, \textbf{GH+CA+TI+LN}). We exclude \textbf{SA} from the qualitative evaluation because DDIM inversion is determinisic, so we cannot compute averaged scores across many seeds. 

\begin{table}[!htbp]
    \centering
    \begin{tabular}{l|llll}
    & \textbf{GH} & \textbf{GH+CA} & \textbf{GH+CA+TI} & \textbf{GH+CA+TI+LN} \\ 
    \hline
    Global Harmonization & 0.93 & 0.90                 & 0.87                 & 0.88                 \\
    Spatial Fidelity & 0.77 & 0.83                 & 0.82                 & 0.93                 \\
    Appearance Fidelity & 0.24 & 0.41                 & 0.51                 & 0.77                
    \end{tabular}
    \caption{Averaged qualitative rubric evaluations highlight how \textbf{CA} improves spatial fidelity, \textbf{TI} improves appearance fidelity, and \textbf{LN} improves both spatial and appearance fidelity, all with minimal loss in harmonization. This table presents the averaged rubric results from 5 human evaluators on 10 image seeds for each of 6 seeds, as described in Section \ref{appendix:qualitative}.}
    \label{table:survey}
\end{table}

The averaged results per method are presented in Table \ref{table:survey}. \textbf{CA} improves spatial fidelity, with an increase in average score from $0.77$ to $0.83$. \textbf{TI} then improves appearance fidelity, increasing the score from $0.41$ to $0.51$. Finally, tuning the harmonization-fidelity tradeoff on a per-layer basis with \textbf{LN} boosts both spatial fidelity ($0.82$ to $0.93$) and appearance fidelity ($0.51$ to $0.77$). Compared to \textbf{GH}, the full \emph{Collage Diffusion} methodology of \textbf{GH+CA+TI+LN} does slightly decrease the harmonization score from $0.93$ to $0.88$; however, nearly all generated images are still well-harmonized, and in exchange we significantly boost spatial fidelity by $0.16$ and appearance fidelity by $0.53$. 

\section{Collage-Conditional Diffusion as Image-To-Image Translation}

\begin{figure}[!htbp]
    \centering
    \begin{adjustbox}{max size={\textwidth}{\textheight}}
    \begin{tabular}[t]{p{.33\linewidth}p{.33\linewidth}p{.33\linewidth}}
        \hfil\textbf{Input Image} & \hfil\textbf{InstructPix2Pix} & \hfil\textbf{PnP Diffusion} \\
        \includegraphics[width=\linewidth,valign=m]{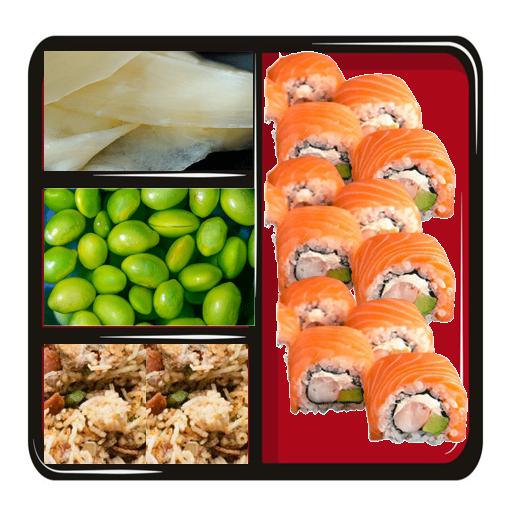} & \includegraphics[width=\linewidth,valign=m]{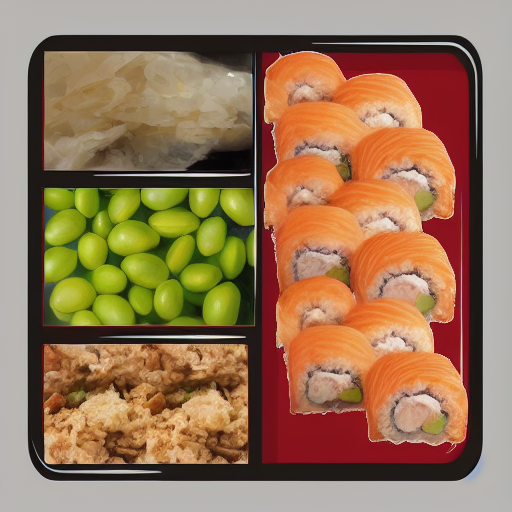} & \includegraphics[width=\linewidth,valign=m]{claim_1/bentoBox/pnp.jpg} \\
        Prompt: ``a bento box with rice, edamame, ginger, and sushi'' & Prompt: ``change style from collage to realistic'' & Prompt: ``a bento box with rice, edamame, ginger, and sushi'' \newline Negative Prompt: ``a collage'' \\
    \end{tabular}
    \end{adjustbox}
    \caption{Image-to-image methods that aim to preserve structure are ineffective at collage-conditional diffusion. }
    \label{fig:PnP}
\end{figure}

As mentioned in Section \ref{related:img2img}, it is possible to frame collage-conditional diffusion as a controlled image-to-image task---manipulating individual objects or the overall style of an image while keeping the image structure as fixed as possible. 
Given access to these methods, is it even necessary to leverage individual layer information for collage-conditional diffusion? 
Testing both InstructPix2Pix \cite{brooks2022instructpix2pix} and Plug-and-Play Diffusion \cite{tumanyan2022plug} (the \textbf{SA} method in Section \ref{exp:mainResults}), Figure \ref{fig:PnP} highlights how both methods have a minimal impact in terms of harmonizing the input bento box image---the sushi still aren't oriented in a way that fits the bento box, etc.---and Plug-and-Play Diffusion accidentally removes both the ginger and parts of the sushi from the image. 

\begin{figure}[!htbp]
    \centering
    \begin{adjustbox}{max size={\textwidth}{\textheight}}
    \begin{tabular}[t]{p{.33\linewidth}p{.33\linewidth}p{.33\linewidth}}
        \hfil\textbf{Input Image} & \hfil\textbf{InstructPix2Pix} & \hfil\textbf{PnP Diffusion} \\
        \includegraphics[width=\linewidth,valign=m]{claim_1/bentoBox/init.jpg} & \includegraphics[width=\linewidth,valign=m]{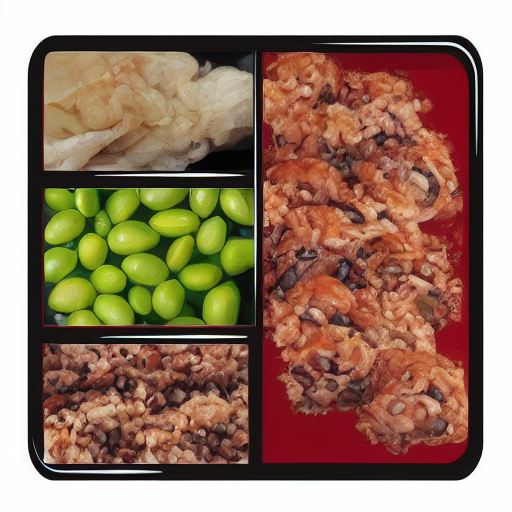} & \includegraphics[width=\linewidth,valign=m]{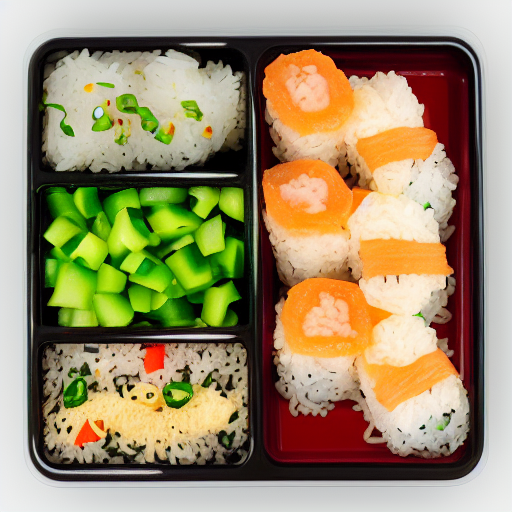} \\
        Prompt: ``a bento box with rice, edamame, ginger, and sushi'' & Prompt: ``replace edamame with black beans'' & Prompt: ``a bento box with rice, black beans, ginger, and sushi'' \newline Negative Prompt: ``a collage'' \\
    \end{tabular}
    \end{adjustbox}
    \caption{Image-to-image methods that aim to preserve structure struggle to handle prompts with many objects. }
    \label{fig:PnPCA}
\end{figure}

We also test the capacity of InstructPix2Pix and Plug-and-Play Diffusion to map complex compositional prompts to the appropriate regions of the image by attempting to replace the edamame in the bento box with black beans.
Figure \ref{fig:PnPCA} highlights the failure of both techniques for the task---InstructPix2Pix replaces the the rice and parts of the sushi with a rice-bean hybrid, while Plug-and-Play Diffusion turns the edamame into a green chopped vegetable while turning the ginger into rice. 

\section{Note on Iterative Inpainting}

\begin{figure}[!htbp]
    \centering
    \begin{adjustbox}{max size={\textwidth}{\textheight}}
    \begin{tabular}[t]{p{.4\linewidth}p{.4\linewidth}}
        \hfil\textbf{Input Layers (5)} & \hfil\textbf{Paint By Example} \\
        \includegraphics[width=\linewidth,valign=m]{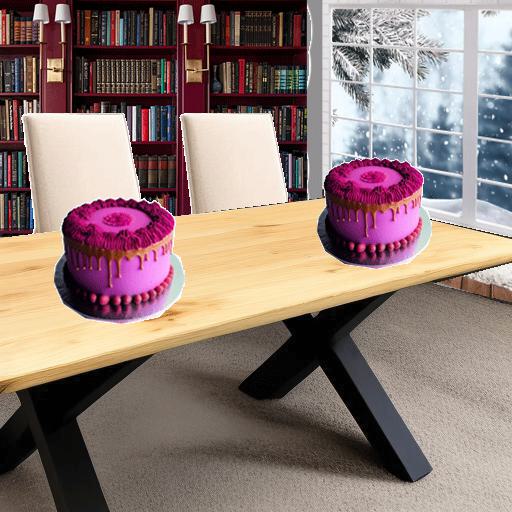} & \includegraphics[width=\linewidth,valign=m]{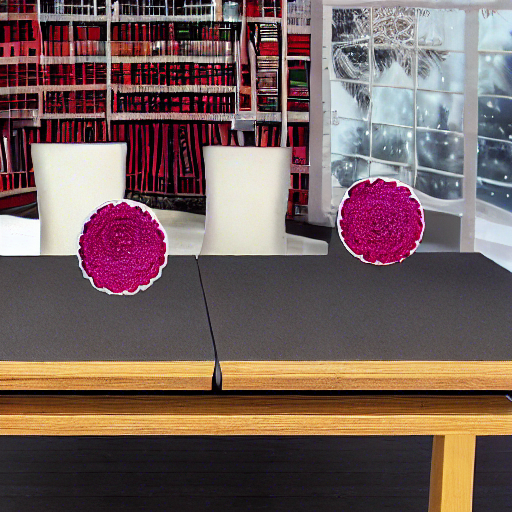} \\
    \end{tabular}
    \end{adjustbox}
    \vspace{0.5em}
    \caption{Iterative inpainting-based algorithms struggle with harmonizing many input layers, as the orientations and lighting of the generated objects don't quite fit together. }
    \vspace{-0.5em}
    \label{fig:byExample}
\end{figure}

Iterative inpainting-based algorithms such as ``Paint by Example''\cite{yang2023paint} have spatial fidelity due to the provided inpainting masks, and can have appearance fidelity to the input layers, but struggle with harmonizing many input layers; In \cref{fig:byExample}, the orientation and lighting of the cakes, chairs, table, etc. do not fit together. On the other hand, in \emph{Collage Diffusion}, SDEdit-style denoising helps enforce global harmonization not provided by iterative inpainting approaches.

\section{Additional Experimental Details}

We leverage the following negative prompts by scene:
\begin{itemize}
    \item ``Toys'': `collage'
    \item ``Bento Box'': `collage'
    \item ``Cake'': `collage, warped'
    \item ``Veggie Face'': `collage, plastic, bowl'
    \item ``Striped Sweater'': `collage, backlit, ugly, disfigured, deformed'
    \item ``Ceramic Bowl'': `collage, ugly, disfigured, warped'
    \item ``Red Skirt'': `'
\end{itemize}

\noindent In Figures \ref{fig:fruit_def} through \ref{fig:sweater_def}, we illustrate individual layers for some of the collages tested:

\begin{figure}[!htbp]
    \centering
    \includegraphics[width=0.8\linewidth]{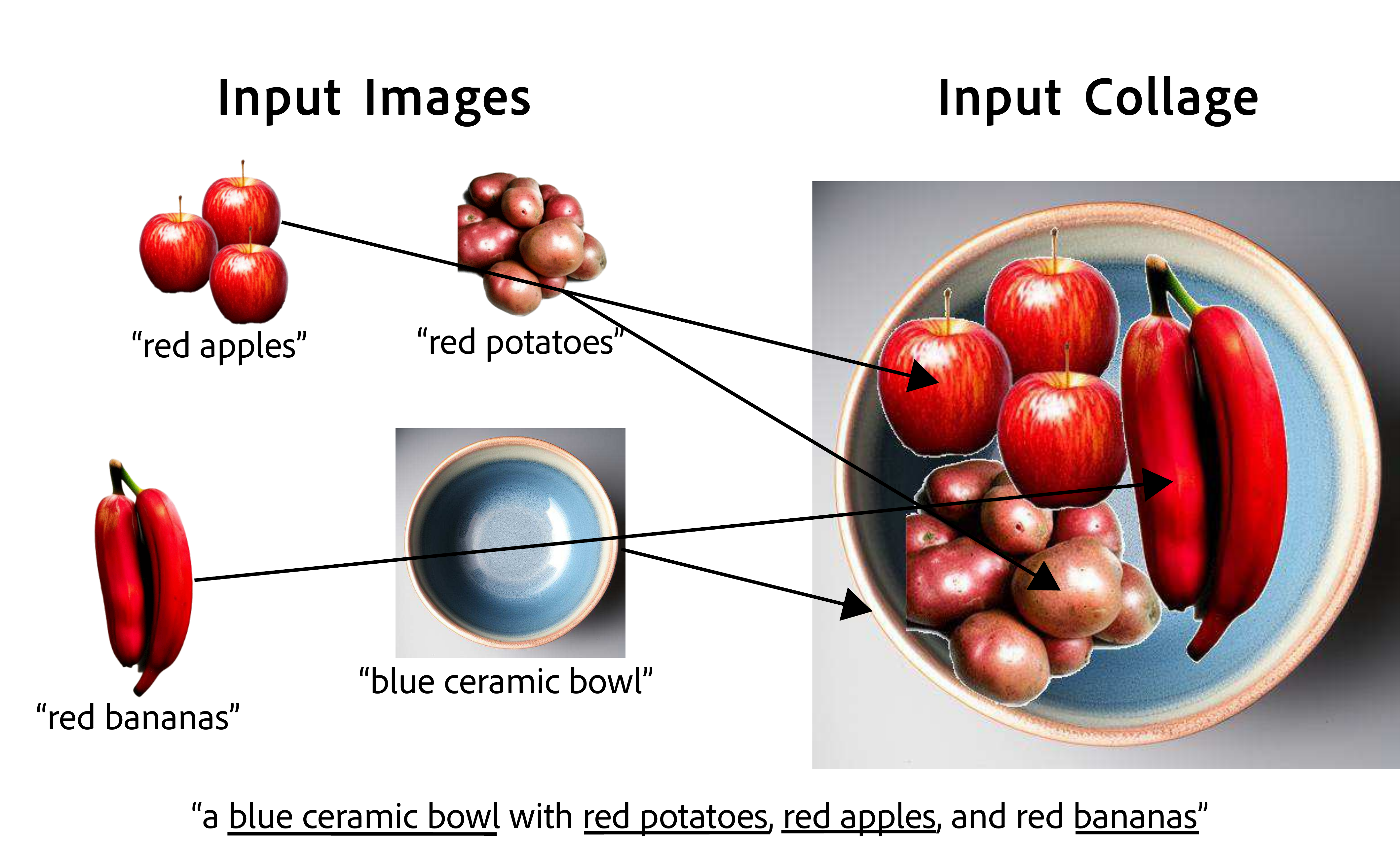}
    \caption{\textbf{Fruit} collage definition}
    \label{fig:fruit_def}
\end{figure}

\begin{figure}[!htbp]
    \centering
    \includegraphics[width=0.8\linewidth]{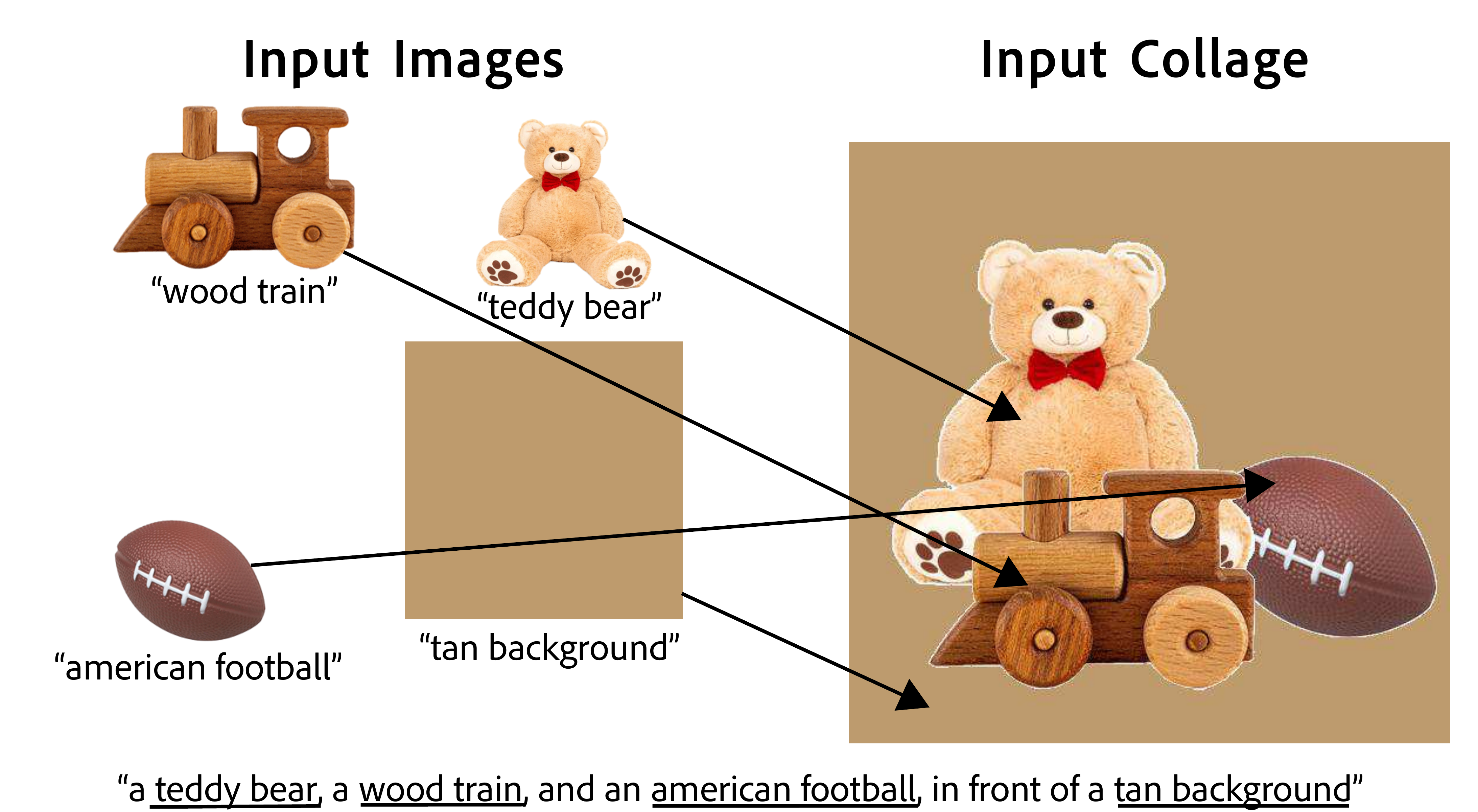}
    \caption{\textbf{Toys} collage definition}
    \label{fig:toy_def}
\end{figure}

\begin{figure}[!htbp]
    \centering
    \includegraphics[width=0.8\linewidth]{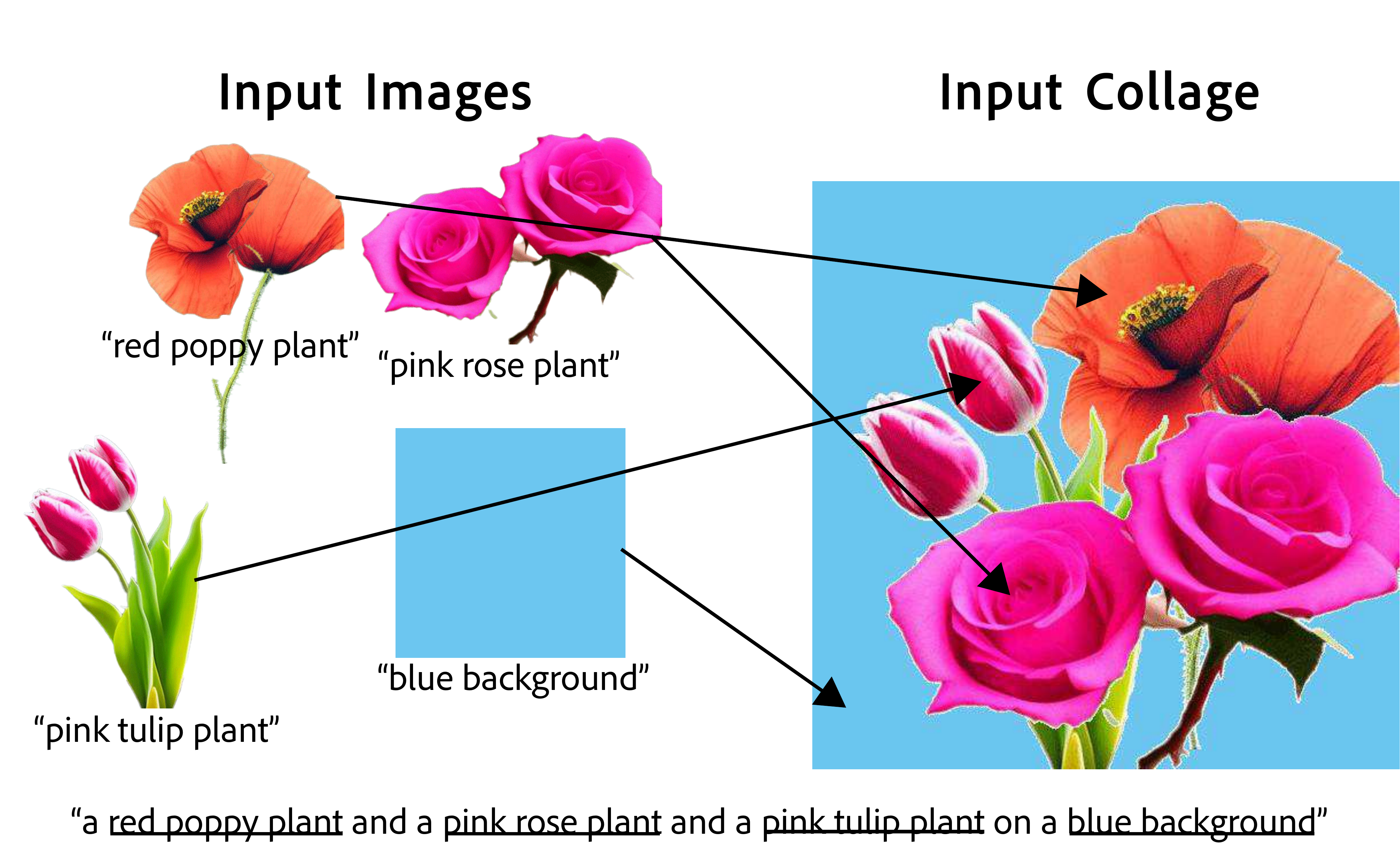}
    \caption{\textbf{Flowers} collage definition}
    \label{fig:bento_def}
\end{figure}

\begin{figure}[!htbp]
    \centering
    \includegraphics[width=0.8\linewidth]{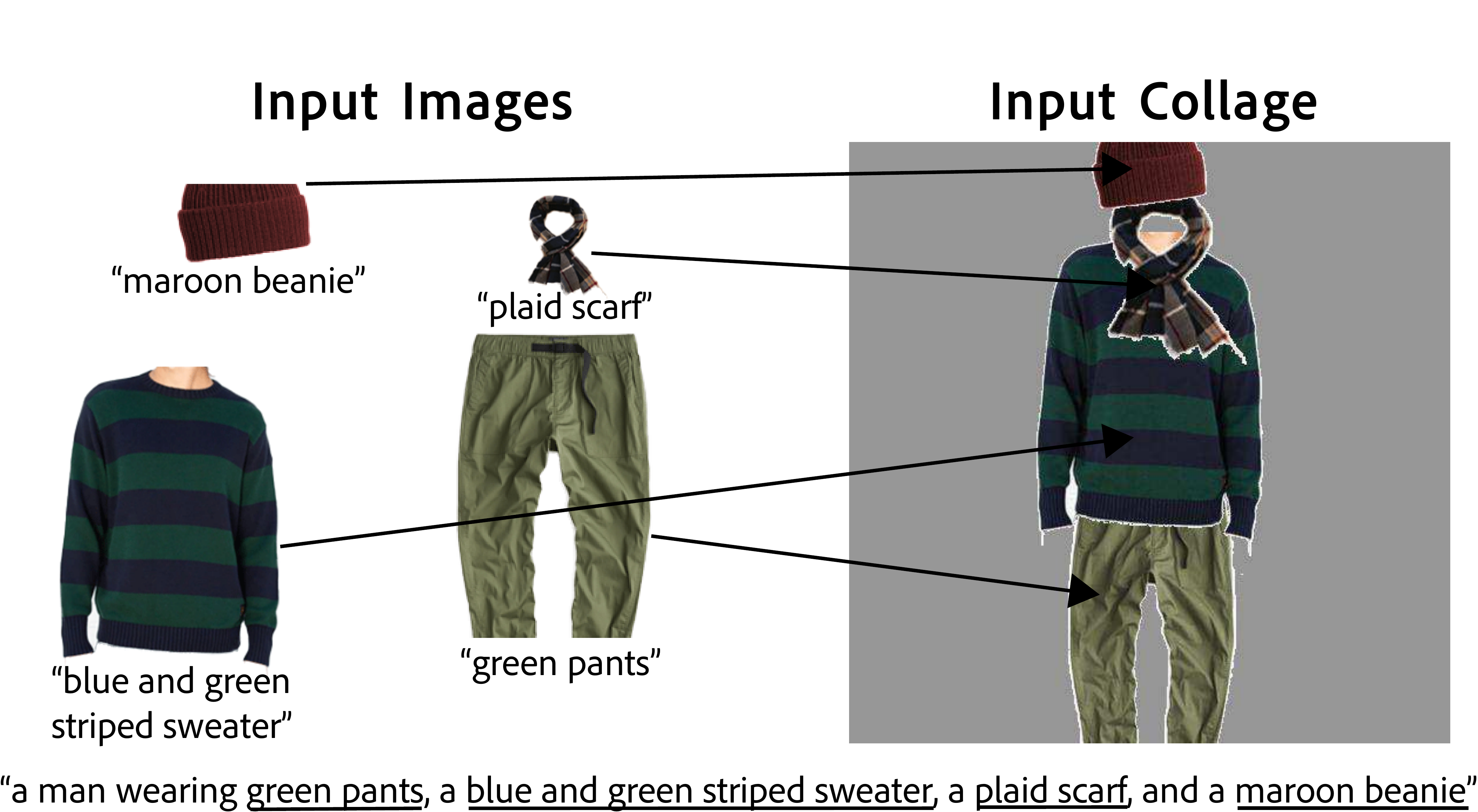}
    \caption{\textbf{Striped Sweater} collage definition}
    \label{fig:sweater_def}
\end{figure}

\pagebreak

\section{Automatic parameter tuning}

\begin{figure}
    \begin{adjustbox}{max size={\linewidth}{\textheight}}
        \begin{tabular}[t]{p{.2\linewidth}|p{.2\linewidth}|p{.2\linewidth}p{.2\linewidth}p{.2\linewidth}}
            \hfil\textbf{Prompt} & \hfil\textbf{Input Layers} &  & \hfil\textbf{Outputs} & \\
        \hline
        {\begin{small}A \ul{house} with a \ul{pink cherry blossom} next to a \ul{swimming pool} with a stone pool deck in the \ul{backyard}, \ul{sky with birds flying} in the background\end{small}} & \includegraphics[width=\linewidth,valign=t]{interface_images/house_input.png}
         & \includegraphics[width=\linewidth,valign=t]{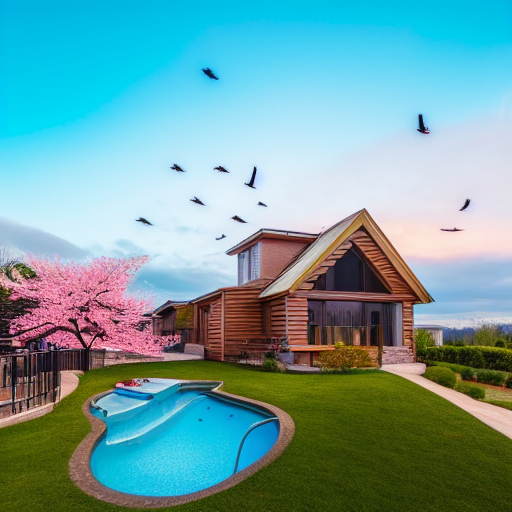} & \includegraphics[width=\linewidth,valign=t]{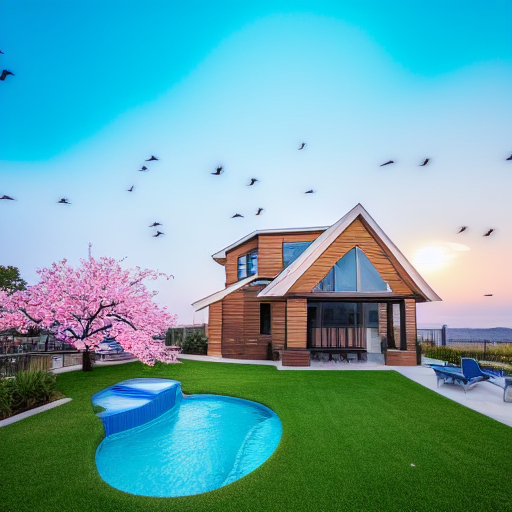} & \includegraphics[width=\linewidth,valign=t]{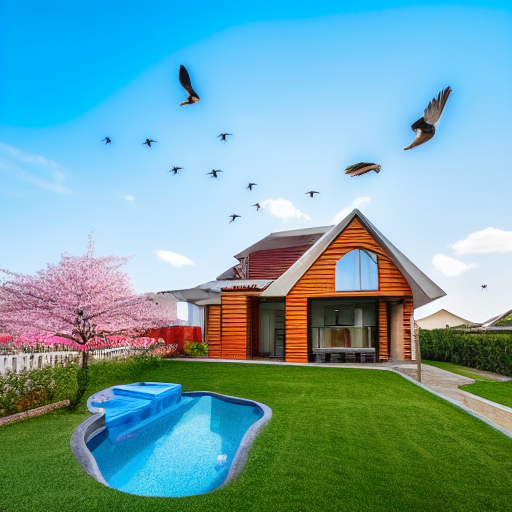} \\
         {\begin{small}A \ul{pirate ship} moving across a \ul{stormy ocean with waves} colliding into a \ul{rocky shore} containing a \ul{lighthouse} on top, \ul{dark storm clouds} with lightning in the background\end{small}} & \includegraphics[width=\linewidth,valign=t]{interface_images/ship_input.png}
         & \includegraphics[width=\linewidth,valign=t]{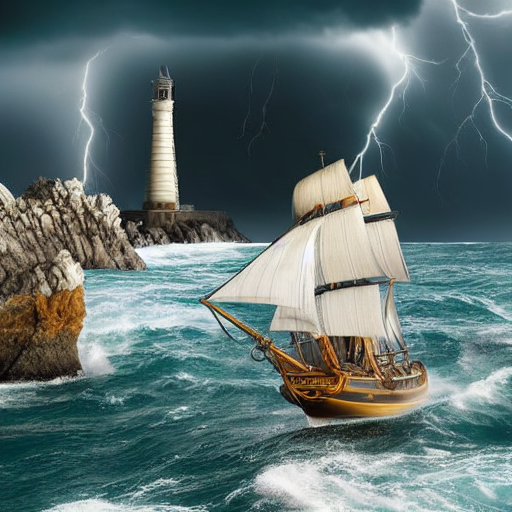} & \includegraphics[width=\linewidth,valign=t]{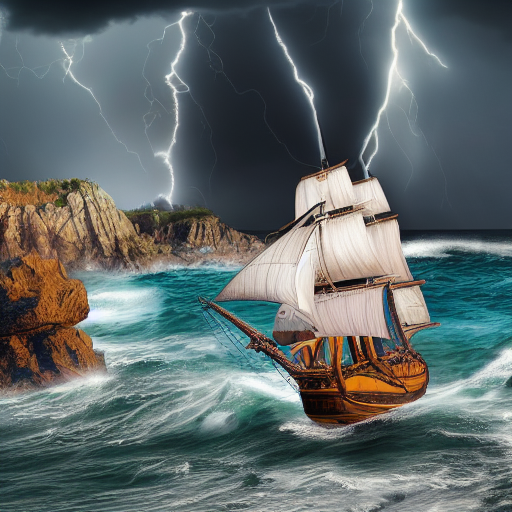} & \includegraphics[width=\linewidth,valign=t]{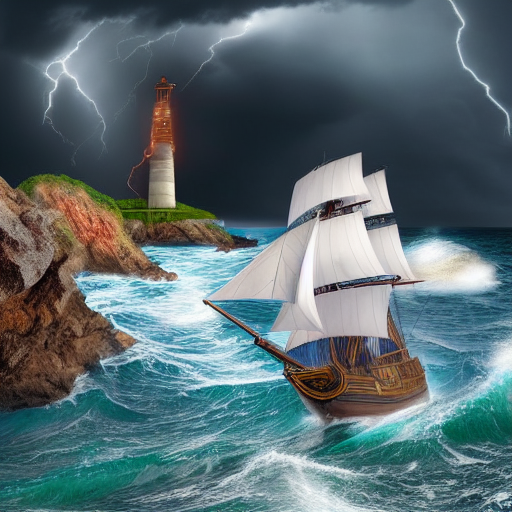} \\
        {\begin{small}A \ul{room} with a \ul{couch with pillows} in the center, \ul{wooden table} with a \ul{lamp} on top, \ul{window} with a \ul{potted plant} on the sill, \ul{carpet} with a \ul{blue mug} on top of a \ul{wooden coffee table}\end{small}}& \includegraphics[width=\linewidth,valign=t]{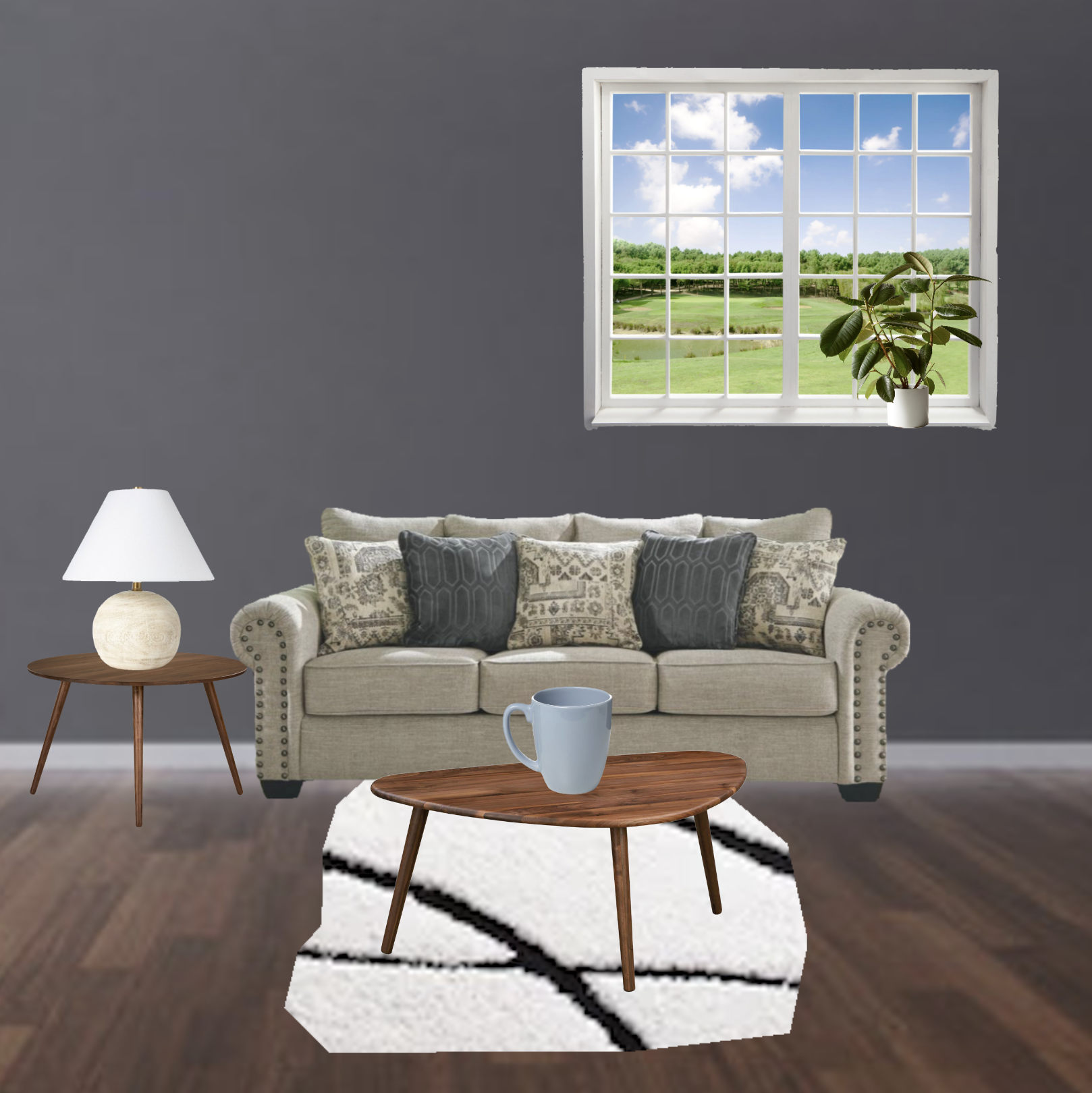}
         & \includegraphics[width=\linewidth,valign=t]{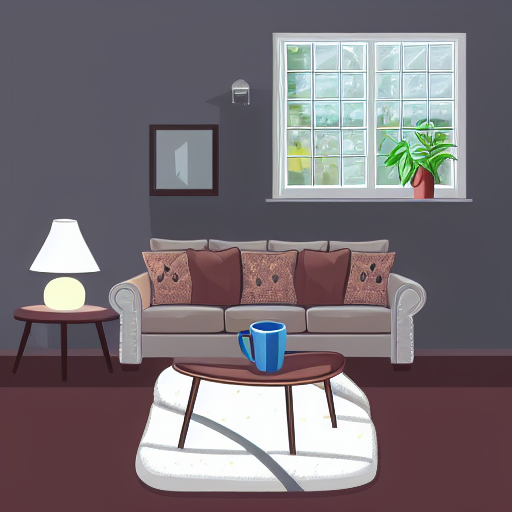} & \includegraphics[width=\linewidth,valign=t]{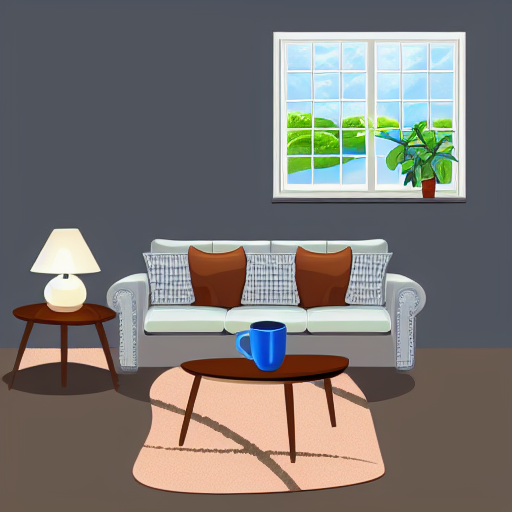} & \includegraphics[width=\linewidth,valign=t]{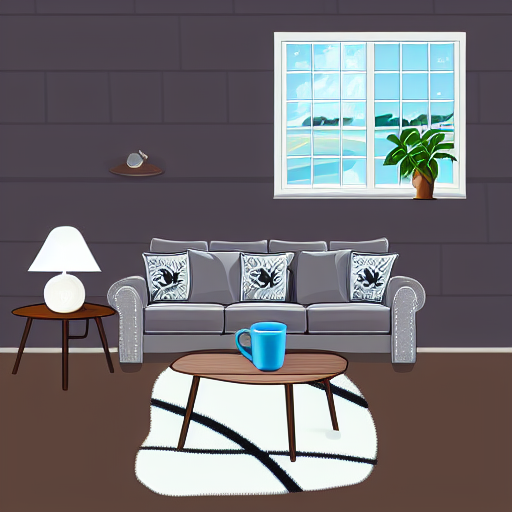} \\
        \end{tabular}
    \end{adjustbox}
    \caption{Our automated heuristic-based parameter adjustment algorithm generates compelling images across several seeds for all three scenes. Even for the living room scene, which contains nine layers of varying sizes, the algorithm is able to generate a high-quality image with automated parameters.}
    \label{fig:autoparams}
\end{figure}

We devise an automatic heuristic-based parameter adjustment algorithm to aid the user in navigating the design space of parameters governing spatial fidelity, appearance fidelity, and harmonization. We utilize the following heuristics in the algorithm:
\begin{enumerate}
    \item The noise strength is set to a lower value for foreground objects and a higher value for background objects. Conversely, canny edge strength via ControlNet is set to a higher value for foreground objects and a lower value for background objects.  These heuristics are chosen because users tend to prefer preserving the visual appearance of subjects in the foreground while trading off visual fidelity in the background for harmonization. Foreground objects and background objects are determined via layer order.
    \item Cross attention strengths are set to a higher values for foreground and smaller objects; they are set to lower values for background and larger objects. Objects in the foreground and smaller objects tend to be omitted with low cross attention strengths. Users generally care more about these layers since they tend to be important components of the image composition. We determine whether an object is “small” or “large” by evaluating the scale of the size relative to the entire canvas.
\end{enumerate}

In \cref{fig:autoparams}, the automatic parameter adjustment algorithm is able to generate compelling images with high spatial fidelity, appearance fidelity, and harmonization for scenes of varying complexity. Even for the living room scene, which contains nine layers of varying sizes, the algorithm is able to generate a high-quality image with automated parameters.

\section{Robustness to Random Seed}

\begin{figure}[!htbp]
    \centering
\begin{adjustbox}{max size={\textwidth}{\textheight}}
    \begin{tabular}[t]{p{.0\linewidth}p{.2\linewidth}|p{.2\linewidth}p{.2\linewidth}p{.2\linewidth}p{.2\linewidth}}
        & \hfil\textbf{Input Layers} & \hfil\textbf{GH} & \hfil\textbf{GH+CA} & \hfil\textbf{GH+CA+TI} & \hfil\textbf{GH+CA+TI+LN}\\
        & \includegraphics[width=\linewidth,valign=m]{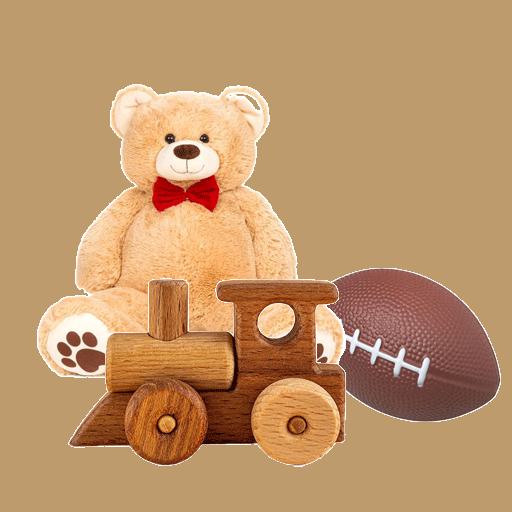} & \includegraphics[width=\linewidth,valign=m]{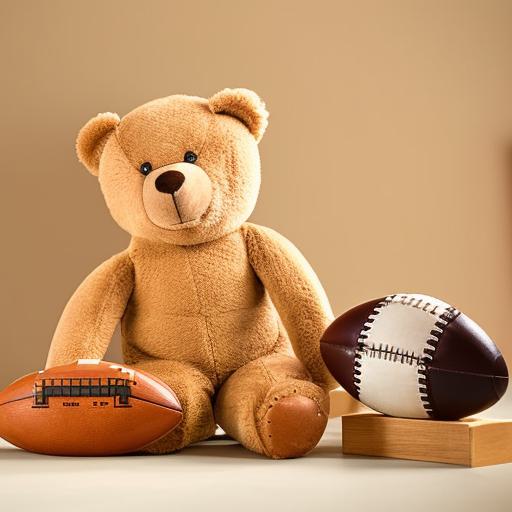} & \includegraphics[width=\linewidth,valign=m]{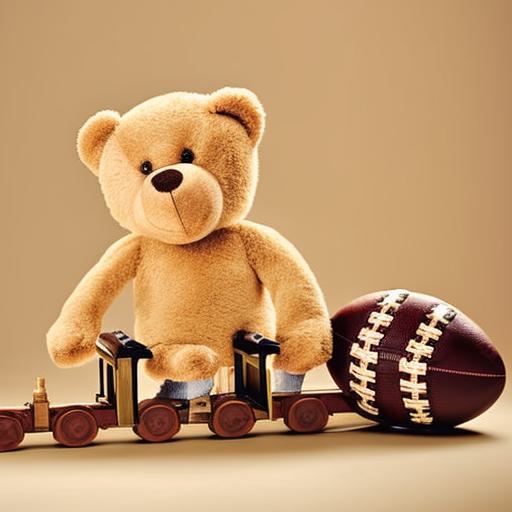} & \includegraphics[width=\linewidth,valign=m]{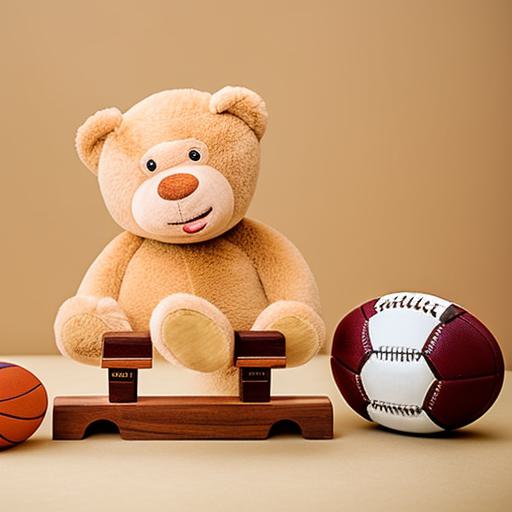} & \includegraphics[width=\linewidth,valign=m]{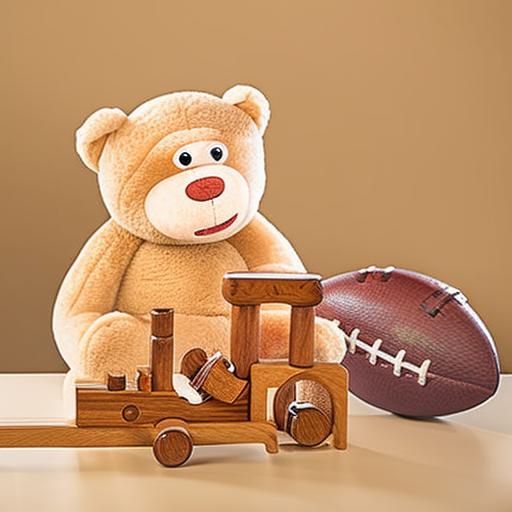} \\
        & \includegraphics[width=\linewidth,valign=m]{claim_1/toyBoxFinal/init.jpg} & \includegraphics[width=\linewidth,valign=m]{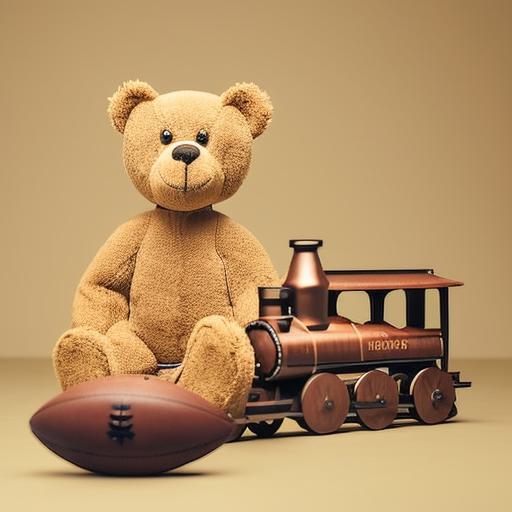} & \includegraphics[width=\linewidth,valign=m]{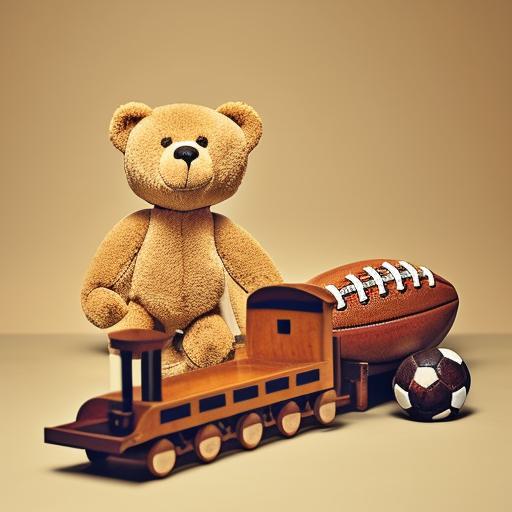} & \includegraphics[width=\linewidth,valign=m]{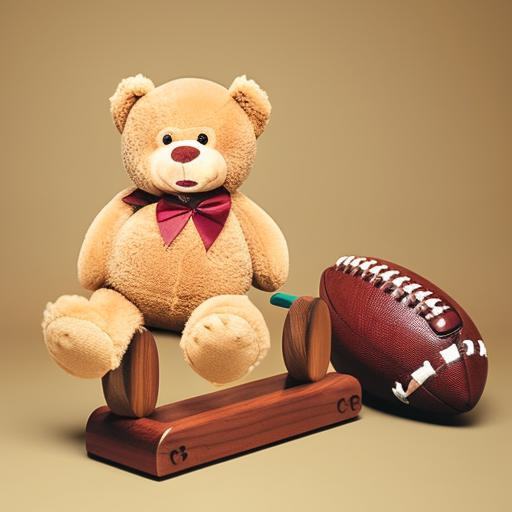} & \includegraphics[width=\linewidth,valign=m]{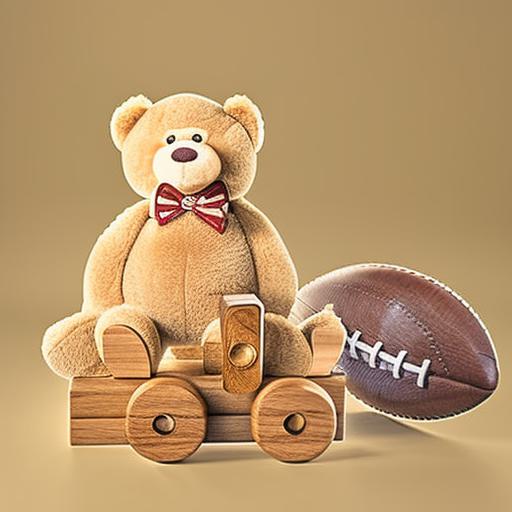} \\
        & \includegraphics[width=\linewidth,valign=m]{claim_1/toyBoxFinal/init.jpg} & \includegraphics[width=\linewidth,valign=m]{claim_1/toyBoxFinal/img2img-no_cac-no_ft-no_mask/8.jpg} & \includegraphics[width=\linewidth,valign=m]{claim_1/toyBoxFinal/img2img-with_cac-no_ft-no_mask/8.jpg} & \includegraphics[width=\linewidth,valign=m]{claim_1/toyBoxFinal/img2img-with_cac-with_ft-no_mask/8.jpg} & \includegraphics[width=\linewidth,valign=m]{claim_1/toyBoxFinal/img2img-with_cac-with_ft-with_mask/8.jpg} \\
        & \includegraphics[width=\linewidth,valign=m]{claim_1/toyBoxFinal/init.jpg} & \includegraphics[width=\linewidth,valign=m]{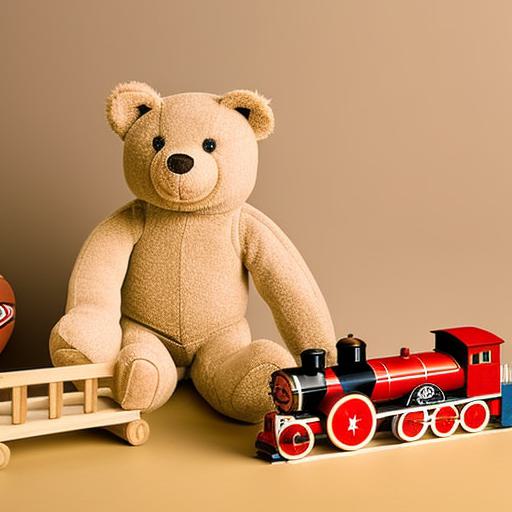} & \includegraphics[width=\linewidth,valign=m]{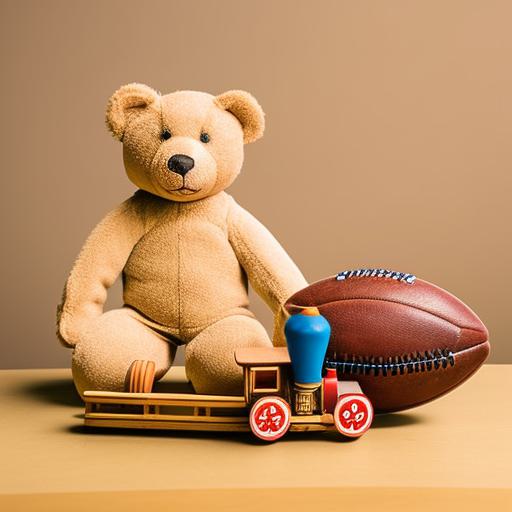} & \includegraphics[width=\linewidth,valign=m]{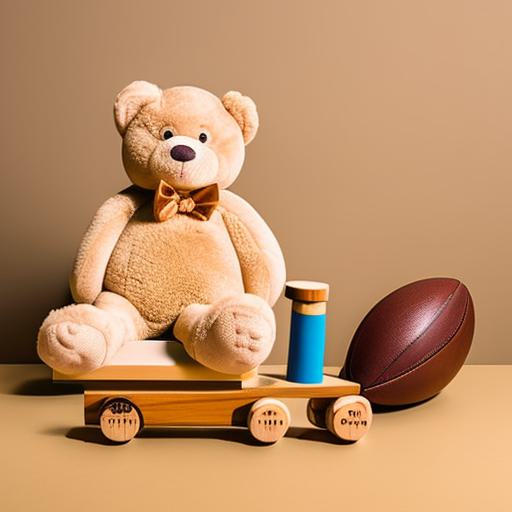} & \includegraphics[width=\linewidth,valign=m]{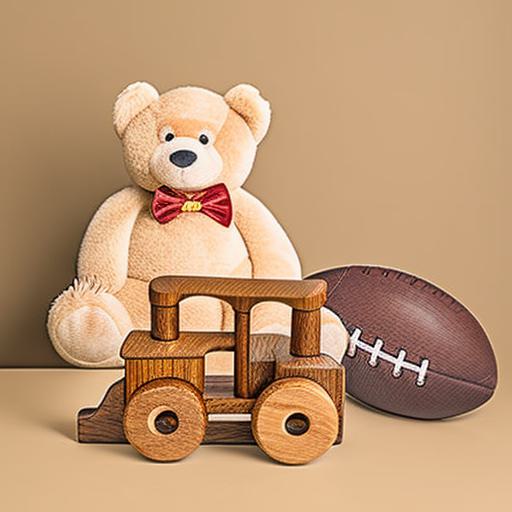} \\
        & \includegraphics[width=\linewidth,valign=m]{claim_1/toyBoxFinal/init.jpg} & \includegraphics[width=\linewidth,valign=m]{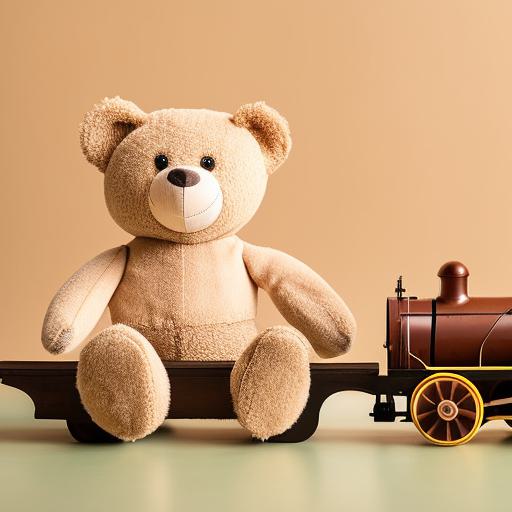} & \includegraphics[width=\linewidth,valign=m]{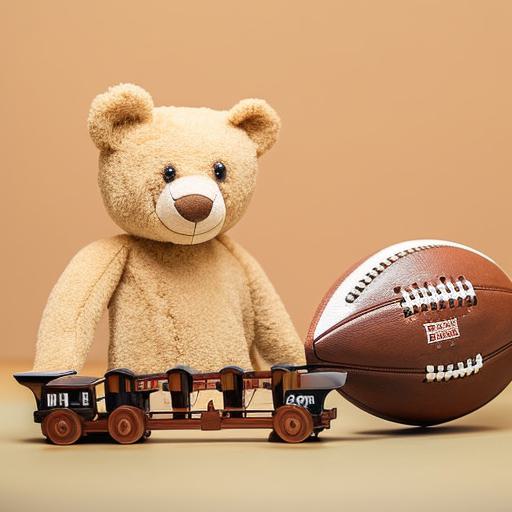} & \includegraphics[width=\linewidth,valign=m]{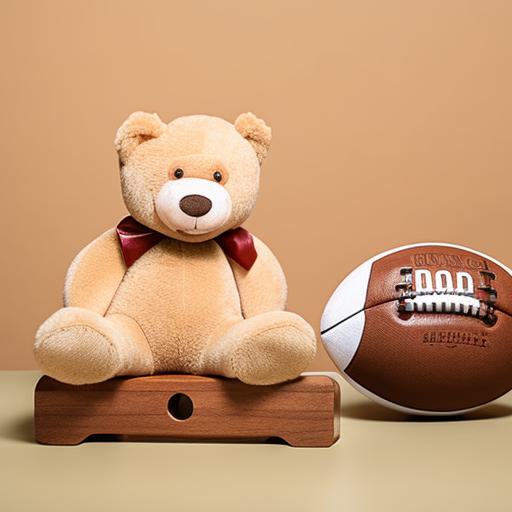} & \includegraphics[width=\linewidth,valign=m]{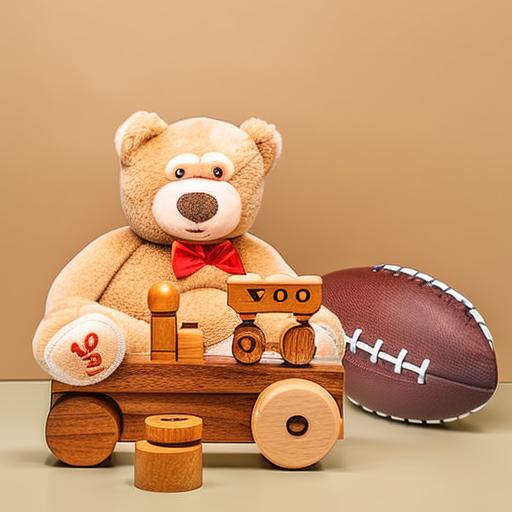} \\
        & \includegraphics[width=\linewidth,valign=m]{claim_1/toyBoxFinal/init.jpg} & \includegraphics[width=\linewidth,valign=m]{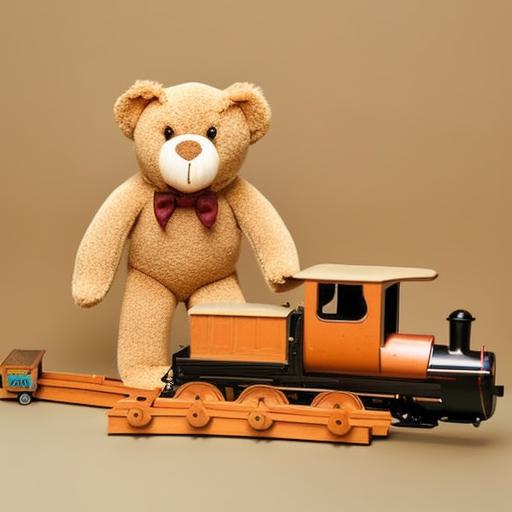} & \includegraphics[width=\linewidth,valign=m]{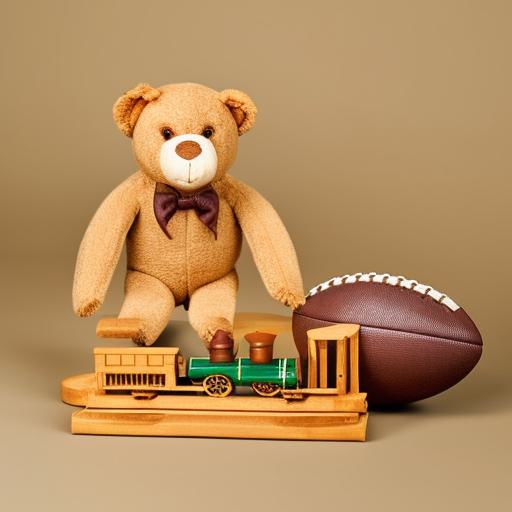} & \includegraphics[width=\linewidth,valign=m]{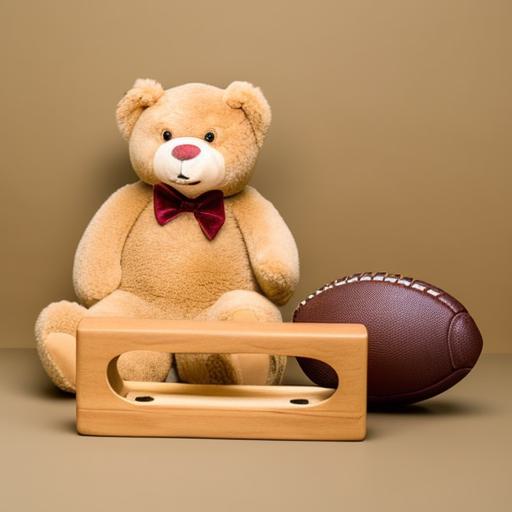} & \includegraphics[width=\linewidth,valign=m]{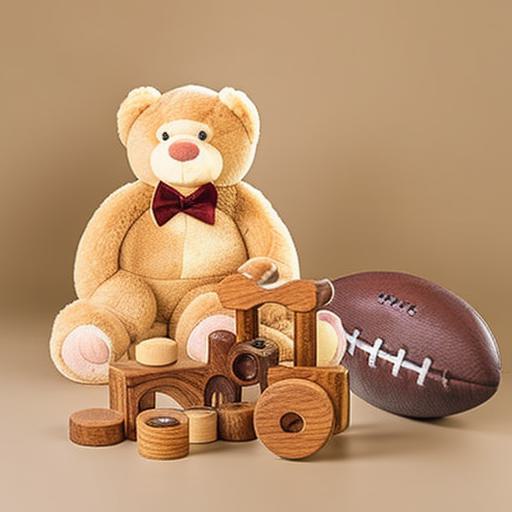} \\
    \end{tabular}
\end{adjustbox}
\caption{ ``a \ul{teddy bear}, a \ul{wood train}, and an \ul{american football}, in front of a \ul{tan background}'' }
\label{fig:toyBox}
\end{figure}

\begin{figure}[!htbp]
    \begin{adjustbox}{max size={\textwidth}{\textheight}}
    \begin{tabular}[t]{p{.0\linewidth}p{.2\linewidth}|p{.2\linewidth}p{.2\linewidth}p{.2\linewidth}p{.2\linewidth}}
        & \hfil\textbf{Input Layers} & \hfil\textbf{GH} & \hfil\textbf{GH+CA} & \hfil\textbf{GH+CA+TI} & \hfil\textbf{GH+CA+TI+LN}\\
        & \includegraphics[width=\linewidth,valign=m]{claim_1/bentoBoxFinal/init.jpg} & \includegraphics[width=\linewidth,valign=m]{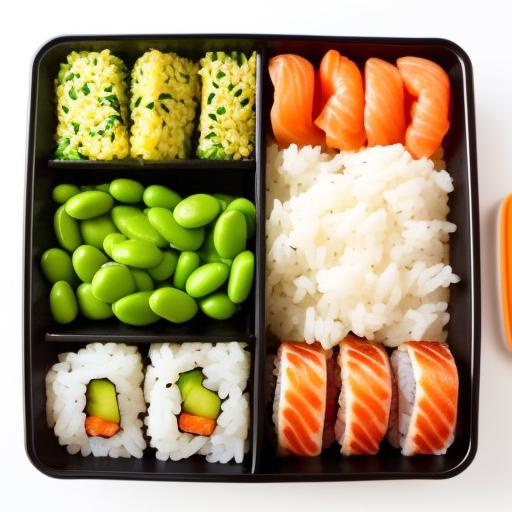} & \includegraphics[width=\linewidth,valign=m]{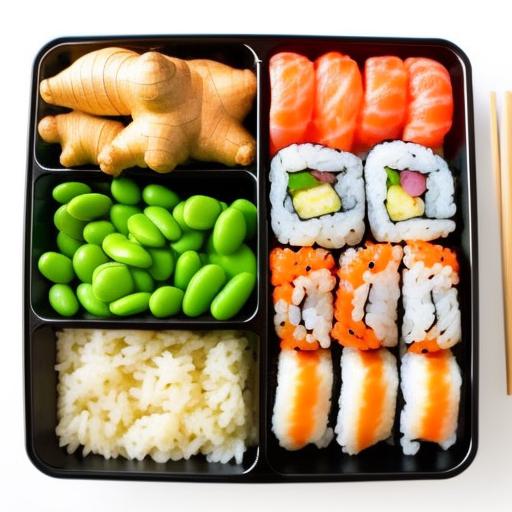} & \includegraphics[width=\linewidth,valign=m]{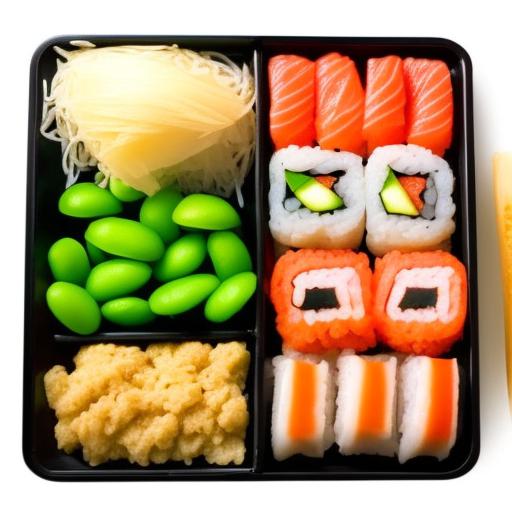} & \includegraphics[width=\linewidth,valign=m]{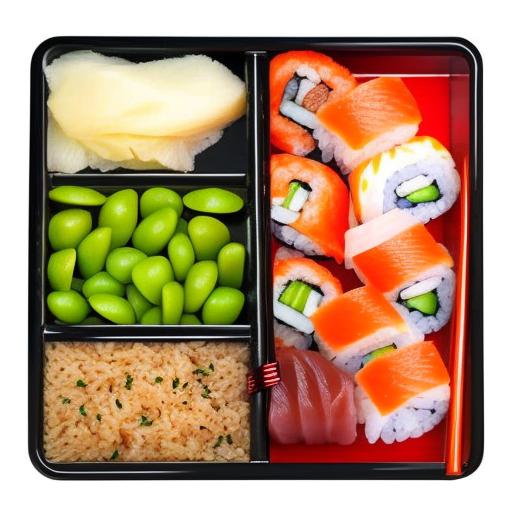} \\
        & \includegraphics[width=\linewidth,valign=m]{claim_1/bentoBoxFinal/init.jpg} & \includegraphics[width=\linewidth,valign=m]{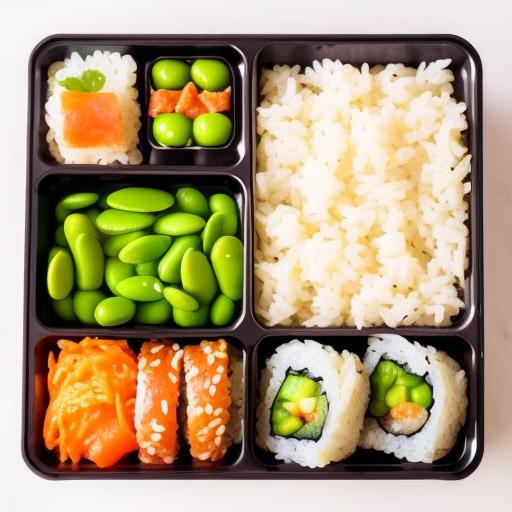} & \includegraphics[width=\linewidth,valign=m]{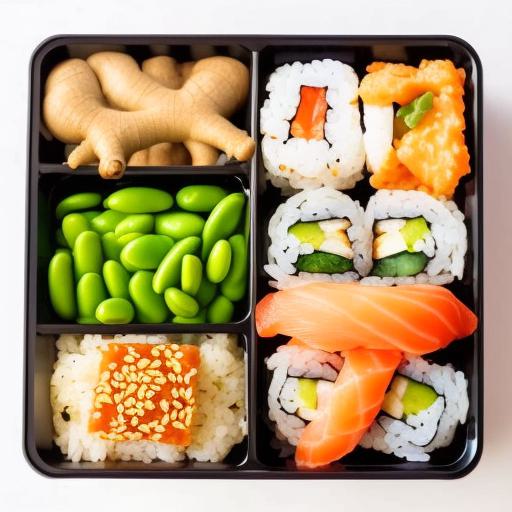} & \includegraphics[width=\linewidth,valign=m]{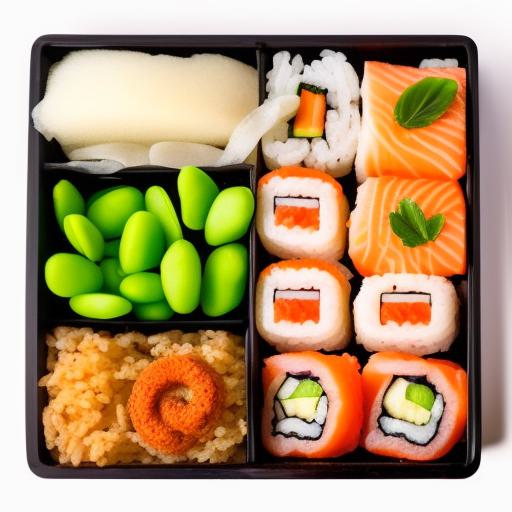} & \includegraphics[width=\linewidth,valign=m]{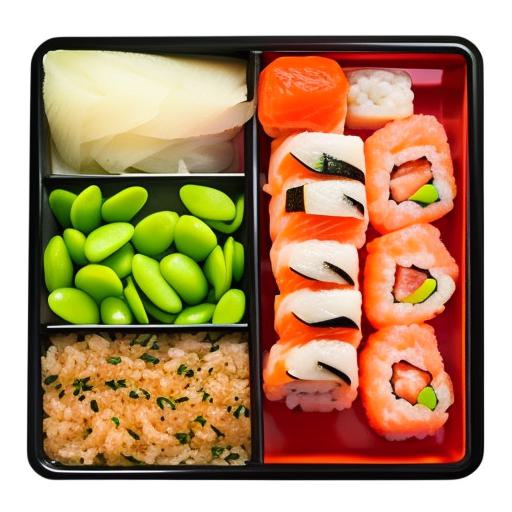} \\
        & \includegraphics[width=\linewidth,valign=m]{claim_1/bentoBoxFinal/init.jpg} & \includegraphics[width=\linewidth,valign=m]{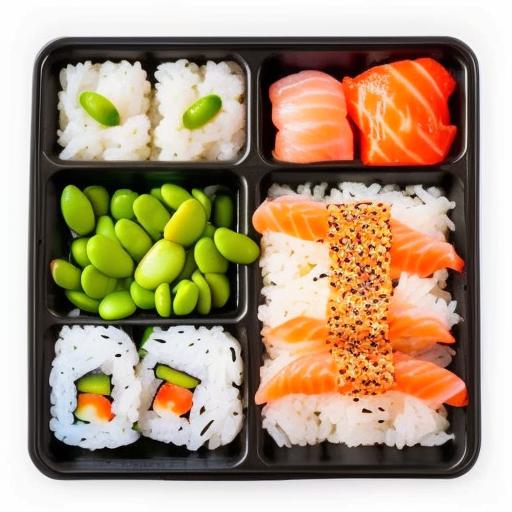} & \includegraphics[width=\linewidth,valign=m]{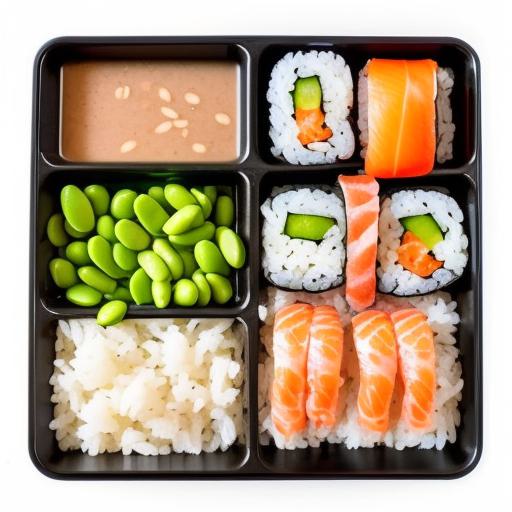} & \includegraphics[width=\linewidth,valign=m]{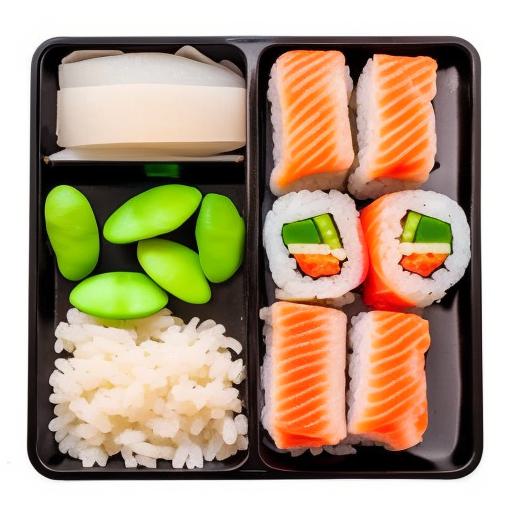} & \includegraphics[width=\linewidth,valign=m]{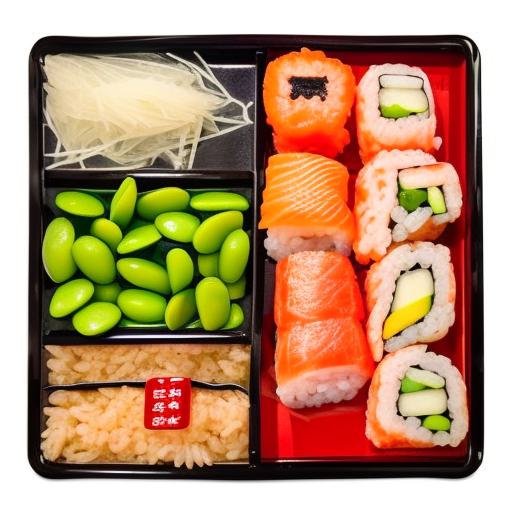} \\
        & \includegraphics[width=\linewidth,valign=m]{claim_1/bentoBoxFinal/init.jpg} & \includegraphics[width=\linewidth,valign=m]{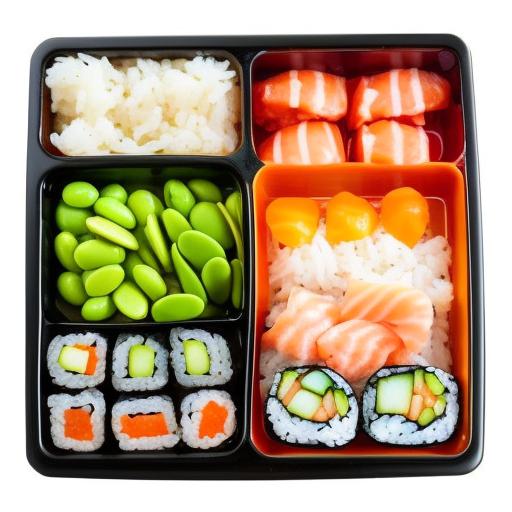} & \includegraphics[width=\linewidth,valign=m]{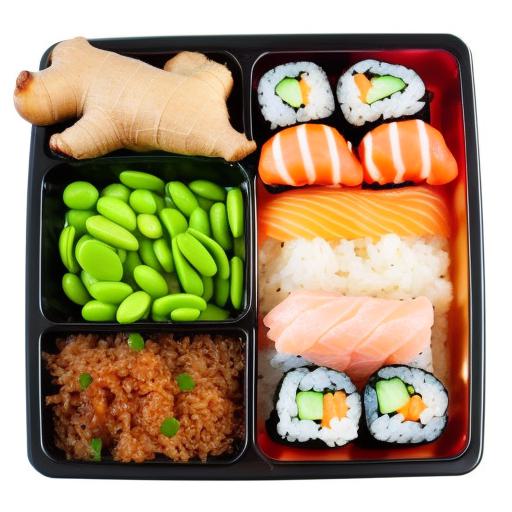} & \includegraphics[width=\linewidth,valign=m]{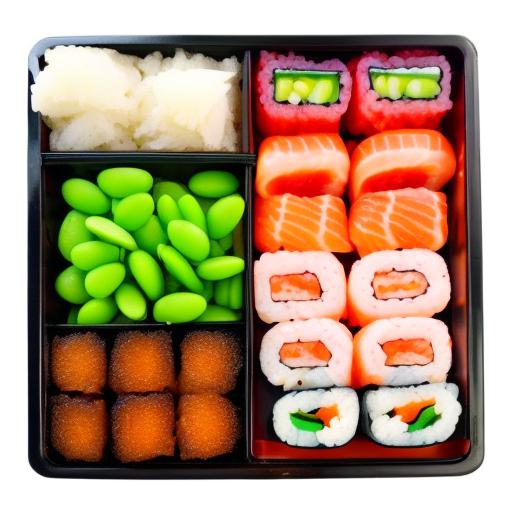} & \includegraphics[width=\linewidth,valign=m]{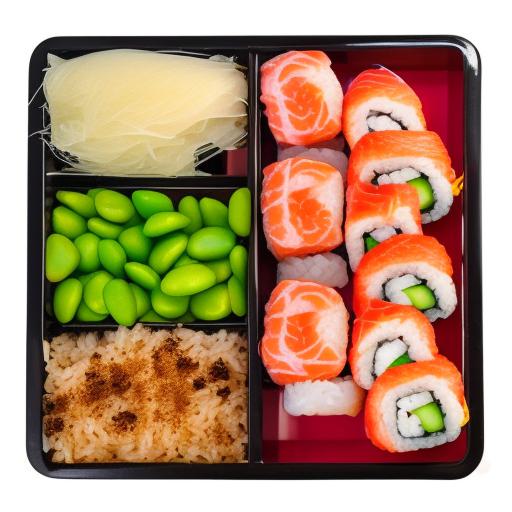} \\
        & \includegraphics[width=\linewidth,valign=m]{claim_1/bentoBoxFinal/init.jpg} & \includegraphics[width=\linewidth,valign=m]{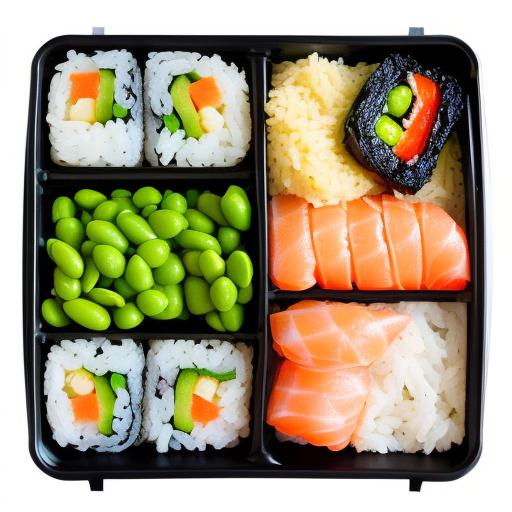} & \includegraphics[width=\linewidth,valign=m]{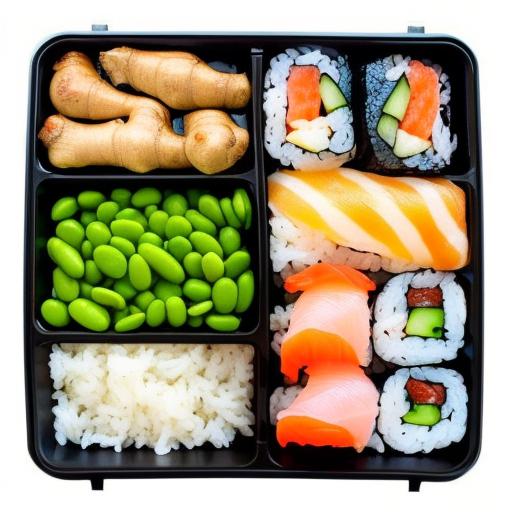} & \includegraphics[width=\linewidth,valign=m]{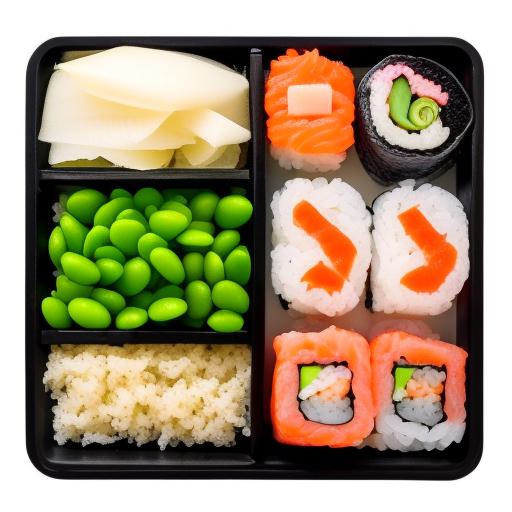} & \includegraphics[width=\linewidth,valign=m]{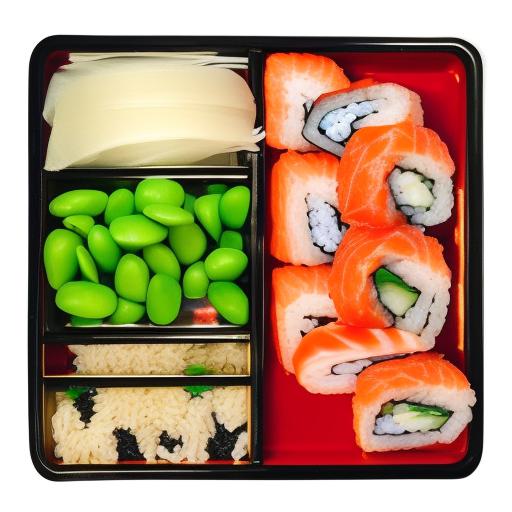} \\
        & \includegraphics[width=\linewidth,valign=m]{claim_1/bentoBoxFinal/init.jpg} & \includegraphics[width=\linewidth,valign=m]{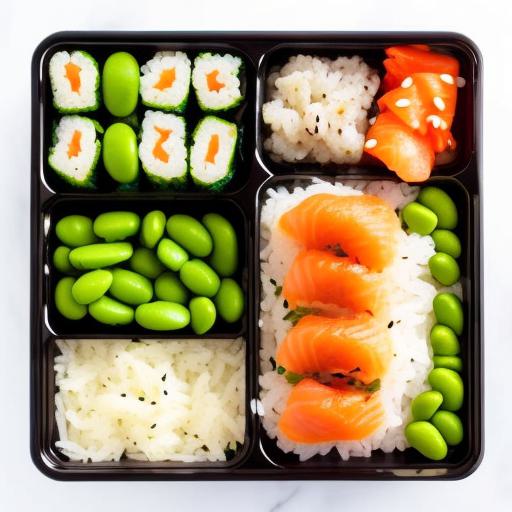} & \includegraphics[width=\linewidth,valign=m]{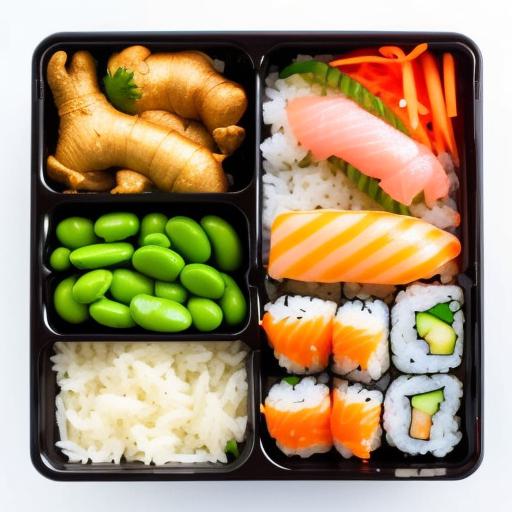} & \includegraphics[width=\linewidth,valign=m]{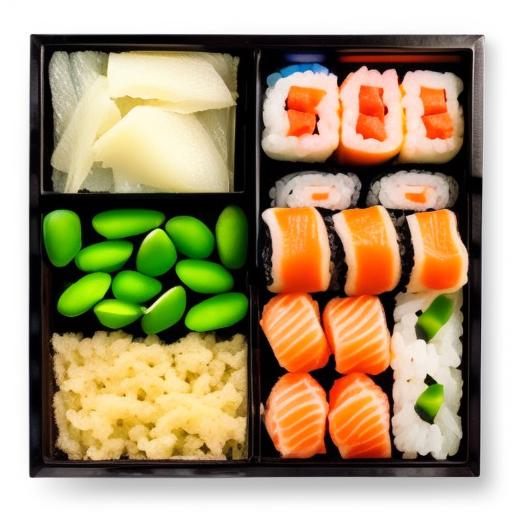} & \includegraphics[width=\linewidth,valign=m]{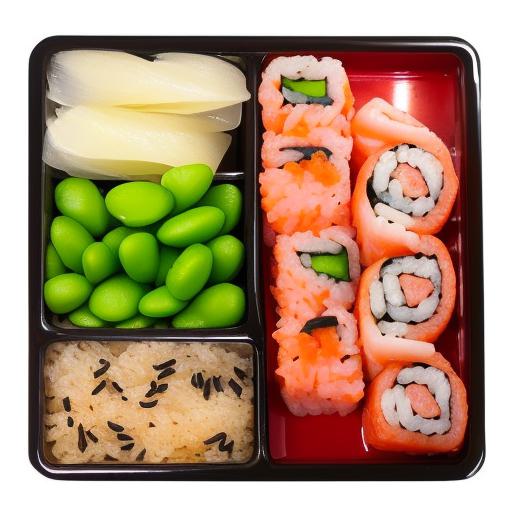} \\
    \end{tabular}
\end{adjustbox}
\caption{ ``a \ul{bento box} with \ul{rice}, \ul{edamame}, \ul{ginger}, and \ul{sushi}'' }
\label{fig:bentoBox}
\end{figure}

\begin{figure}[!htbp]
    \begin{adjustbox}{max size={\textwidth}{\textheight}}
        \begin{tabular}[t]{p{.0\linewidth}p{.2\linewidth}|p{.2\linewidth}p{.2\linewidth}p{.2\linewidth}p{.2\linewidth}}
        & \hfil\textbf{Input Layers} & \hfil\textbf{GH} & \hfil\textbf{GH+CA} & \hfil\textbf{GH+CA+TI} & \hfil\textbf{GH+CA+TI+LN}\\
        & \includegraphics[width=\linewidth,valign=m]{claim_1/arcimboldoFinal/init.jpg} & \includegraphics[width=\linewidth,valign=m]{claim_1/arcimboldoFinal/img2img-no_cac-no_ft-no_mask/6.jpg} & \includegraphics[width=\linewidth,valign=m]{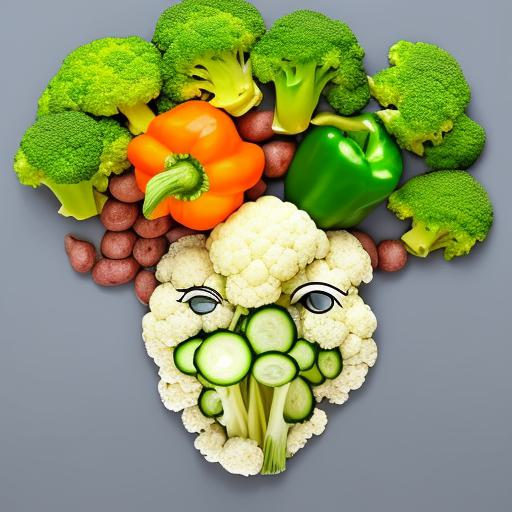} & \includegraphics[width=\linewidth,valign=m]{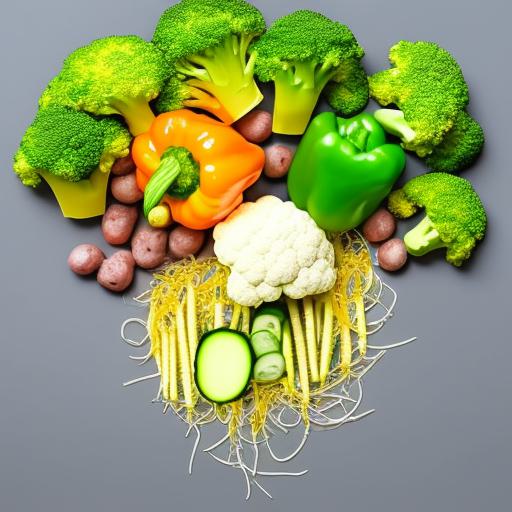} & \includegraphics[width=\linewidth,valign=m]{claim_1/arcimboldoFinal/img2img-with_cac-with_ft-with_mask/6.jpg} \\
        & \includegraphics[width=\linewidth,valign=m]{claim_1/arcimboldoFinal/init.jpg} & \includegraphics[width=\linewidth,valign=m]{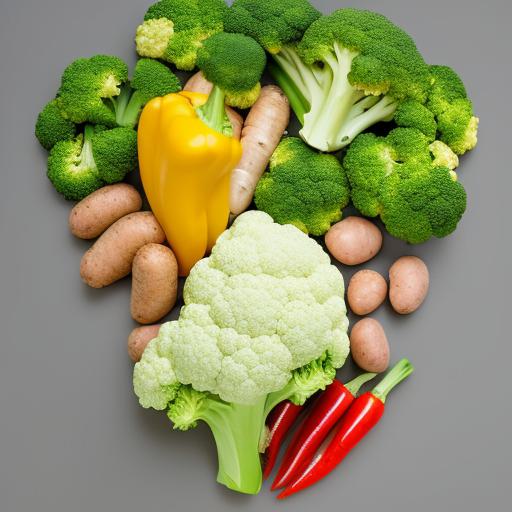} & \includegraphics[width=\linewidth,valign=m]{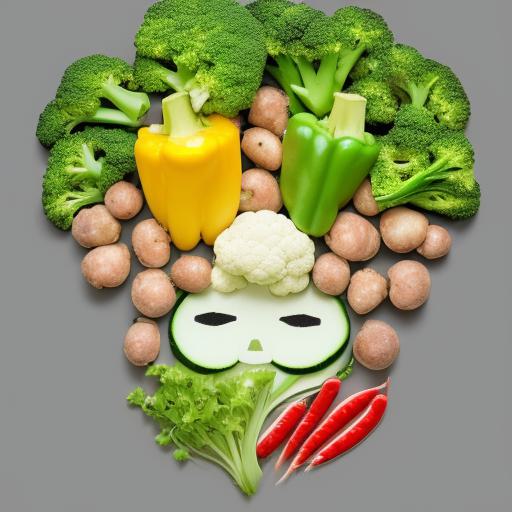} & \includegraphics[width=\linewidth,valign=m]{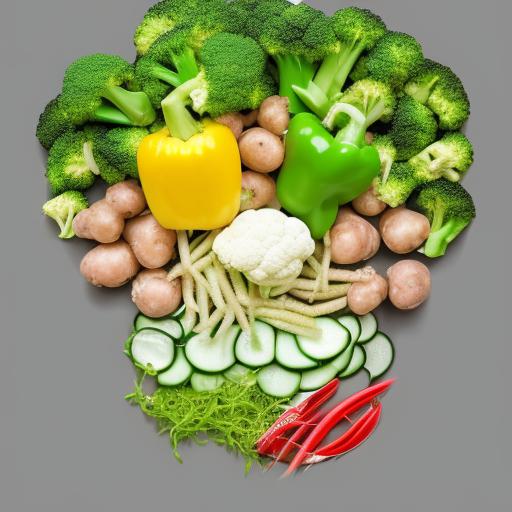} & \includegraphics[width=\linewidth,valign=m]{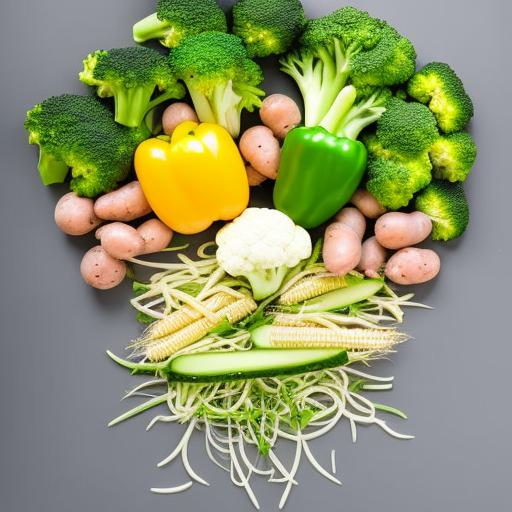} \\
        & \includegraphics[width=\linewidth,valign=m]{claim_1/arcimboldoFinal/init.jpg} & \includegraphics[width=\linewidth,valign=m]{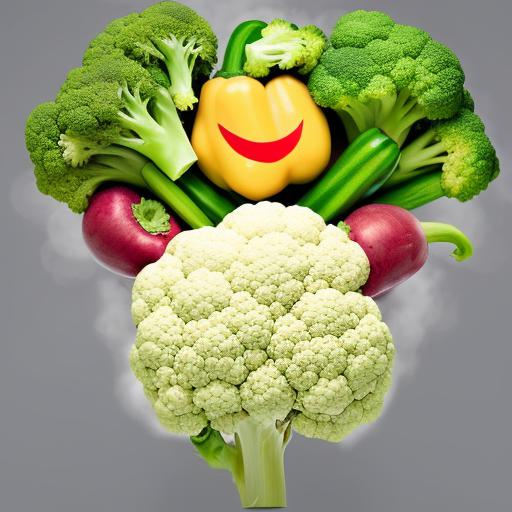} & \includegraphics[width=\linewidth,valign=m]{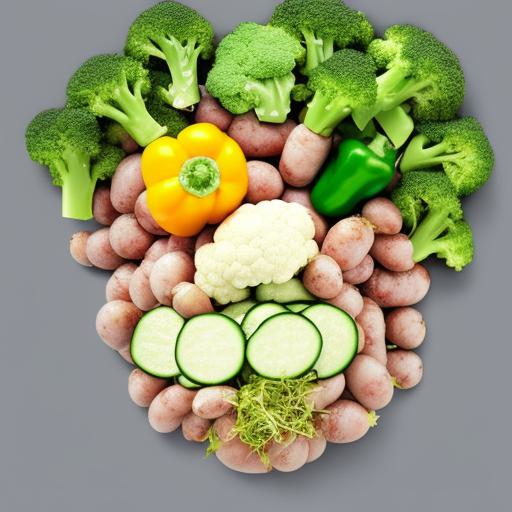} & \includegraphics[width=\linewidth,valign=m]{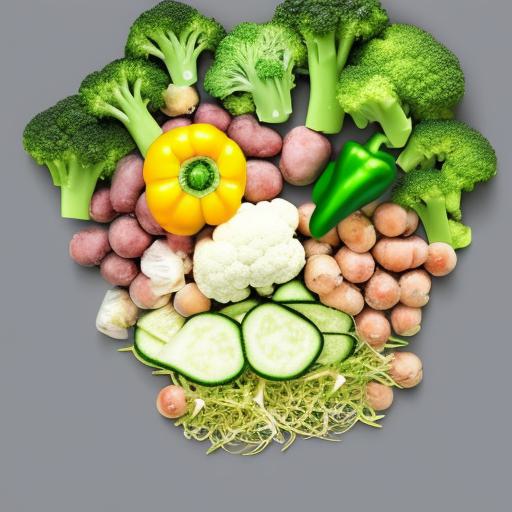} & \includegraphics[width=\linewidth,valign=m]{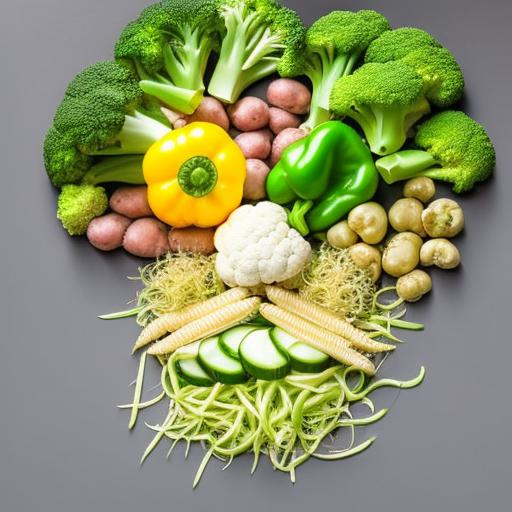} \\
        & \includegraphics[width=\linewidth,valign=m]{claim_1/arcimboldoFinal/init.jpg} & \includegraphics[width=\linewidth,valign=m]{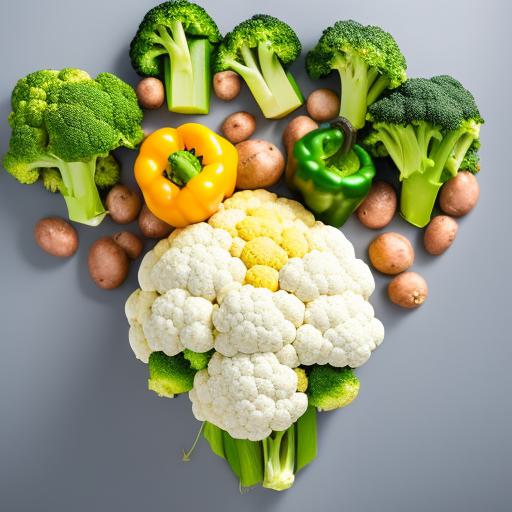} & \includegraphics[width=\linewidth,valign=m]{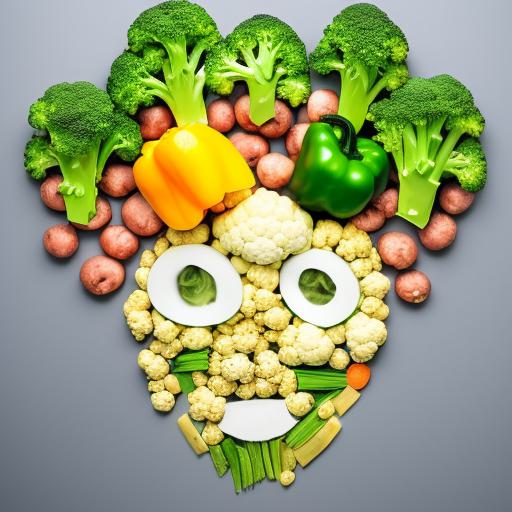} & \includegraphics[width=\linewidth,valign=m]{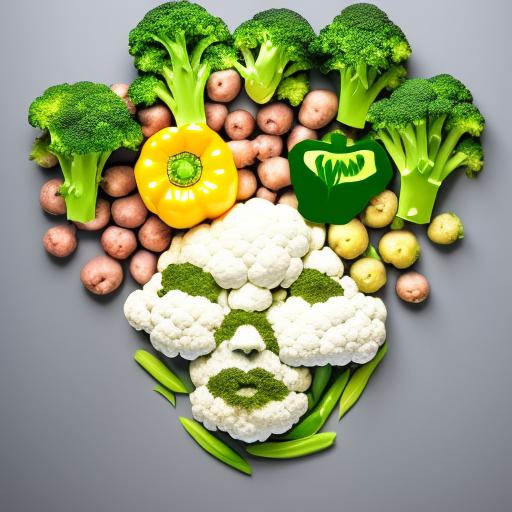} & \includegraphics[width=\linewidth,valign=m]{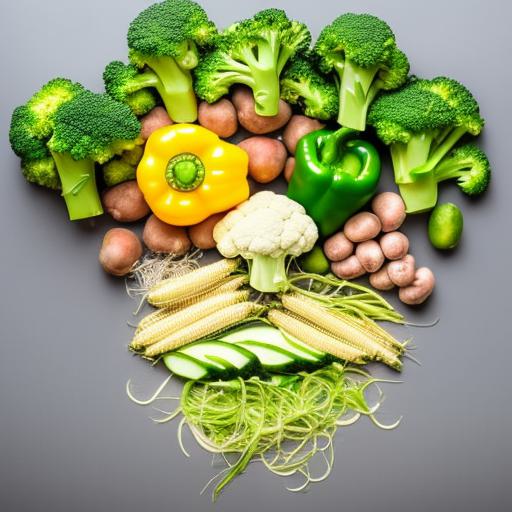} \\
        & \includegraphics[width=\linewidth,valign=m]{claim_1/arcimboldoFinal/init.jpg} & \includegraphics[width=\linewidth,valign=m]{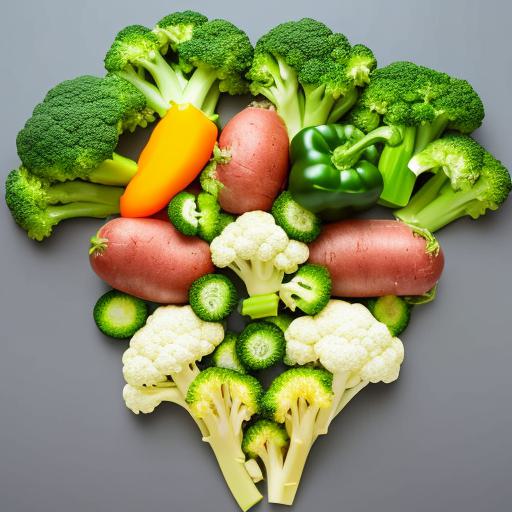} & \includegraphics[width=\linewidth,valign=m]{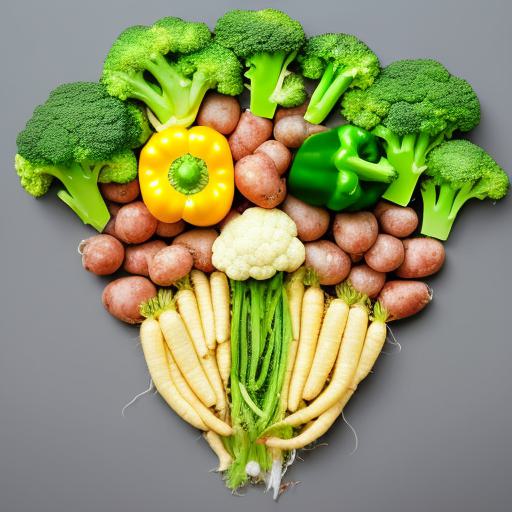} & \includegraphics[width=\linewidth,valign=m]{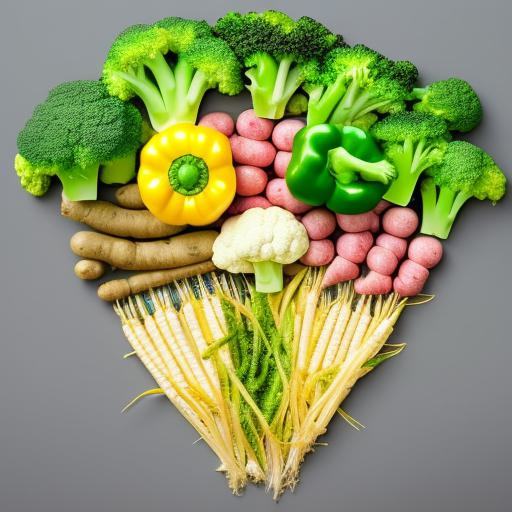} & \includegraphics[width=\linewidth,valign=m]{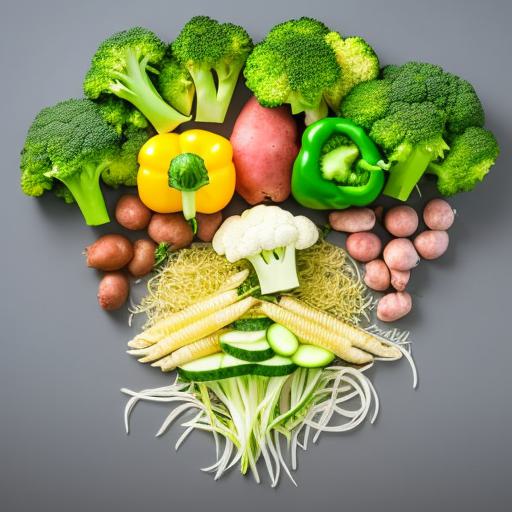} \\
        & \includegraphics[width=\linewidth,valign=m]{claim_1/arcimboldoFinal/init.jpg} & \includegraphics[width=\linewidth,valign=m]{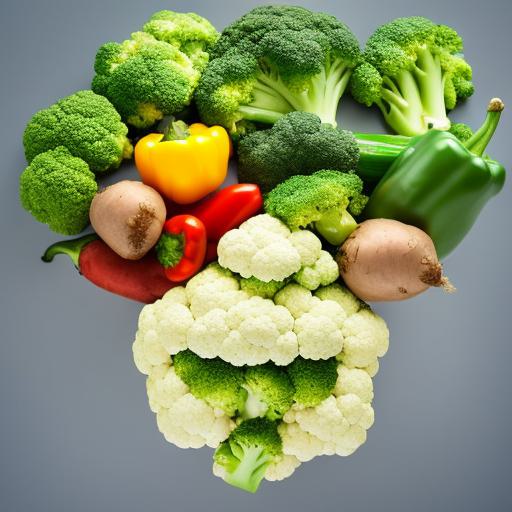} & \includegraphics[width=\linewidth,valign=m]{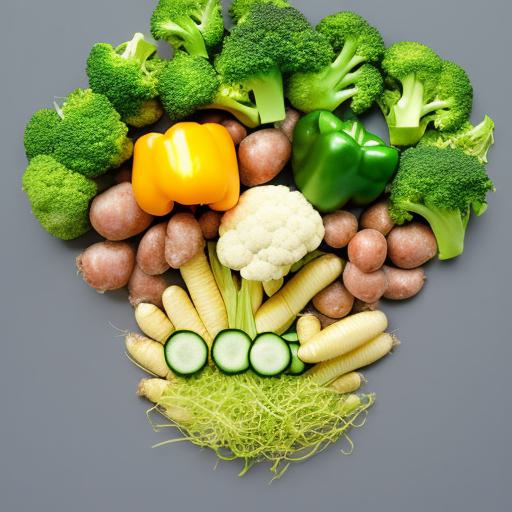} & \includegraphics[width=\linewidth,valign=m]{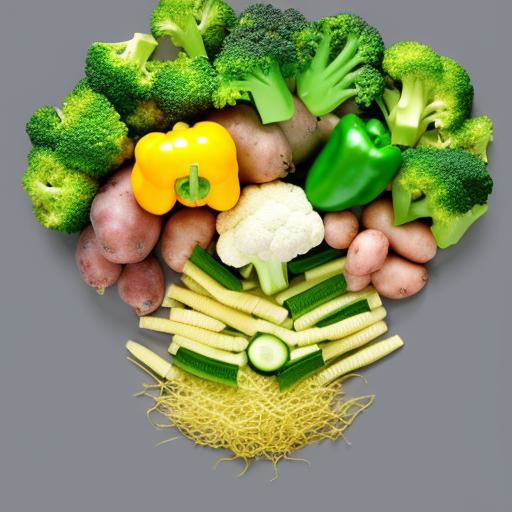} & \includegraphics[width=\linewidth,valign=m]{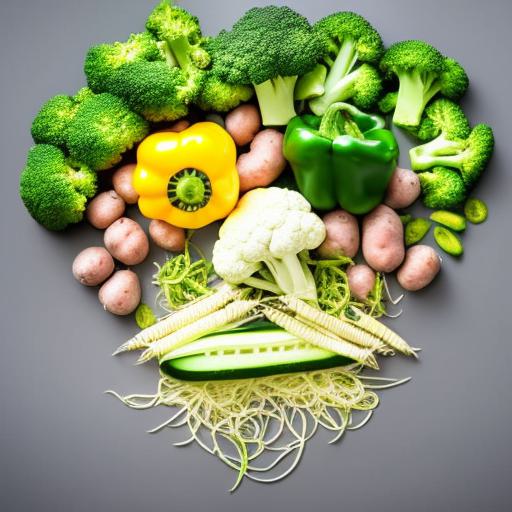} \\
    \end{tabular}
\end{adjustbox}
\caption{Prompt: ``a face made of vegetables, including a \ul{yellow bell pepper} and a \ul{green bell pepper}, a \ul{white cauliflower}, \ul{red potatoes}, \ul{baby corn}, \ul{small cucumber}, \ul{bean sprouts}, and \ul{floret broccoli}, on a \ul{grey background}''}
\label{fig:arcimboldo}
\end{figure}

\begin{figure}[!htbp]
    \centering
    \begin{adjustbox}{max size={\textwidth}{\textheight}}
    \begin{tabular}[t]{p{.0\linewidth}p{.2\linewidth}|p{.2\linewidth}p{.2\linewidth}p{.2\linewidth}p{.2\linewidth}}
        & \hfil\textbf{Input Layers} & \hfil\textbf{GH} & \hfil\textbf{GH+CA} & \hfil\textbf{GH+CA+TI} & \hfil\textbf{GH+CA+TI+LN}\\
         & \includegraphics[width=\linewidth,valign=m]{claim_1/cakeTable3Final/init.jpg} & \includegraphics[width=\linewidth,valign=m]{claim_1/cakeTable3Final/img2img-no_cac-no_ft-no_mask/0.jpg} & \includegraphics[width=\linewidth,valign=m]{claim_1/cakeTable3Final/img2img-with_cac-no_ft-no_mask/0.jpg} & \includegraphics[width=\linewidth,valign=m]{claim_1/cakeTable3Final/img2img-with_cac-with_ft-no_mask/0.jpg} & \includegraphics[width=\linewidth,valign=m]{claim_1/cakeTable3Final/img2img-with_cac-with_ft-with_mask/0.jpg} \\
        & \includegraphics[width=\linewidth,valign=m]{claim_1/cakeTable3Final/init.jpg} & \includegraphics[width=\linewidth,valign=m]{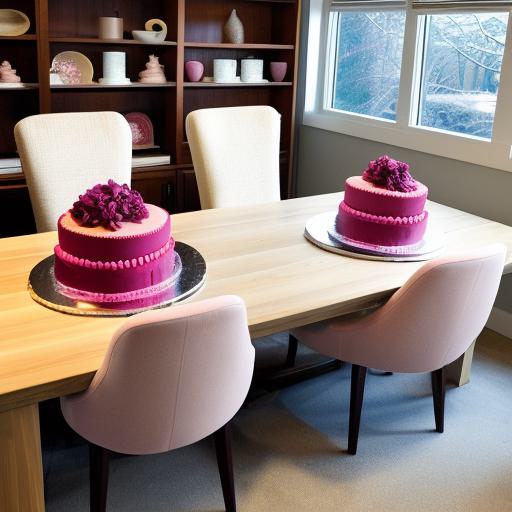} & \includegraphics[width=\linewidth,valign=m]{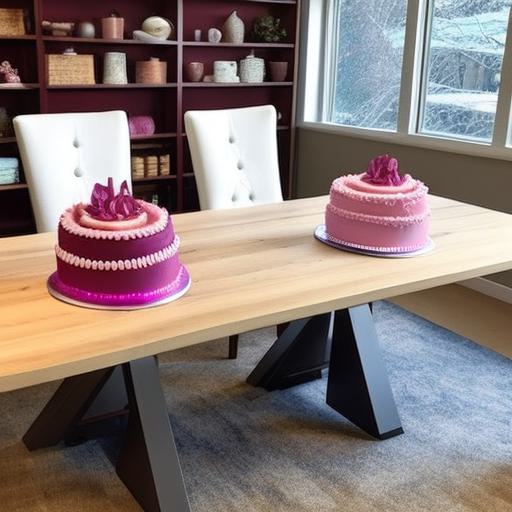} & \includegraphics[width=\linewidth,valign=m]{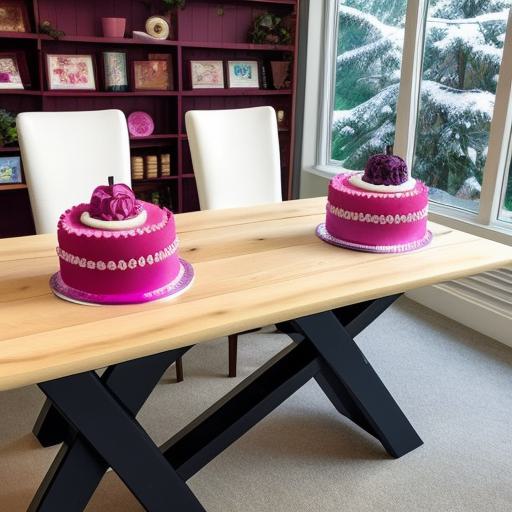} & \includegraphics[width=\linewidth,valign=m]{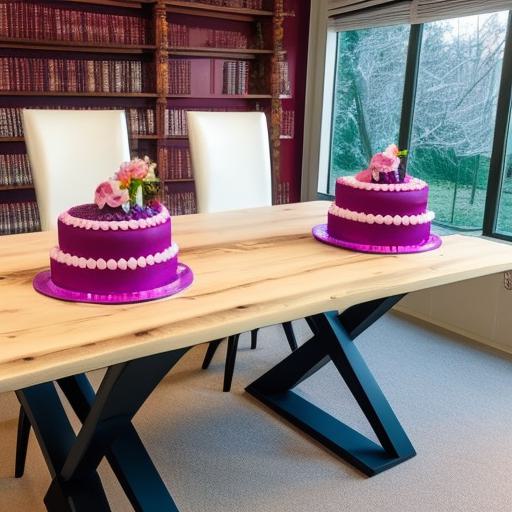} \\
        & \includegraphics[width=\linewidth,valign=m]{claim_1/cakeTable3Final/init.jpg} & \includegraphics[width=\linewidth,valign=m]{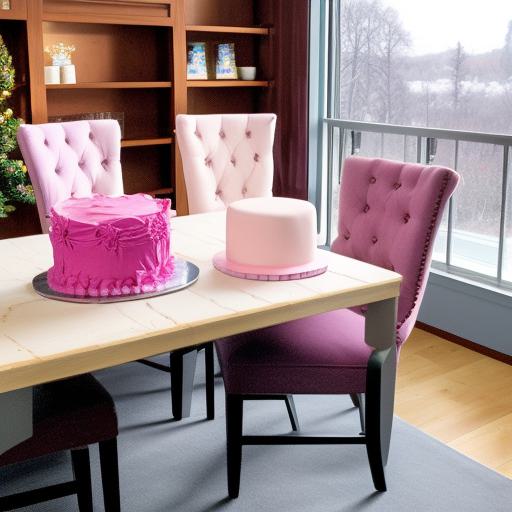} & \includegraphics[width=\linewidth,valign=m]{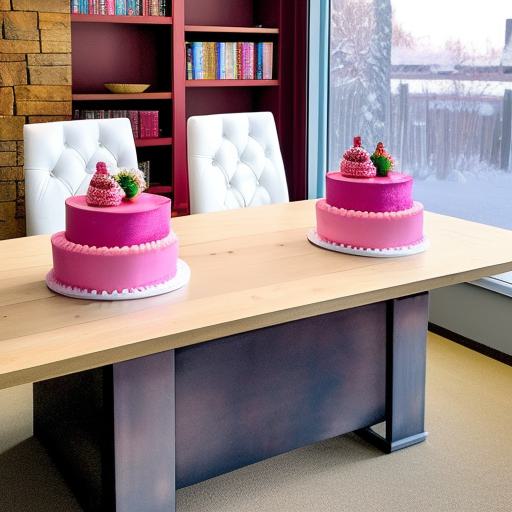} & \includegraphics[width=\linewidth,valign=m]{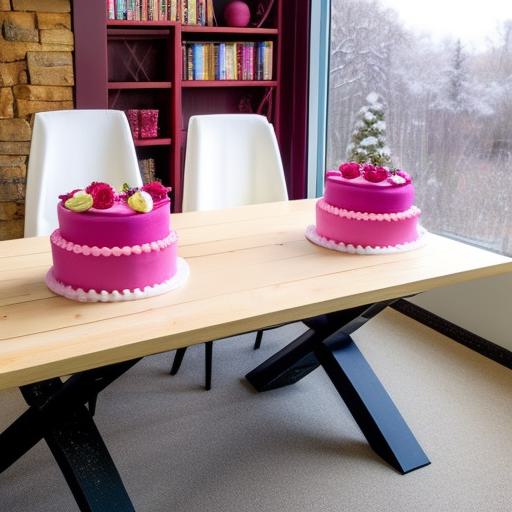} & \includegraphics[width=\linewidth,valign=m]{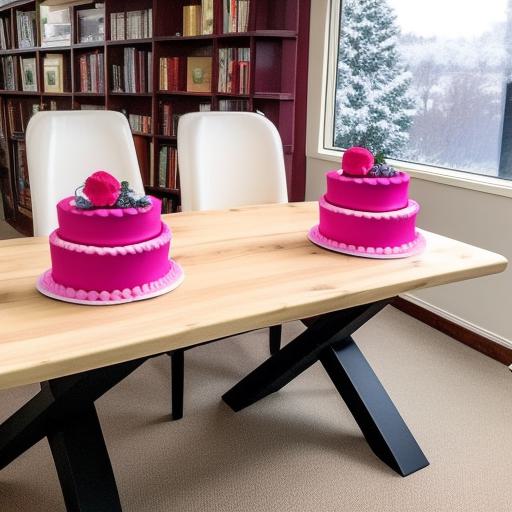} \\
        & \includegraphics[width=\linewidth,valign=m]{claim_1/cakeTable3Final/init.jpg} & \includegraphics[width=\linewidth,valign=m]{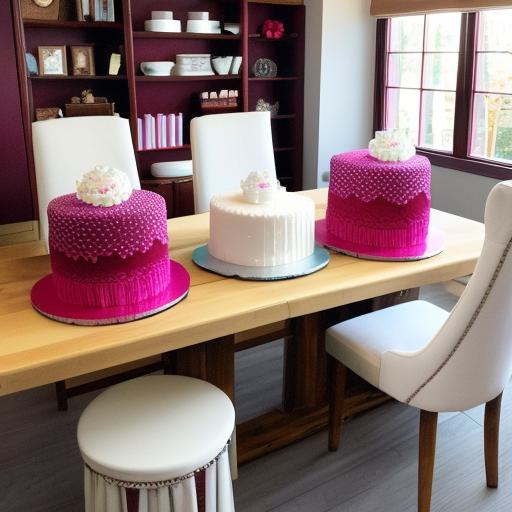} & \includegraphics[width=\linewidth,valign=m]{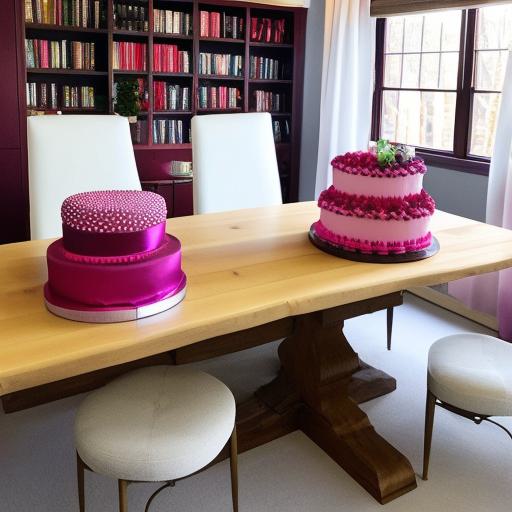} & \includegraphics[width=\linewidth,valign=m]{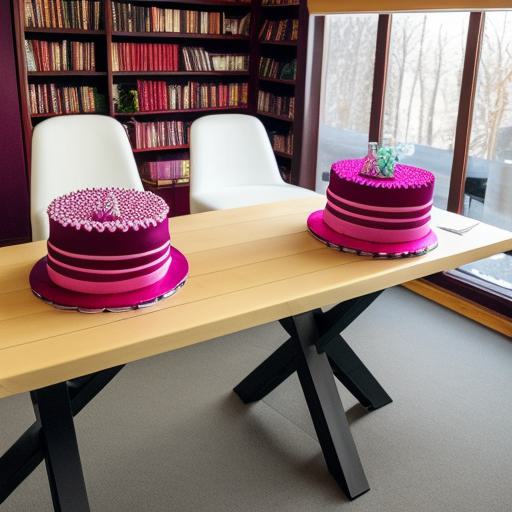} & \includegraphics[width=\linewidth,valign=m]{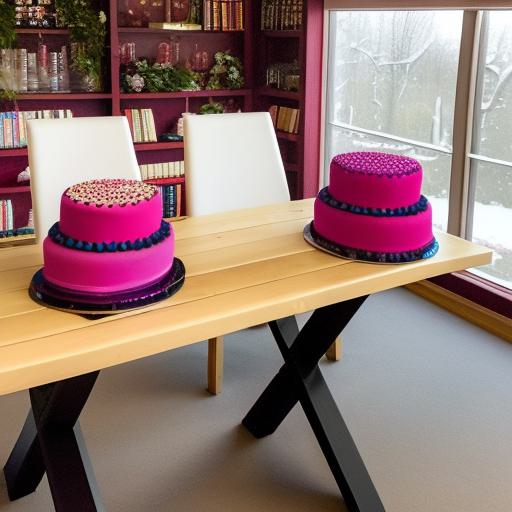} \\
        & \includegraphics[width=\linewidth,valign=m]{claim_1/cakeTable3Final/init.jpg} & \includegraphics[width=\linewidth,valign=m]{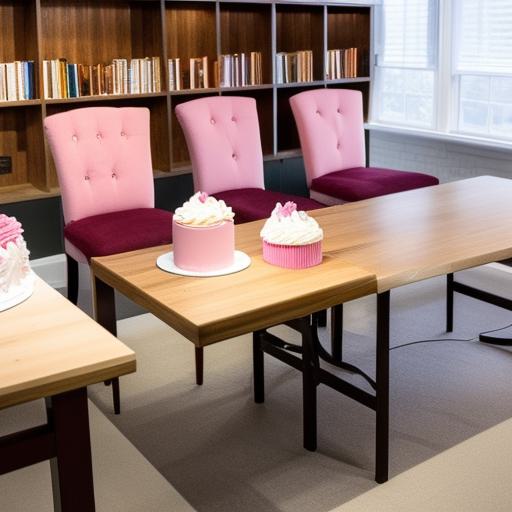} & \includegraphics[width=\linewidth,valign=m]{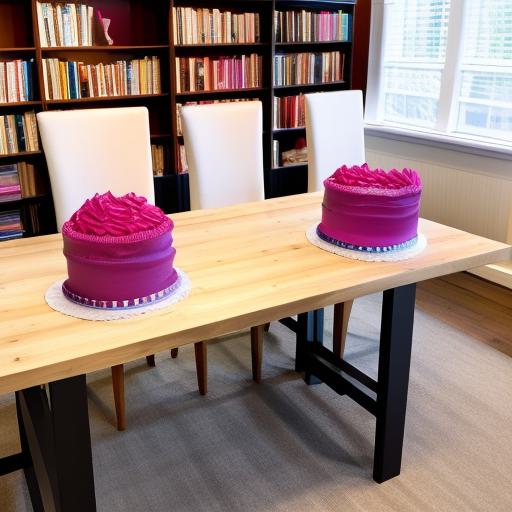} & \includegraphics[width=\linewidth,valign=m]{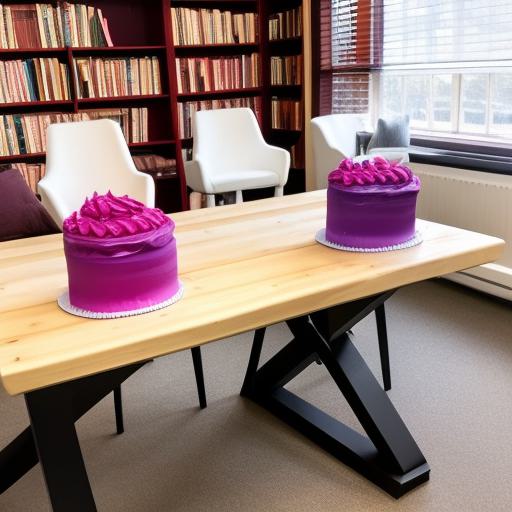} & \includegraphics[width=\linewidth,valign=m]{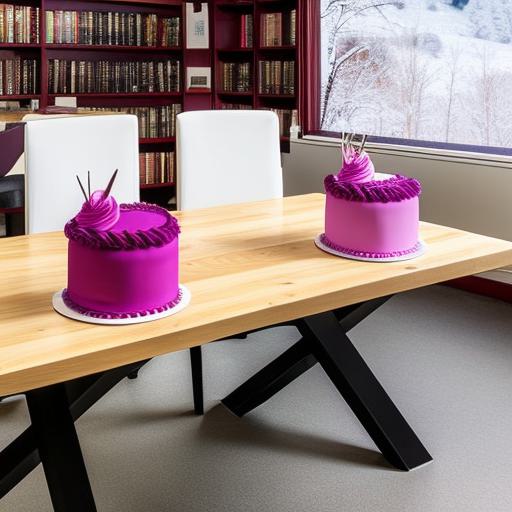} \\
        & \includegraphics[width=\linewidth,valign=m]{claim_1/cakeTable3Final/init.jpg} & \includegraphics[width=\linewidth,valign=m]{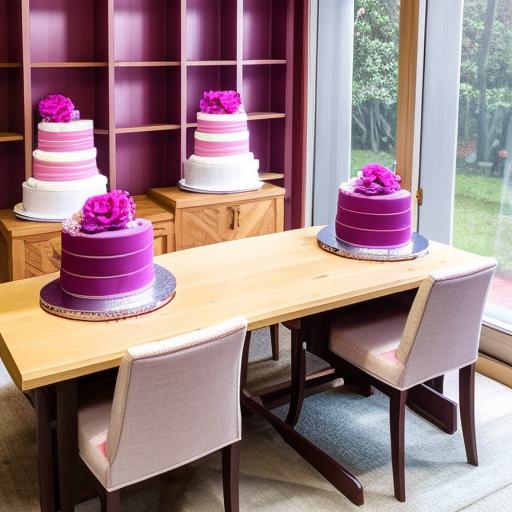} & \includegraphics[width=\linewidth,valign=m]{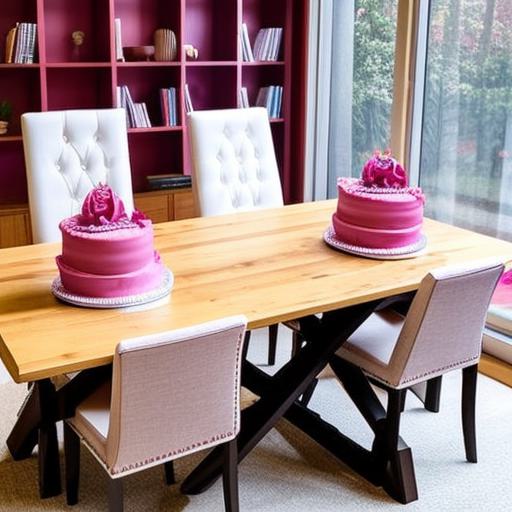} & \includegraphics[width=\linewidth,valign=m]{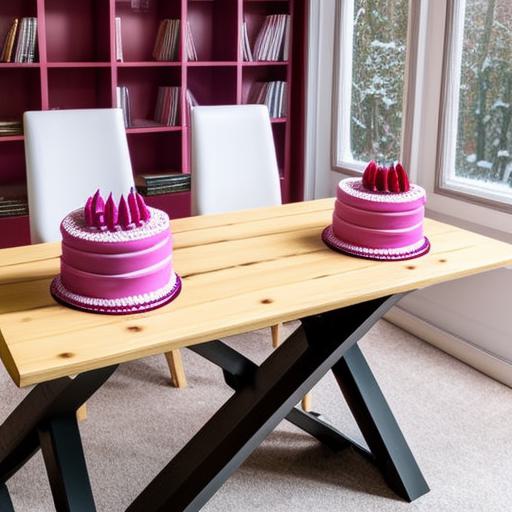} & \includegraphics[width=\linewidth,valign=m]{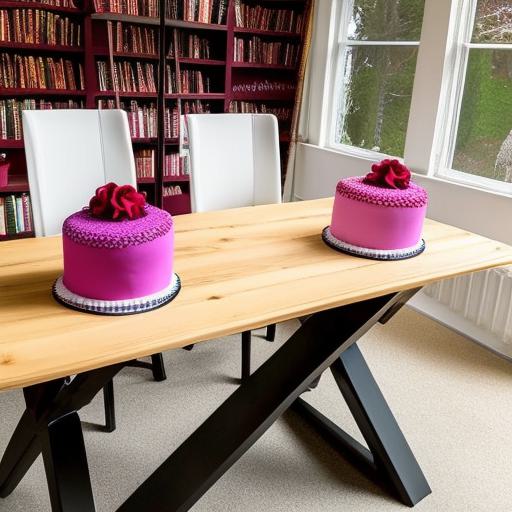} \\
    \end{tabular}
\end{adjustbox}
\caption{ ``a \ul{wood table} with two \ul{white chairs} behind, two \ul{decorated cakes} on top, \ul{maroon bookshelves} behind, and \ul{winter window}'' }
\label{fig:cakeTable3}
\end{figure}

\begin{figure}[!htbp]
    \centering
    \begin{adjustbox}{max size={\textwidth}{\textheight}}
    \begin{tabular}[t]{p{.0\linewidth}p{.2\linewidth}|p{.2\linewidth}p{.2\linewidth}p{.2\linewidth}p{.2\linewidth}}
        & \hfil\textbf{Input Layers} & \hfil\textbf{GH} & \hfil\textbf{GH+CA} & \hfil\textbf{GH+CA+TI} & \hfil\textbf{GH+CA+TI+LN}\\
        & \includegraphics[width=\linewidth,valign=m]{claim_1/fruit4Final/init.jpg} & \includegraphics[width=\linewidth,valign=m]{claim_1/fruit4Final/img2img-no_cac-no_ft-no_mask/6.jpg} & \includegraphics[width=\linewidth,valign=m]{claim_1/fruit4Final/img2img-with_cac-no_ft-no_mask/6.jpg} & \includegraphics[width=\linewidth,valign=m]{claim_1/fruit4Final/img2img-with_cac-with_ft-no_mask/6.jpg} & \includegraphics[width=\linewidth,valign=m]{claim_1/fruit4Final/img2img-with_cac-with_ft-with_mask/6.jpg} \\
         & \includegraphics[width=\linewidth,valign=m]{claim_1/fruit4Final/init.jpg} & \includegraphics[width=\linewidth,valign=m]{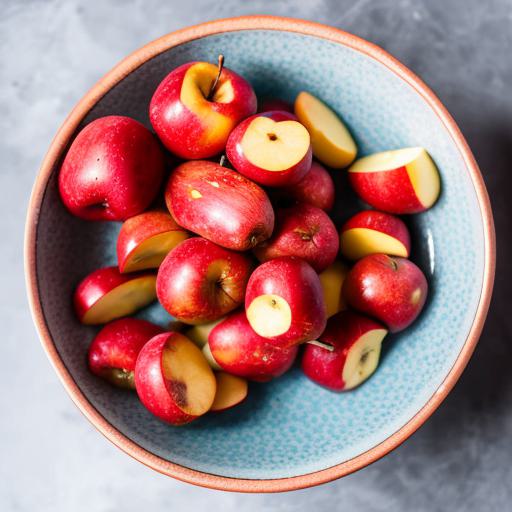} & \includegraphics[width=\linewidth,valign=m]{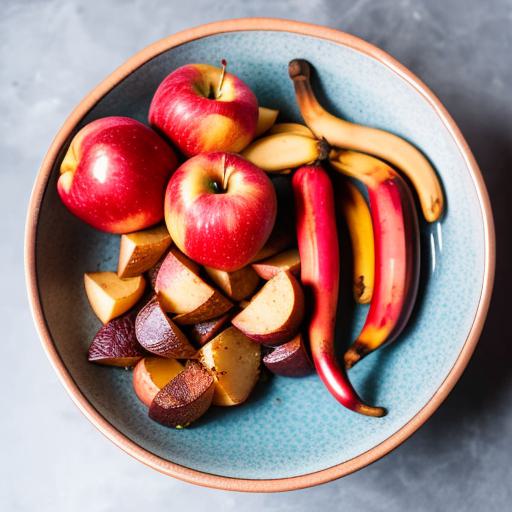} & \includegraphics[width=\linewidth,valign=m]{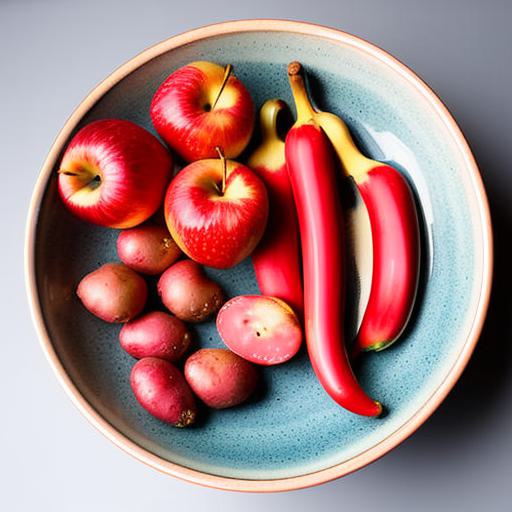} & \includegraphics[width=\linewidth,valign=m]{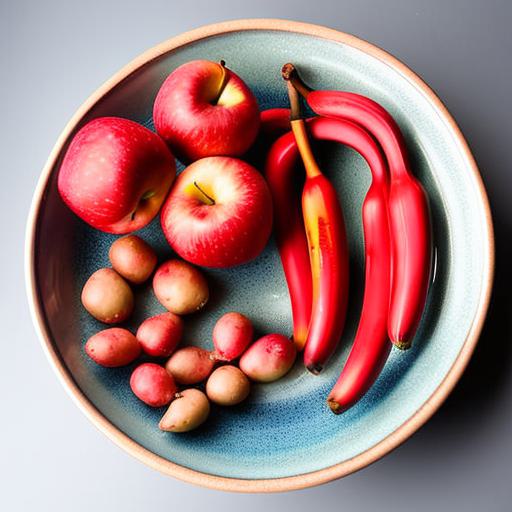} \\
        & \includegraphics[width=\linewidth,valign=m]{claim_1/fruit4Final/init.jpg} & \includegraphics[width=\linewidth,valign=m]{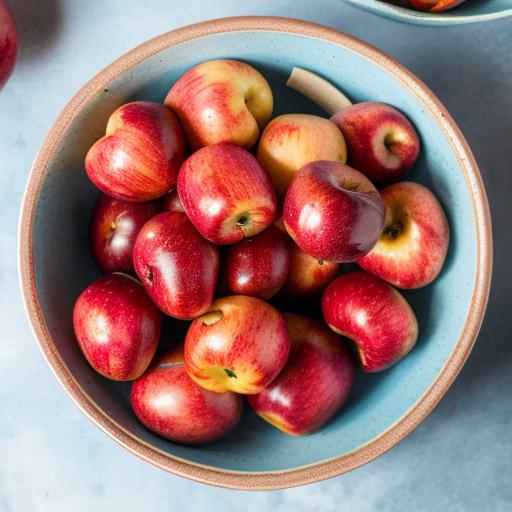} & \includegraphics[width=\linewidth,valign=m]{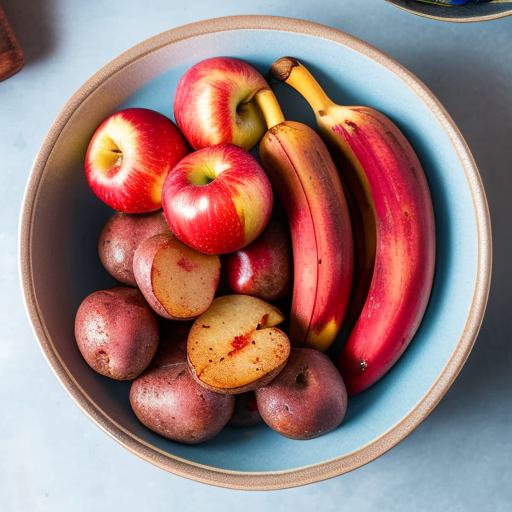} & \includegraphics[width=\linewidth,valign=m]{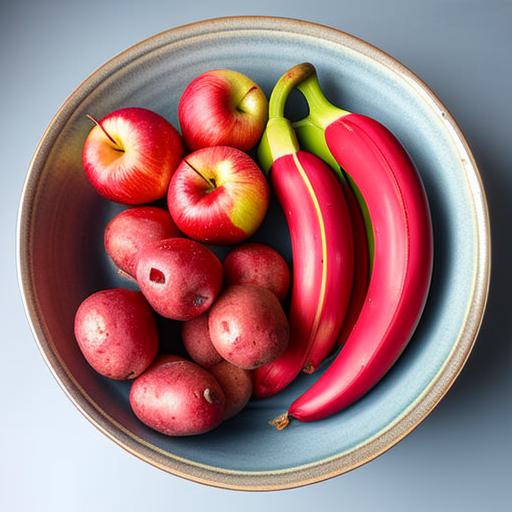} & \includegraphics[width=\linewidth,valign=m]{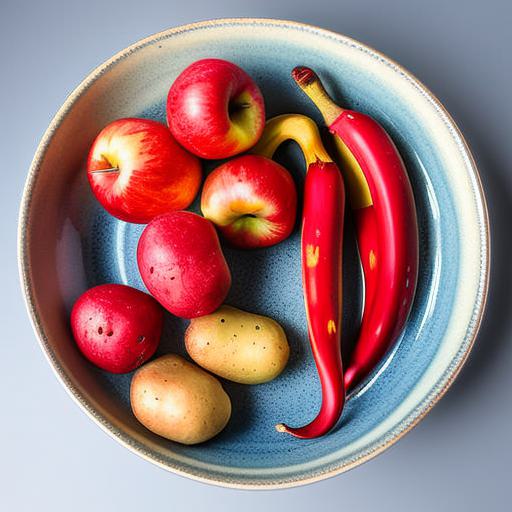} \\
        & \includegraphics[width=\linewidth,valign=m]{claim_1/fruit4Final/init.jpg} & \includegraphics[width=\linewidth,valign=m]{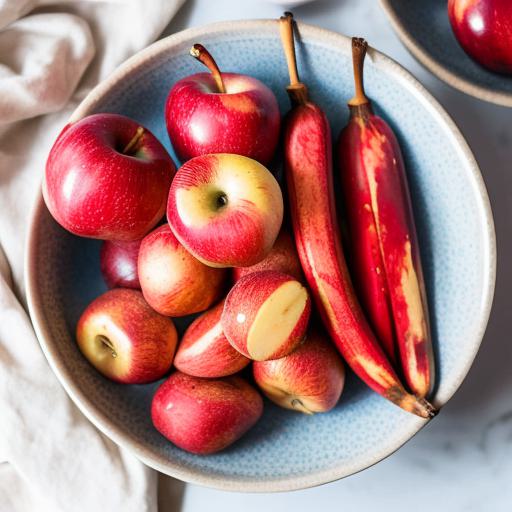} & \includegraphics[width=\linewidth,valign=m]{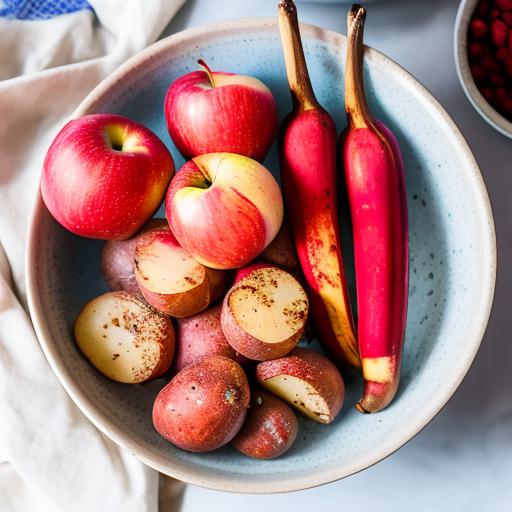} & \includegraphics[width=\linewidth,valign=m]{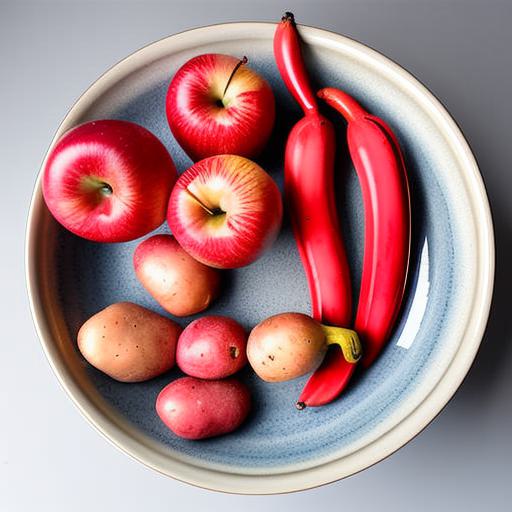} & \includegraphics[width=\linewidth,valign=m]{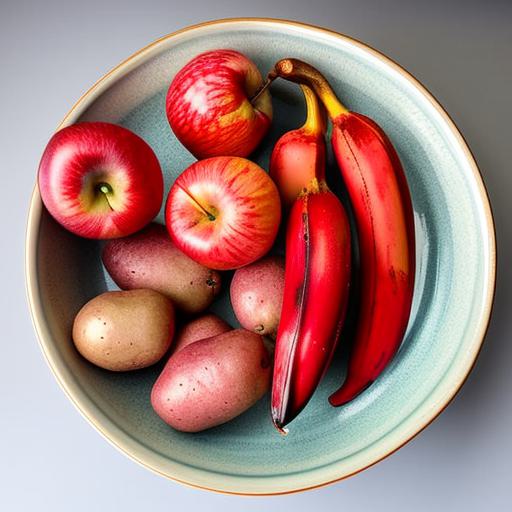} \\
        & \includegraphics[width=\linewidth,valign=m]{claim_1/fruit4Final/init.jpg} & \includegraphics[width=\linewidth,valign=m]{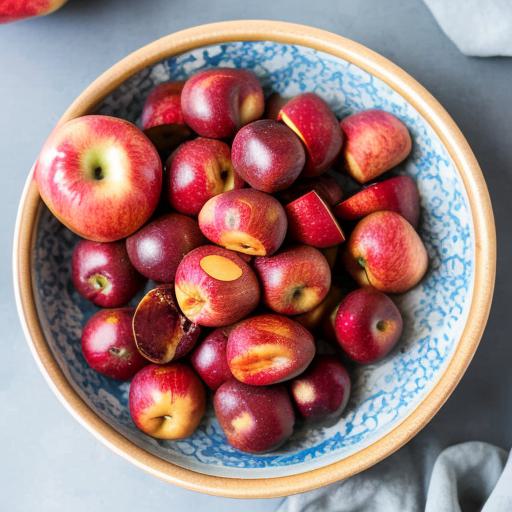} & \includegraphics[width=\linewidth,valign=m]{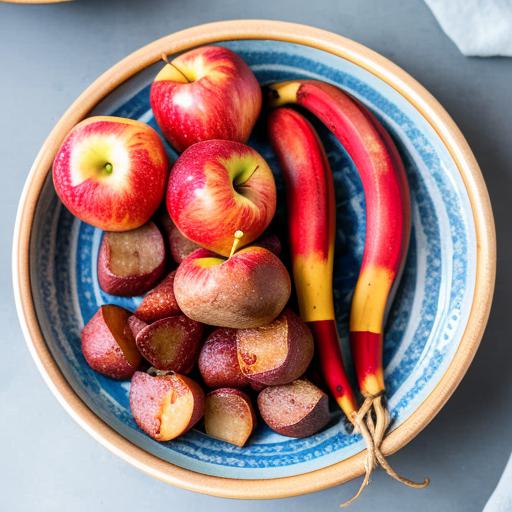} & \includegraphics[width=\linewidth,valign=m]{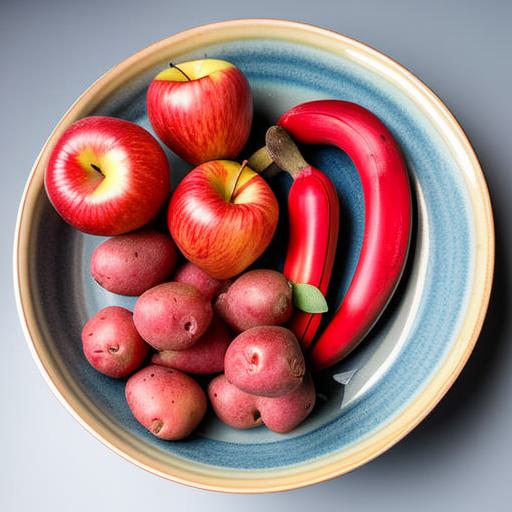} & \includegraphics[width=\linewidth,valign=m]{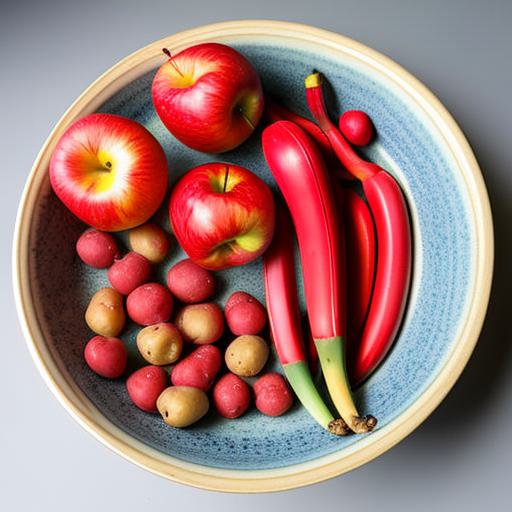} \\
        & \includegraphics[width=\linewidth,valign=m]{claim_1/fruit4Final/init.jpg} & \includegraphics[width=\linewidth,valign=m]{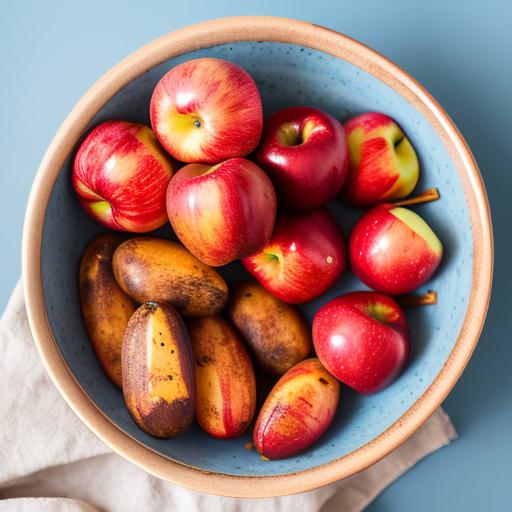} & \includegraphics[width=\linewidth,valign=m]{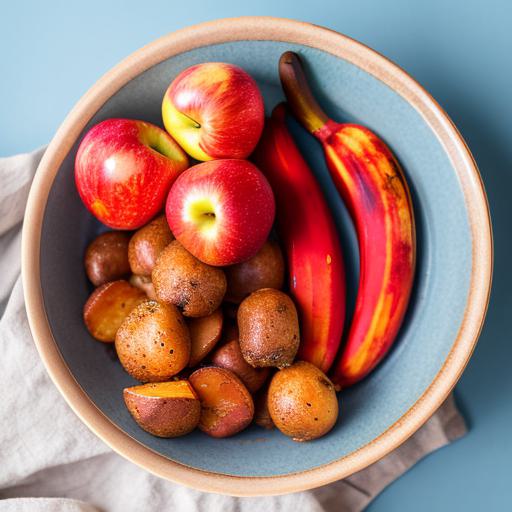} & \includegraphics[width=\linewidth,valign=m]{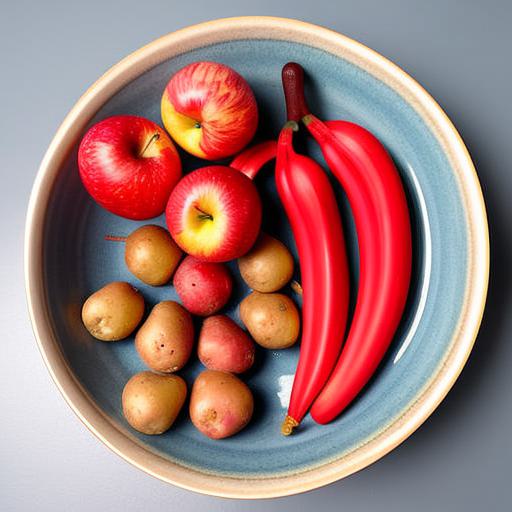} & \includegraphics[width=\linewidth,valign=m]{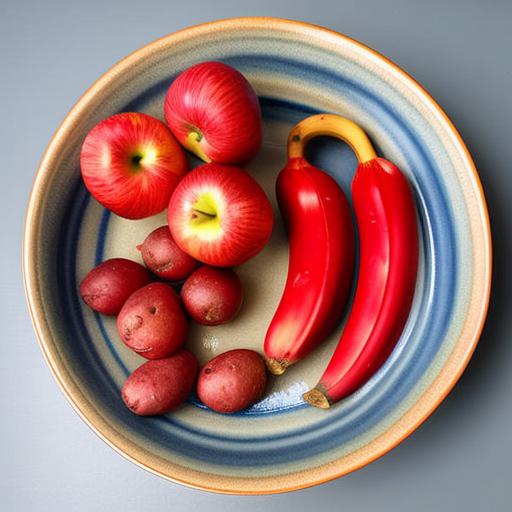} \\
    \end{tabular}
\end{adjustbox}
\caption{ ``a \ul{blue ceramic bowl} with \ul{red potatoes}, \ul{red apples}, and \ul{red bananas}'' }
\label{fig:fruit4}
\end{figure}

\begin{figure}[!htbp]
    \centering
    \begin{adjustbox}{max size={\textwidth}{\textheight}}
    \begin{tabular}[t]{p{.0\linewidth}p{.2\linewidth}|p{.2\linewidth}p{.2\linewidth}p{.2\linewidth}p{.2\linewidth}}
        & \hfil\textbf{Input Layers} & \hfil\textbf{GH} & \hfil\textbf{GH+CA} & \hfil\textbf{GH+CA+TI} & \hfil\textbf{GH+CA+TI+LN}\\
         & \includegraphics[width=\linewidth,valign=m]{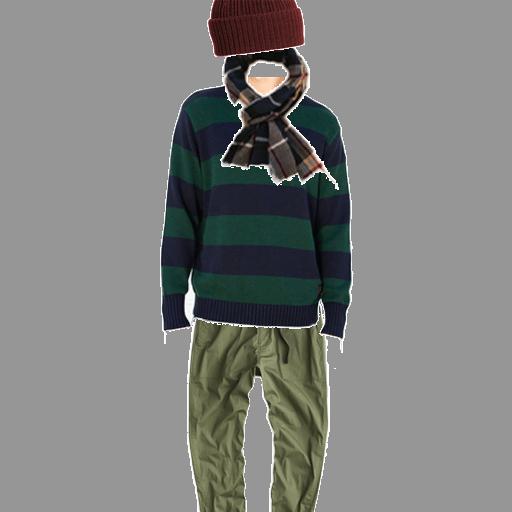} & \includegraphics[width=\linewidth,valign=m]{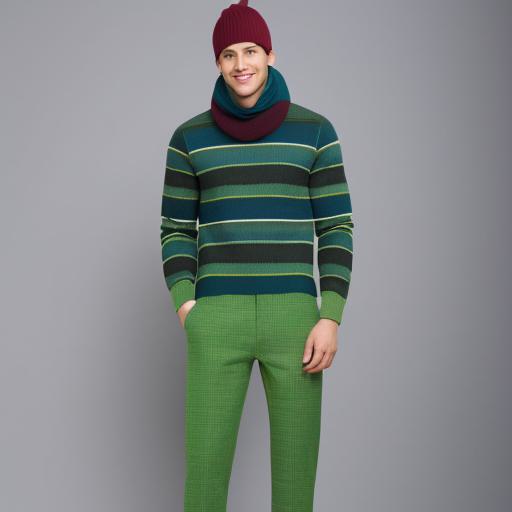} & \includegraphics[width=\linewidth,valign=m]{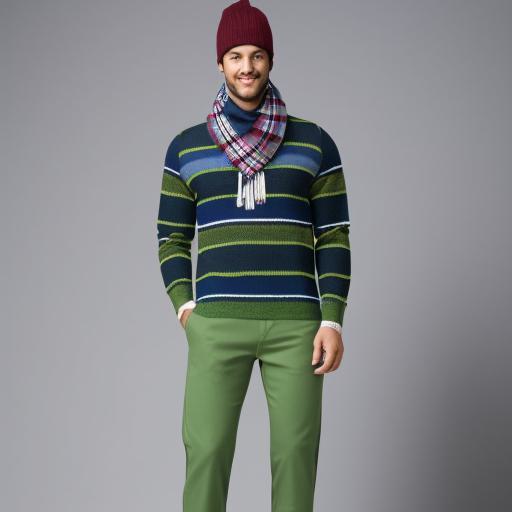} & \includegraphics[width=\linewidth,valign=m]{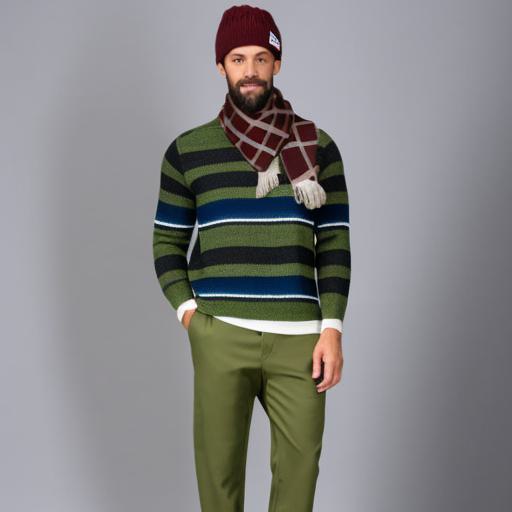} & \includegraphics[width=\linewidth,valign=m]{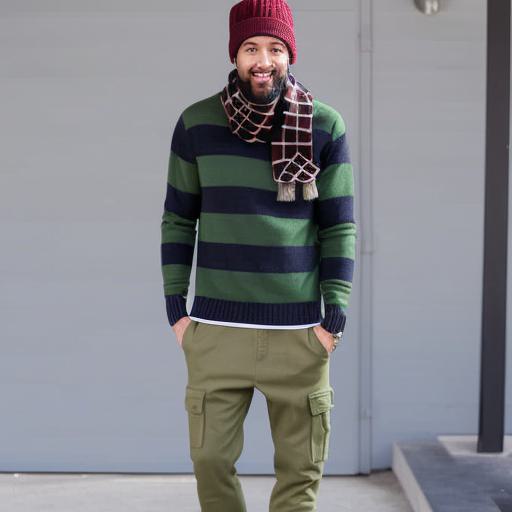} \\
        & \includegraphics[width=\linewidth,valign=m]{claim_1/clothing4Final/init.jpg} & \includegraphics[width=\linewidth,valign=m]{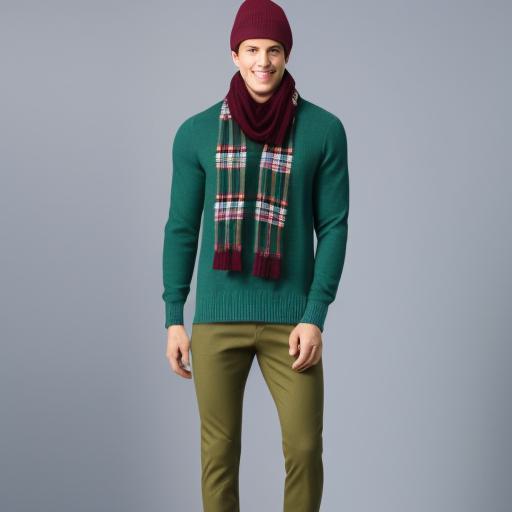} & \includegraphics[width=\linewidth,valign=m]{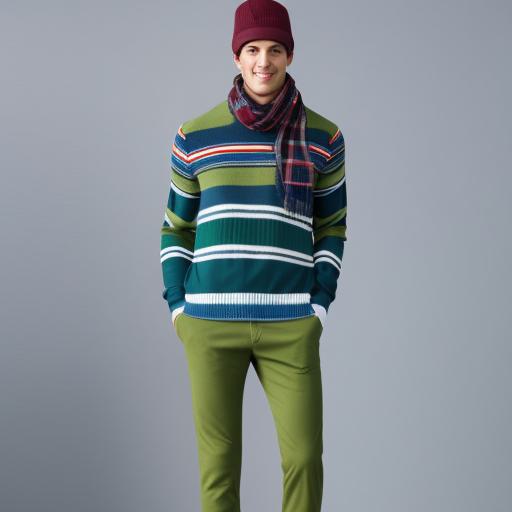} & \includegraphics[width=\linewidth,valign=m]{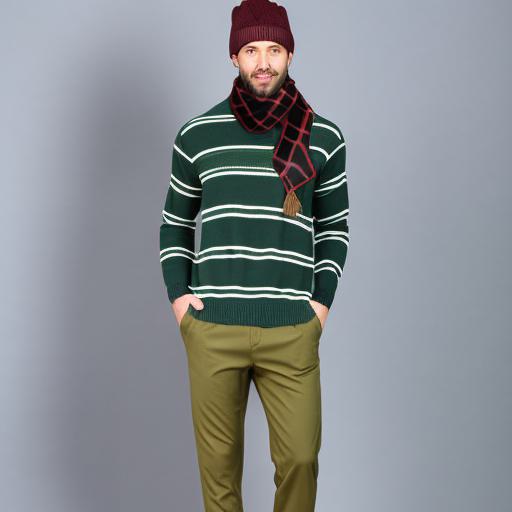} & \includegraphics[width=\linewidth,valign=m]{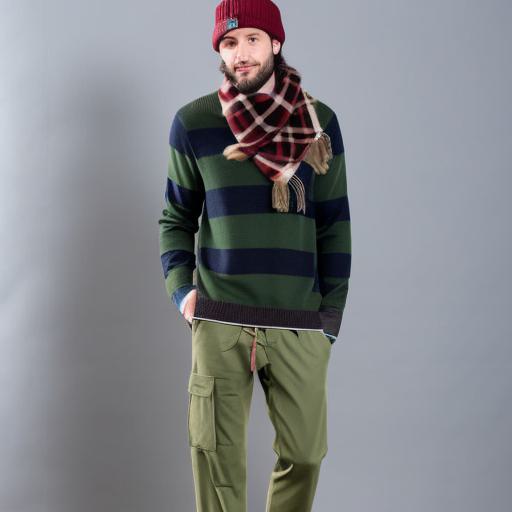} \\
        & \includegraphics[width=\linewidth,valign=m]{claim_1/clothing4Final/init.jpg} & \includegraphics[width=\linewidth,valign=m]{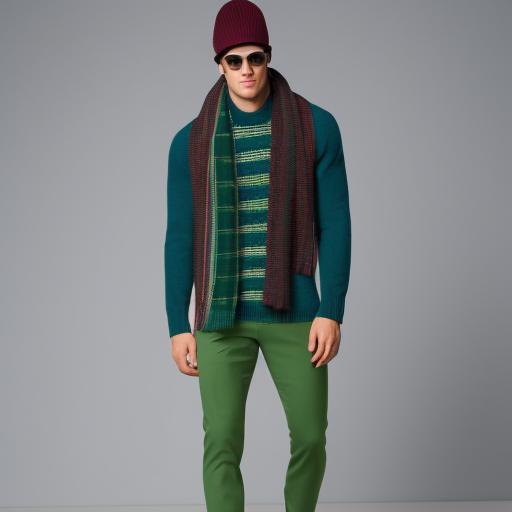} & \includegraphics[width=\linewidth,valign=m]{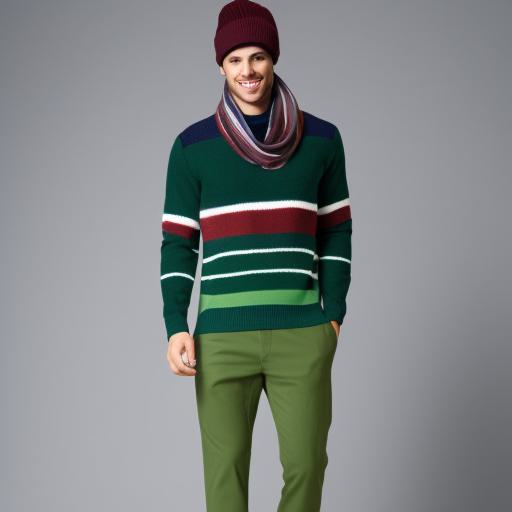} & \includegraphics[width=\linewidth,valign=m]{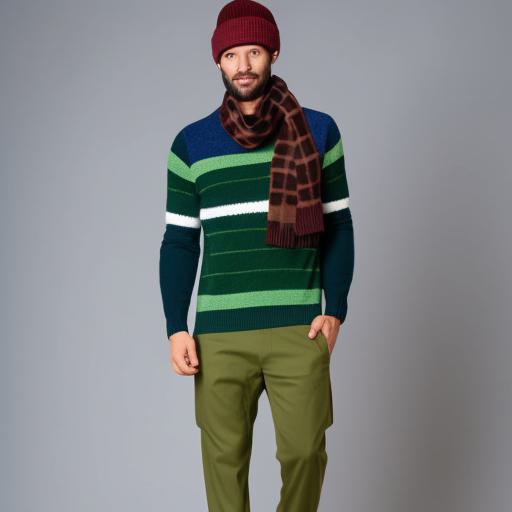} & \includegraphics[width=\linewidth,valign=m]{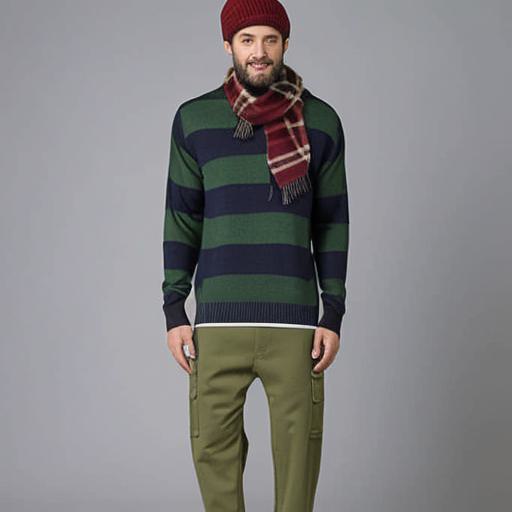} \\
        & \includegraphics[width=\linewidth,valign=m]{claim_1/clothing4Final/init.jpg} & \includegraphics[width=\linewidth,valign=m]{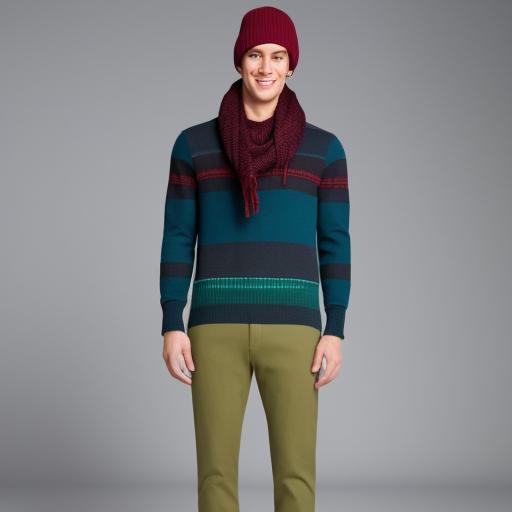} & \includegraphics[width=\linewidth,valign=m]{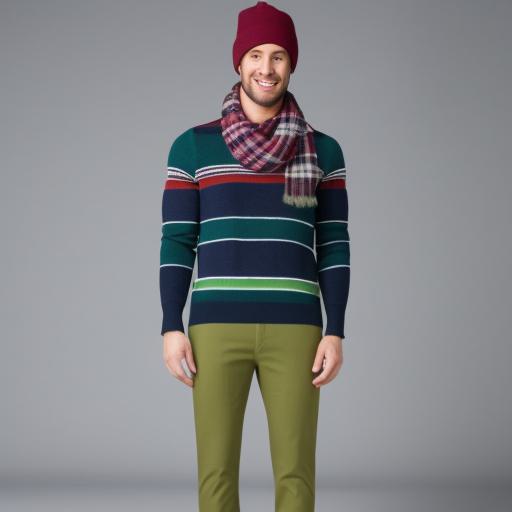} & \includegraphics[width=\linewidth,valign=m]{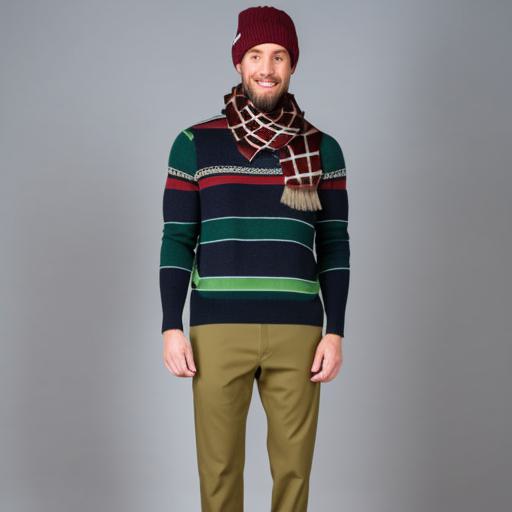} & \includegraphics[width=\linewidth,valign=m]{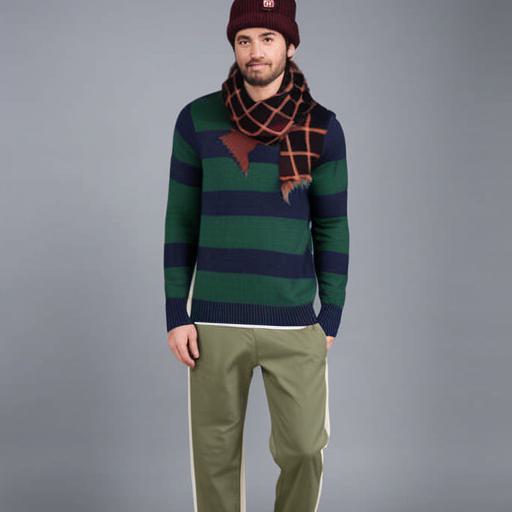} \\
        & \includegraphics[width=\linewidth,valign=m]{claim_1/clothing4Final/init.jpg} & \includegraphics[width=\linewidth,valign=m]{claim_1/clothing4Final/img2img-no_cac-no_ft-no_mask/7.jpg} & \includegraphics[width=\linewidth,valign=m]{claim_1/clothing4Final/img2img-with_cac-no_ft-no_mask/7.jpg} & \includegraphics[width=\linewidth,valign=m]{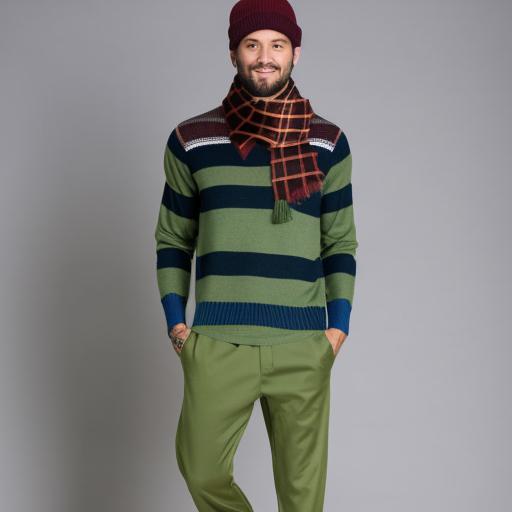} & \includegraphics[width=\linewidth,valign=m]{claim_1/clothing4Final/img2img-with_cac-with_ft-with_mask/7.jpg} \\
        & \includegraphics[width=\linewidth,valign=m]{claim_1/clothing4Final/init.jpg} & \includegraphics[width=\linewidth,valign=m]{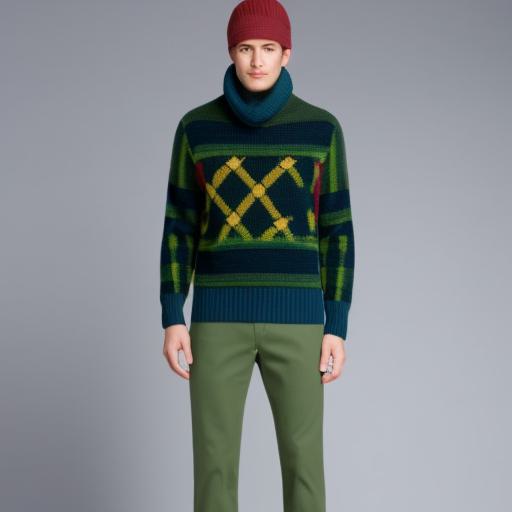} & \includegraphics[width=\linewidth,valign=m]{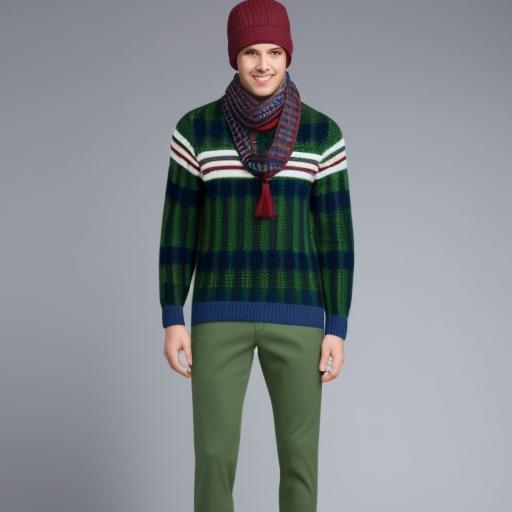} & \includegraphics[width=\linewidth,valign=m]{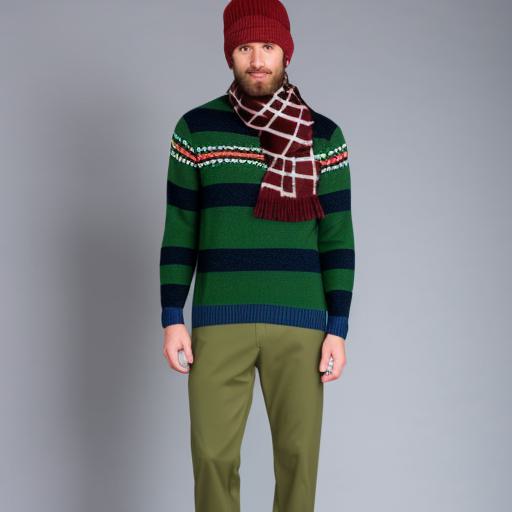} & \includegraphics[width=\linewidth,valign=m]{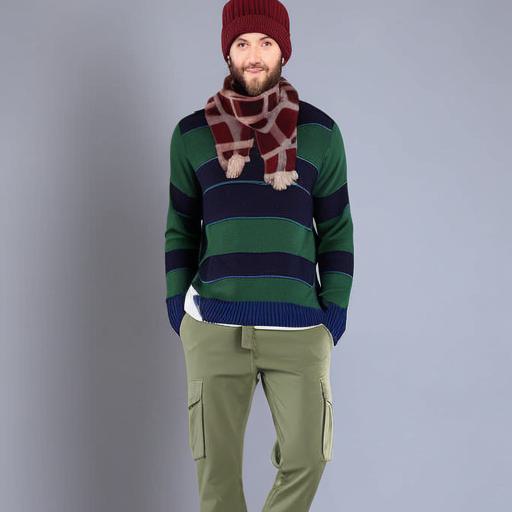} \\
    \end{tabular}
\end{adjustbox}
\caption{ ``a man wearing \ul{green pants}, a \ul{blue and green striped sweater}, a \ul{plaid scarf}, and a \ul{maroon beanie}'' }
\label{fig:clothing4}
\end{figure}

\begin{figure}[!htbp]
    \centering
    \begin{adjustbox}{max size={\textwidth}{\textheight}}
    \begin{tabular}[t]{p{.0\linewidth}p{.2\linewidth}|p{.2\linewidth}p{.2\linewidth}p{.2\linewidth}p{.2\linewidth}}
        & \hfil\textbf{Input Layers} & \hfil\textbf{GH} & \hfil\textbf{GH+CA} & \hfil\textbf{GH+CA+TI} & \hfil\textbf{GH+CA+TI+LN}\\
         & \includegraphics[width=\linewidth,valign=m]{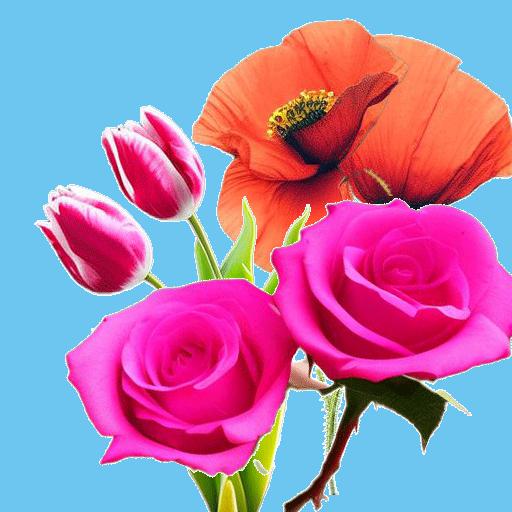} & \includegraphics[width=\linewidth,valign=m]{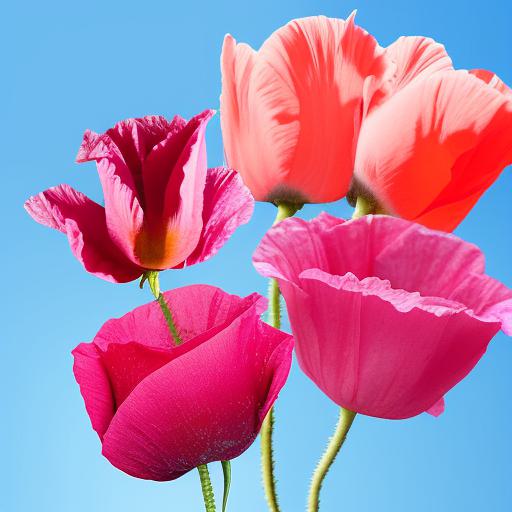} & \includegraphics[width=\linewidth,valign=m]{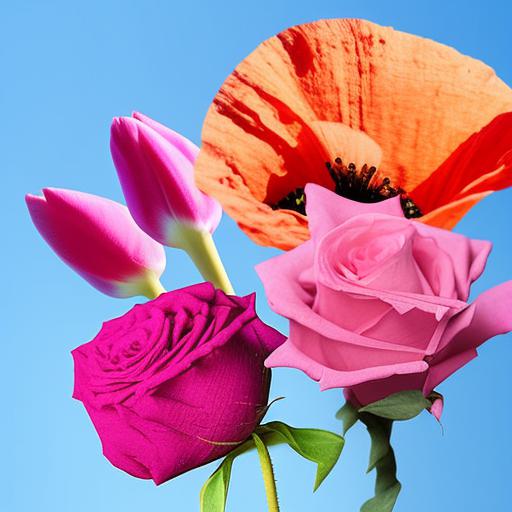} & \includegraphics[width=\linewidth,valign=m]{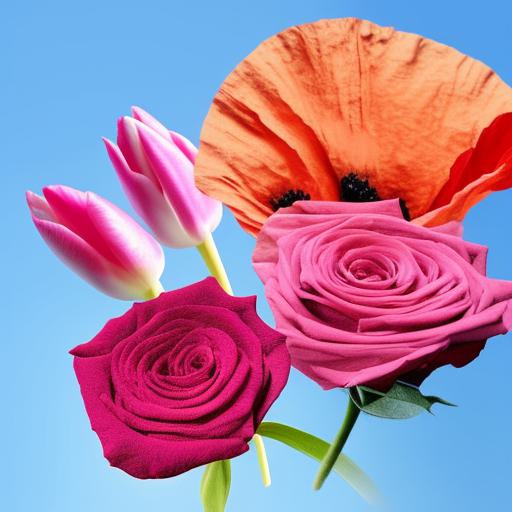} & \includegraphics[width=\linewidth,valign=m]{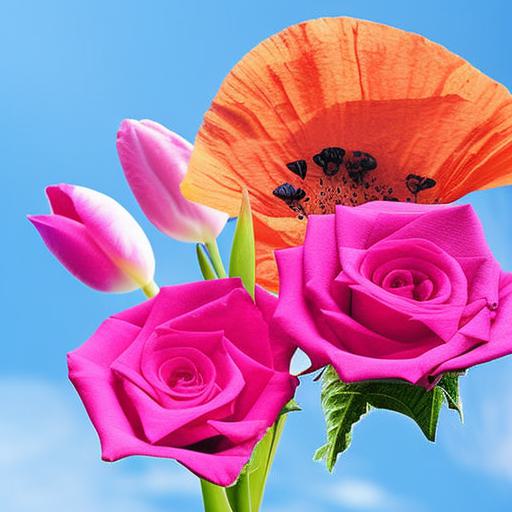} \\
        & \includegraphics[width=\linewidth,valign=m]{claim_1/flowers3Final/init.jpg} & \includegraphics[width=\linewidth,valign=m]{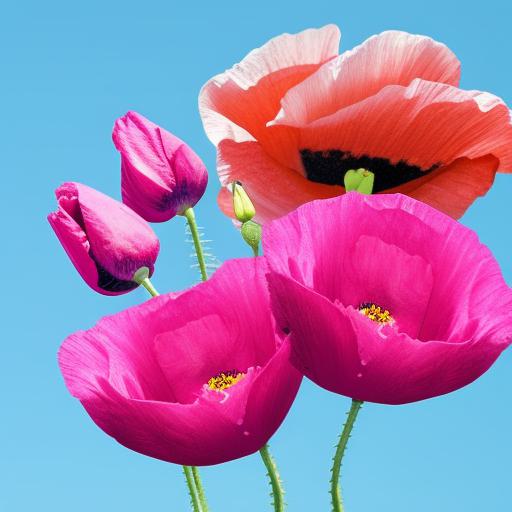} & \includegraphics[width=\linewidth,valign=m]{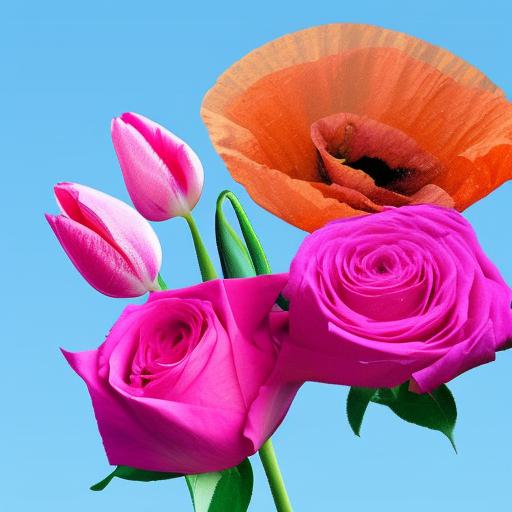} & \includegraphics[width=\linewidth,valign=m]{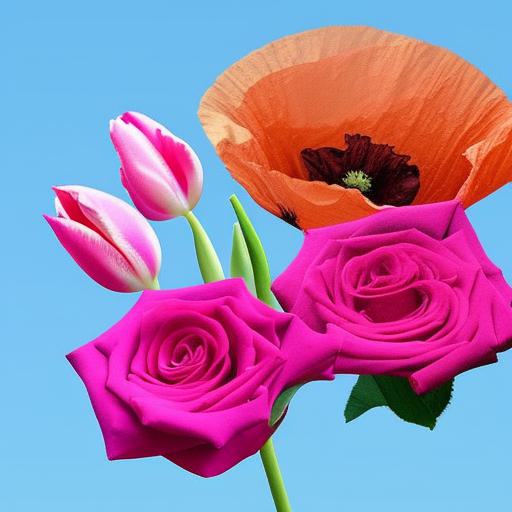} & \includegraphics[width=\linewidth,valign=m]{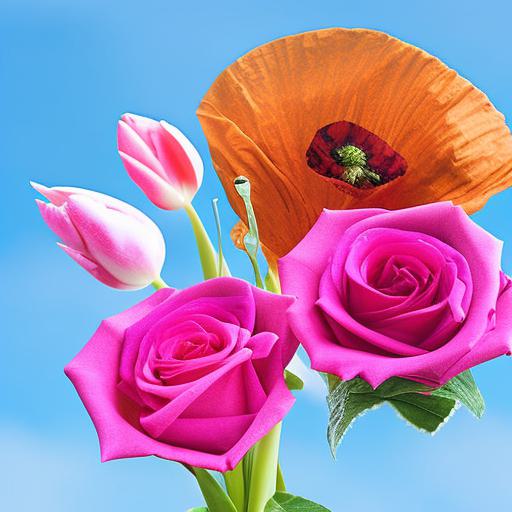} \\
        & \includegraphics[width=\linewidth,valign=m]{claim_1/flowers3Final/init.jpg} & \includegraphics[width=\linewidth,valign=m]{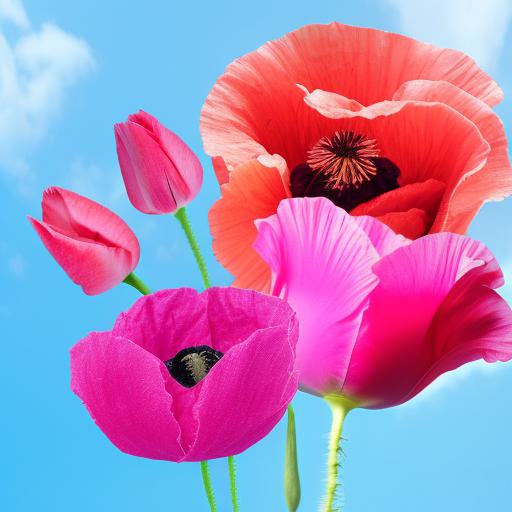} & \includegraphics[width=\linewidth,valign=m]{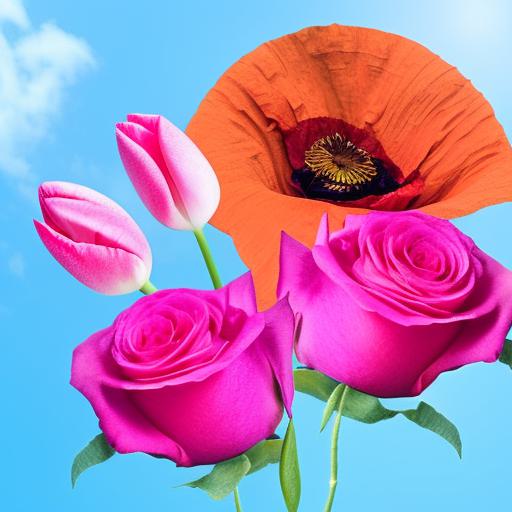} & \includegraphics[width=\linewidth,valign=m]{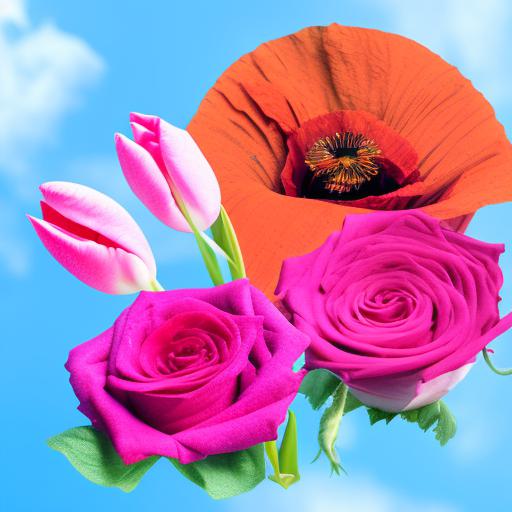} & \includegraphics[width=\linewidth,valign=m]{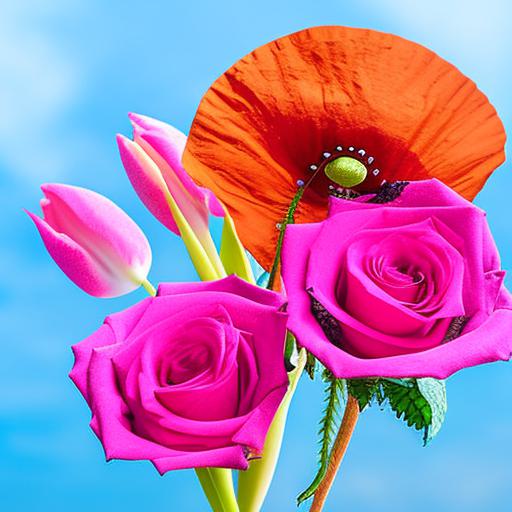} \\
        & \includegraphics[width=\linewidth,valign=m]{claim_1/flowers3Final/init.jpg} & \includegraphics[width=\linewidth,valign=m]{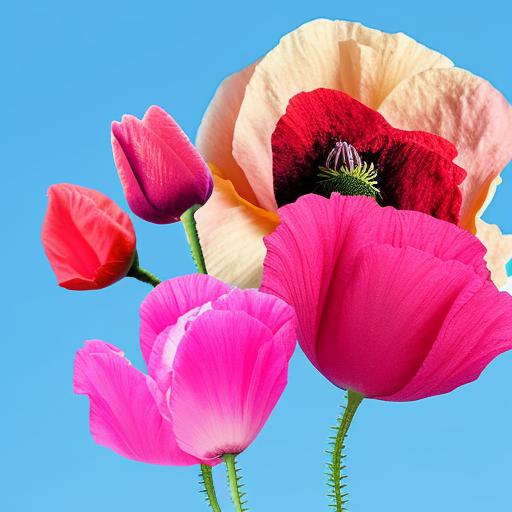} & \includegraphics[width=\linewidth,valign=m]{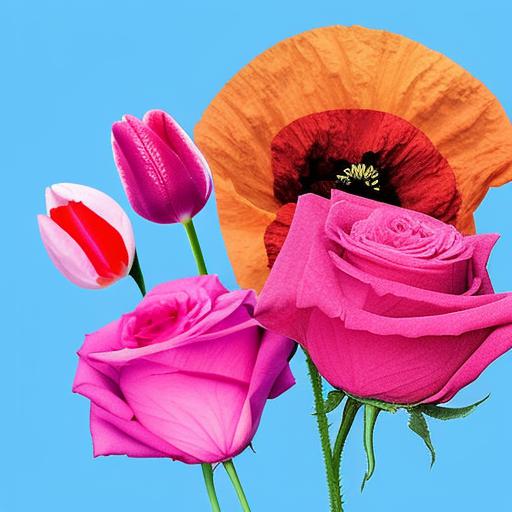} & \includegraphics[width=\linewidth,valign=m]{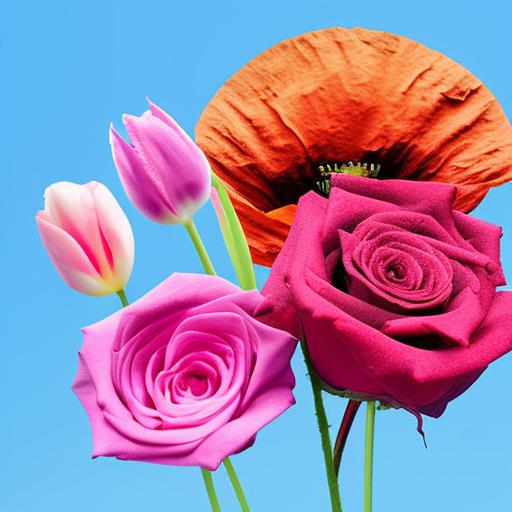} & \includegraphics[width=\linewidth,valign=m]{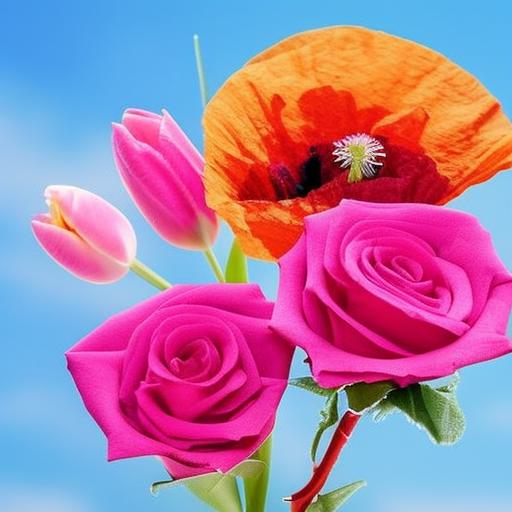} \\
        & \includegraphics[width=\linewidth,valign=m]{claim_1/flowers3Final/init.jpg} & \includegraphics[width=\linewidth,valign=m]{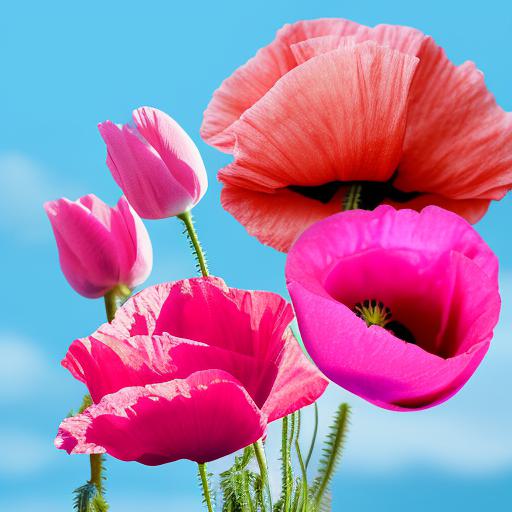} & \includegraphics[width=\linewidth,valign=m]{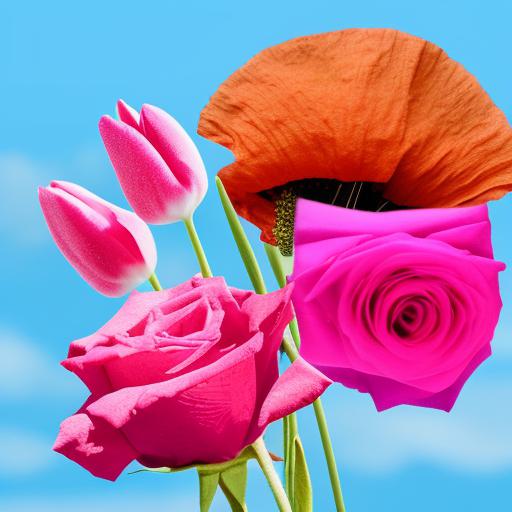} & \includegraphics[width=\linewidth,valign=m]{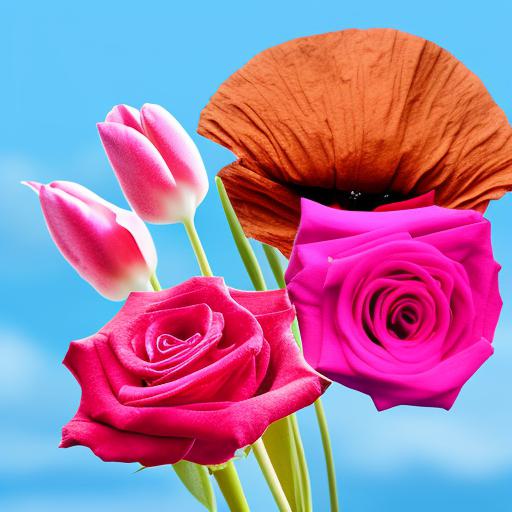} & \includegraphics[width=\linewidth,valign=m]{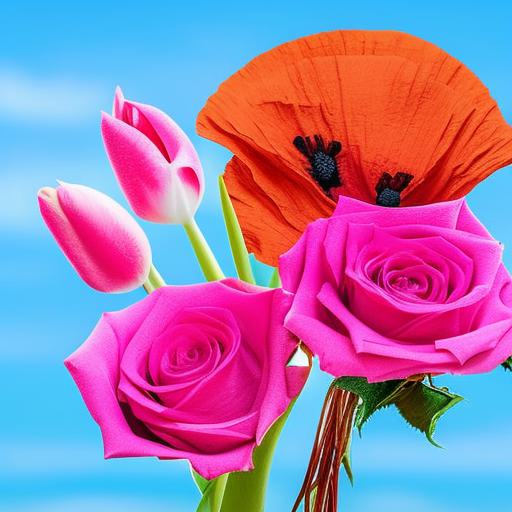} \\
        & \includegraphics[width=\linewidth,valign=m]{claim_1/flowers3Final/init.jpg} & \includegraphics[width=\linewidth,valign=m]{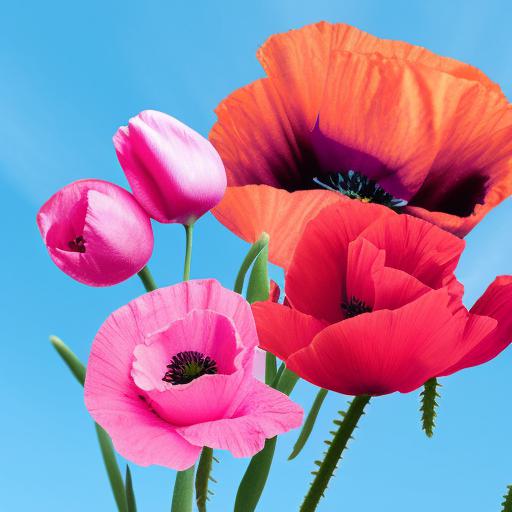} & \includegraphics[width=\linewidth,valign=m]{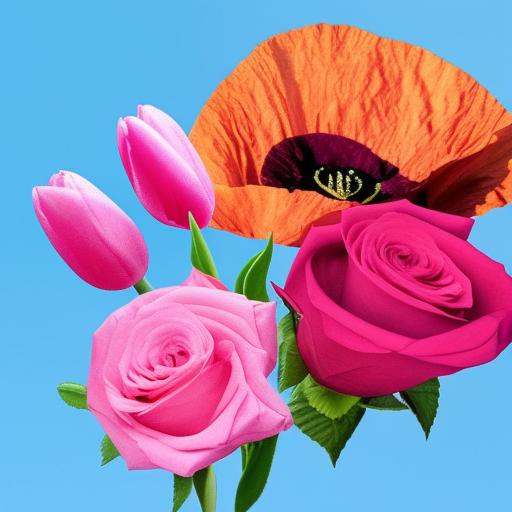} & \includegraphics[width=\linewidth,valign=m]{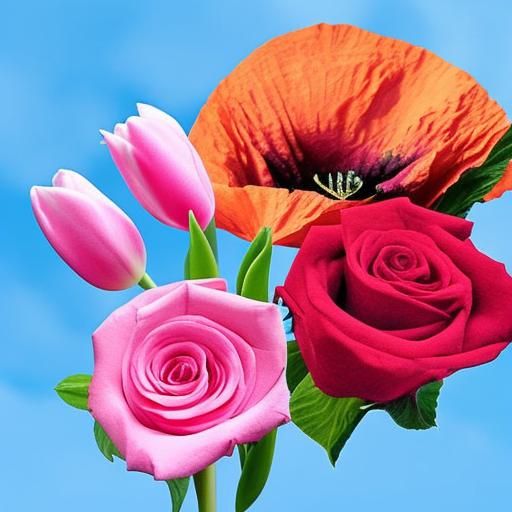} & \includegraphics[width=\linewidth,valign=m]{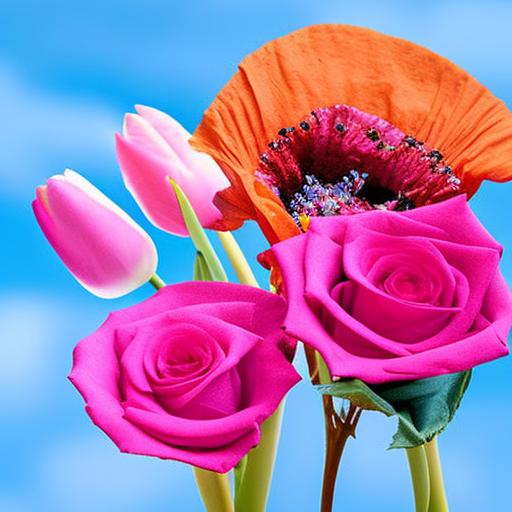} \\
    \end{tabular}
\end{adjustbox}
\caption{ ``a \ul{red poppy plant} and a \ul{pink rose plant} and a \ul{pink tulip plant} on a \ul{blue background}'' }
\label{fig:flowers3}
\end{figure}

Figures \ref{fig:toyBox} through \ref{fig:flowers3} contain additional results with different noise seeds for \textbf{GH}, \textbf{GH+CA}, and \textbf{GH+CA+TI}, highlighting that the trends of \textbf{CA} improving spatial fidelity and \textbf{TI} improving appearance fidelity hold across noise seeds for the collages tested. Additional results are not included for \textbf{SA} because the Plug-and-Play algorithm generates noise seeds through DDIM inversion, not at random \cite{tumanyan2022plug}.  

\end{document}